\documentclass{article} 
\usepackage{iclr2023_conference,times}
\usepackage{algorithm}
\usepackage{algorithmic}
\usepackage{newfloat}
\usepackage{listings}

\usepackage{amsmath,amsfonts,bm}

\def\eqref#1{equation~\ref{#1}}

\def\1{\bm{1}}

\def\va{{\bm{a}}}

\def\vs{{\bm{s}}}

\def\mQ{{\bm{Q}}}

\DeclareMathAlphabet{\mathsfit}{\encodingdefault}{\sfdefault}{m}{sl}
\SetMathAlphabet{\mathsfit}{bold}{\encodingdefault}{\sfdefault}{bx}{n}

\def\gA{{\mathcal{A}}}

\def\gP{{\mathcal{P}}}

\def\gS{{\mathcal{S}}}
\def\gT{{\mathcal{T}}}

\def\sP{{\mathbb{P}}}

\newcommand{\E}{\mathbb{E}}

\newcommand{\R}{\mathbb{R}}

\DeclareMathOperator*{\argmax}{arg\,max}
\DeclareMathOperator*{\argmin}{arg\,min}

\usepackage[colorlinks=true,citecolor=blue]{hyperref}
\usepackage{url}
\usepackage{amsmath}
\usepackage{graphicx}
\usepackage{booktabs}
\usepackage{multirow}
\usepackage{makecell}
\usepackage{amsthm}
\usepackage{xcolor}
\usepackage{tcolorbox} 

\usepackage{enumitem}
\usepackage{multirow} 
\usepackage{array}
\usepackage{graphicx}
\usepackage{tikz}
\usepackage{callouts}
\usepackage{bbm}
\usepackage{caption,subcaption}
\usepackage{mathtools}
\usepackage{amsthm}
\usepackage[textfont=small,labelfont=small]{caption}
\usepackage{wrapfig}
\usepackage{hyperref}
\usepackage{amsfonts}
\usepackage{lineno,hyperref}

\newtheorem{example}{Example}
\newtheorem{problem}{Problem}
\newtheorem{theorem}{Theorem}
\newtheorem{lemma}{Lemma} 
\newtheorem{assumption}{Assumption} 
 
\newtheorem{remark}{Remark}

\newtheorem{definition}{Definition}

\newcounter{bxincomm}
\definecolor{aqua}{rgb}{0.00,0.67,0.80}

\newcommand{\eg}{\textit{e}.\textit{g}., }
\newcommand{\ie}{\textit{i}.\textit{e}., }

\title{LLQL: Logistic Likelihood Q-Learning for Reinforcement Learning}
 \author{Outongyi Lv \\
   Shanghai Jiao Tong University \\
   \texttt{harry\_lv@sjtu.edu.cn} \\
   \And
   Bingxin Zhou \\
   Shanghai Jiao Tong University \\
   \texttt{bingxin.zhou@sjtu.edu.cn} \\
}

\iclrfinalcopy 

\begin{document}
\maketitle

\begin{abstract}
Modern reinforcement learning (RL) can be categorized into online and offline variants. As a pivotal aspect of both online and offline RL, current research on the Bellman equation revolves primarily around optimization techniques and performance enhancement rather than exploring the inherent structural properties of the Bellman error, such as its distribution characteristics. This study investigates the distribution of the Bellman approximation error through iterative exploration of the Bellman equation with the observation that the Bellman error approximately follows the Logistic distribution. Based on this, we proposed the utilization of the Logistic maximum likelihood function ($\rm LLoss$) as an alternative to the commonly used mean squared error ($\rm MSELoss$) that assumes a Normal distribution for Bellman errors. We validated the hypotheses through extensive numerical experiments across diverse online and offline environments. In particular, we applied the Logistic correction to loss functions in various RL baseline methods and observed that the results with $\rm LLoss$ consistently outperformed the MSE counterparts. We also conducted the Kolmogorov–Smirnov tests to confirm the reliability of the Logistic distribution. Moreover, our theory connects the Bellman error to the proportional reward scaling phenomenon by providing a distribution-based analysis. Furthermore, we applied the \emph{Bias–Variance decomposition} for sampling from the Logistic distribution. The theoretical and empirical insights of this study lay a valuable foundation for future investigations and enhancements centered on the distribution of Bellman error.
\end{abstract}

\section{Introduction}
\label{sec:intro}
Modern Deep Reinforcement Learning (RL) has witnessed remarkable advancements in diverse applications, encompassing strategy games \citep{mnih2013playing,kaiser2019model,qi2023adaptive} to Capacitated Vehicle Routing Problem (CVRP) problems \citep{kwon2020pomo,hottung2021efficient,bi2022learning}. RL operates by guiding an agent to actively interact with an environment through a series of actions, aiming to maximize the expectation of rewards over time. The cumulative reward concerning the current state is captured by the \emph{Bellman equation} \citep{bellman1954theory}.  Although the Bellman equation's recursive nature theoretically guides conventional RL towards optimal or near-optimal solutions, its computational demands raise concerns, especially when dealing with extensive state and action spaces \citep{patterson2022generalized}. In the realm of online RL,  \emph{Soft Actor Critic} (SAC) \cite{haarnoja2018soft2,haarnoja2018soft}, incorporates the soft Bellman operator to enhance the overall reward. This method greatly improves the performance and stability of the model. It also catalyzes some significant advancements in RL techniques \citep{christodoulou2019soft,ward2019improving,pan2023reinforcement}. While SAC is primarily designed for online RL, on a parallel front, offline RL has underscored concerns regarding substantial overestimation in action (Q-value) estimations \cite{kumar2020conservative}. 
This insight prompted subsequent developments of the Conservative Q-Learning (CQL) framework, sparking renewed interest in refining offline RL methodologies \citep{bayramouglu2021engagement,lyu2022mildly,kostrikov2021offline,garg2023extreme}. These important results provide a good foundation for the development of RL theory.

The conventional practice of employing Bellman equations for Q-iterations has gradually waned in modern RL discourse. Instead, a shift of preference has been observed in updating the iterative Q-function with the maximum-entropy policy to ensure robust modeling and mitigate estimation errors using the Bellman operator \citep{ziebart2010modeling}. SAC deploys an auxiliary policy network to circumvent intractable estimations over log-partitioned Q-values. More recently, Extreme Q-Learning (XQL) \citep{garg2023extreme} defines a novel sample-free objective towards optimal soft-value functions in the maximum entropy RL setting, thereby obviating the necessity for conventional network iterations. These frameworks mark a significant departure from established practices and offer exciting prospects for advancements in RL optimization techniques \citep{hansen2023idql,hejna2023inverse}. 

\begin{figure*}[t]
    \centering
    \includegraphics[width=\textwidth]{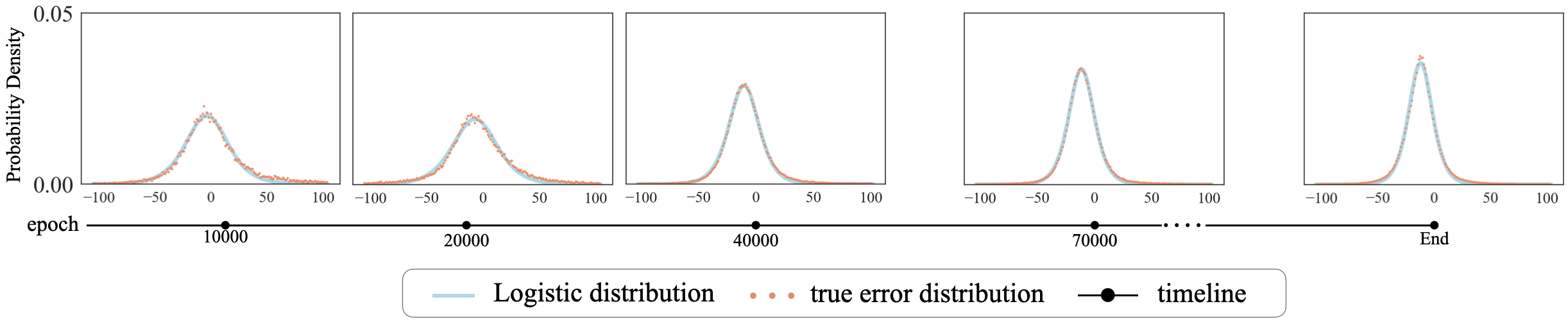}
    \caption{The evolving distributions of Bellman error (defined by (\ref{Eq11})) at different epochs on the online \textbf{LunarLanderContinuous-v2} environment.}
    \label{fig:timeline}
\end{figure*}

In parallel with the evolution of optimization techniques, researchers have exhibited a substantial interest in minimizing the \emph{Bellman error} \citep{baird1995residual}, a metric denoting the disparity between the current state-action value estimation and the value outlined by the Bellman equation. The objective is to precisely represent the value function of state-action pairs under the current policy \citep{geist2017bellman}. Following these efforts, researchers have strived to modify the objective function \citep{feng2019kernel,dai2018sbeed,fujimoto2022should} or optimize the update rules \citep{bas2021logistic,gong2020stable} to minimize the Bellman error. However, despite various attempts to achieve an adequate policy by indirectly addressing the distribution of the Bellman error, there lacks a straightforward analysis of the main properties of the Bellman error, particularly in terms of exploring more suitable error distributions beyond the Normal distribution.

In addition to Bellman errors, another important factor affecting the training effectiveness of RL is the reward scaling problem. It can directly influence whether the agent can find the optimal solution to the goal. However, the notion that a larger scaling factor inherently leads to better agent performance doesn't hold universally, as observed in SAC \citep{haarnoja2018soft}. There's an upper limit to scaling where surpassing it reverses the effect on agent training. The established scaling laws \citep{gao2023scaling, osband2019behaviour} offer valuable insights from the gradient perspective \citep{cabi2019scaling, zhang2020deep}. However, the absence of a theoretically sound scaling bound relegates it to an empirical hyperparameter.

In this paper, we attempt to address the following critical
question: \textit{What distribution better characterizes the Bellman error with an arbitrary Q-network?} We analyzed the close connection between the Logistic distribution and the Bellman error. Building on this foundation, we naturally integrated it with the reward scaling problem and the sampling problem. To the best of our knowledge, stands as the first comprehensive exploration rooted in the Logistic distribution of the Bellman error. Our contributions can be summarized in four key aspects:
\begin{itemize}[leftmargin=*]
\item We define the discrepancy between $Q$ function estimations and true solutions, revealing a characteristic Logistic distribution for the Bellman error. This discovery highlights that, under practical conditions, the Bellman error adheres to a biased Logistic distribution. This challenges the conventional assumption of Normally distributed Bellman error and distinguishes our work from \cite{bas2021logistic}, which focuses on facilitating update rules with the Logistic function. In contrast, our research assumes a Logistic distribution for the Bellman error and establishes a comprehensive theoretical framework supported by rigorous proofs and analysis.
\item We establish a natural connection between the upper limit of reward scaling and the Logistic distribution, offering theoretical proof of this upper limit's existence.
\item We rigorously derive and numerically simulate the sampling error from the Logistic distribution with the Bias-Variance decomposition. This theoretical foundation supports determining the ideal batch size for training neural networks, thereby reducing unnecessary resource consumption.
\item We validate the propositions by eight distinct online environments and nine offline environments. Additionally, for clarity and completeness, we verify the validity of new theorems with toy examples. The results consistently highlight a robust preference for the Logistic distribution of the Bellman error.
\end{itemize}
We believe our theory will be paramount for future research concerning the distribution of Bellman errors.

The structure of this paper is as follows: Section~\ref{sec:2} presents the important fundamental definitions and basic algorithms in the realm of RL. Section~\ref{sec:3}-\ref{sec:LQL} is our main contribution. Section~\ref{sec:3} conducts the Bellman error analysis on Logistic distribution under Gumbel and Normal initialization. Section~\ref{sec:RewardScaling} analyses the natural connection between Logistic distribution and the reward scaling problem. Section~\ref{sec:Sampling} gives the method for sampling from the Logistic distribution. Section~\ref{sec:LQL} provides an alternative formulation for MSELoss. Section~\ref{Section4:Experiment} conducts the numerical experimental analysis and ablation study for our method. Finally, we summarize this paper in Section~\ref{sec:conclusion} and discuss directions for future work.

\section{Preliminaries}
\label{sec:2}
This section provides a concise introduction to foundational concepts in RL. Section~\ref{sec:2.1} presents basic definitions and notations of RL. Section~\ref{sec:2.2} defines the target of RL, following a detailed derivation in \ref{sec:2.3} for Q learning. Section~\ref{sec:2.4} outlines the fundamental definitions and key properties of the Logistic and Gumbel distributions.

\subsection{Concepts and Notations}
\label{sec:2.1}
RL explores the expected cumulative reward by a Markov decision process defined by a tuple $(\gS,\gA,\sP,r,\gamma)$, where $\gS$ and $\gA$ respectively denote the state and action space. $\sP(\bm{s'}|\bm{s,a})$ is the state transition probability from state $\bm{s}$ that drives toward the next state $\bm{s'}$. Here $r$ defines the reward of taking an action $\va$ at the current state $\vs$. For arbitrary state $\vs \in \gS $, the reward obtained by performing arbitrary action $\va \in \gA $ is defined as $r(\vs,\va)$. $\gamma\in(0,1)$ is the discount factor on future rewards. Online and offline RL differs mainly from the distinct behaviors of interaction between an environment and the agent. In online training, the agent is able to interact with the environment and receive immediate feedback. The agent is expected to learn progressively from these feedbacks to facilitate the optimal strategies. In contrast, the interaction between an agent and the environment is unavailable in offline training, in which case the agent will be advised to learn from a large offline dataset to recognize intrinsic patterns that are expected to be generalized to similar environments. While enhancing the generalizability is a more difficult problem in general, the performance of offline RL is significantly inferior to their online counterparts.

\subsection{Objectives in Reinforcement Learning}
\label{sec:2.2}
The target of RL, as defined by the \emph{Actor-Critic} (AC) algorithm \citep{konda1999actor}, is to find the optimal policy $\pi(\va|\vs)$ that maximizes the cumulative discounted reward at a fixed horizon $T$, \ie the finite-horizon discounted objective:
\begin{equation}
\label{Eq1}
    \E_{\va_{t} \sim      \pi(\va_{t}|\vs_{t})}\left[\sum_{t=0}^T\gamma^t r(\vs_{t},\va_{t})\right].
\end{equation}
Alternatively, \emph{Soft AC} (SAC) \citep{haarnoja2018soft2,haarnoja2018soft} encompasses soft conditions in future rewards to learn the policy $\pi$ with a regularization strength $\zeta$ that maximizes:
\begin{equation}
\label{Eq2}
  \E_{\va_t \sim \pi(\va_t|\vs_t)}\left[\sum_{t=0}^T\gamma^t (r(\vs_t,\va_t)-\zeta \log(\pi(\va_t|\vs_t)))\right].
\end{equation}
A more general version proposed in \cite{garg2023extreme} takes the Kullback-Leibler divergence (KL divergence) between the policy $\pi$ and the prior distribution of a reference distribution $\mu$ to augment the reward function in the objective:
\begin{equation}
\label{Eq3}
    \E_{\va_t \sim \pi(\va_t|\vs_t)}\left[\sum_{t=0}^T\gamma^t (r(\vs_t,\va_t)-\zeta \log\frac{\pi(\va_t|\vs_t)}{\mu(\va_t|\vs_t)})\right].
\end{equation}
The reference distribution $\mu(\va|\vs)$ follows different sampling conventions in different types of RL to fit the behavioral policy \citep{neu2017unified}. Specifically, in online RL, it is usually sampled from a uniform distribution, while in offline RL, it is usually sampled from the empirical distribution of the offline training data.

\subsection{(Soft) Bellman Equation}
\label{sec:2.3}
The cumulative discounted reward can be used to formulate the optimal \textit{Bellman iterative equation} for $Q$-learning \citep{watkins1992q}. For conciseness, we derive the equation from (\ref{Eq3}). The same method can be directly applied to the other two variants in (\ref{Eq1}) and (\ref{Eq2}). All these objectives rely on the Bellman equation, \ie
\begin{equation}
\begin{aligned}
\label{Eq4}
  Q^{t+1}(\vs,\va)&=r(\vs,\va)+\gamma \max_{\va'}(Q^{t}(\vs',\va')).
\end{aligned}
\end{equation}
The analysis in this research is based on (\ref{Eq4}). However, for completeness, we also introduce other update methods.

Consider the general form of the optimal Bellman iterative equation from (\ref{Eq3}), which reads:
\begin{equation}
\label{Eq5}
    \hspace{-3mm}
    Q^{k+1}(\vs,\va)\leftarrow \argmin_{Q}\biggl[r(\vs,\va)+\E_{\vs' \sim \sP(\cdot|\vs,\va),\va'\sim \pi}\left[Q(\vs',\va')-\zeta\log\frac{\pi(\va'|\vs')}{\mu(\va'|\vs')}\right]-Q^k(\vs,\va)\biggr]^2.
\end{equation}
The correspondence solution to the Bellman iteration with respect to (\ref{Eq3}) is then:
\begin{equation}
\label{Eq6}
Q^{k+1}(\vs,\va)=r(\vs,\va)+\E_{\vs'\sim \sP(\cdot|\vs,\va),\va'\sim \pi}\biggl[Q^{k}(\vs',\va')-\zeta \log\frac{\pi(\va'|\vs')}{\mu(\va'|\vs')}\biggr].
\end{equation}
To take the optimal strategy with the maximum $Q^{t}(\vs',\va')$ in (\ref{Eq4}), the corresponding $\pi^{*}$ has to satisfy:
\begin{equation}
\label{Eq7}
  \pi^{*}(\va'|\vs')=\argmax_{\pi}(\E_{\vs'\sim \sP(\cdot|\vs,\va),\va'\sim \pi}\left[Q(\vs',\va')-\zeta \log\frac{\pi(\va'|\vs')}{\mu(\va'|\vs')}\right]),
\end{equation}
where $\sum_{\va'}\pi^{*}(\va'|\vs')=1$. Applying the Lagrange multiplier method \citep{bertsekas2014constrained} (see \ref{Appendix4}), we have:
\begin{equation}
\label{Eq8}
    \pi^{*}(\va|\vs)=\frac{\mu(\va|\vs)e^{Q(\vs,\va)/\zeta}}{\sum_\va \mu(\va|\vs)e^{Q(\vs,\va)/\zeta}}.
\end{equation}
Consequently, simplifying (\ref{Eq7}) by (\ref{Eq8}) yields:
\begin{equation}
\label{Eq9}
    \hspace{-2mm}
    \E_{\vs'\sim \sP(\cdot|\vs,\va),\va'\sim \pi}\left[Q(\vs',\va')-\zeta \log\frac{\pi(\va'|\vs')}{\mu(\va'|\vs')}\right]
    \rightarrow \E_{\vs' \sim \sP(\cdot|\vs,\va)}\biggl[\zeta \log\sum_{\va'} \mu(\va'|\vs')e^{Q(\vs',\va')/\zeta}\biggr].
\end{equation} 
The $\max_{\va'}Q(\vs',\va')$ in (\ref{Eq4}) with respect to the optimal policy $\pi^*$ is then
\begin{equation}
\label{Eq10}
    \max_{\va'}Q(\vs',\va')=\E_{\vs'\sim \gP(\cdot|\vs,\va)}[\zeta \log\sum_{\va'} \mu(\va'|\vs')e^{Q(\vs',\va')/\zeta}].
\end{equation}
While it is challenging to estimate the log sum in (\ref{Eq10}), \cite{garg2023extreme} employed a Gumbel regression-based approach to circumvent the need for sampling estimation.

\subsection{Gumbel and Logistic Distribution}
\label{sec:2.4}
Before delving into the subsequent theoretical analysis, we introduce the probability density function (PDF), cumulative distribution function (CDF), and expectation of the Gumbel distribution and the Logistic distribution.

\begin{definition}[Gumbel Distribution]
If a random variable $x$ follows a $\rm{Gumbel}$ distribution, \ie $x\sim \rm Gumbel(\lambda,\eta)$ with the location parameter $\lambda$ and positive scale parameter $\eta$, then its expectation is $\lambda+\eta v$, where $v\simeq 0.58$ is the Euler–Mascheroni constant. The corresponding PDF and CDF are:
\begin{align*}
    {\rm PDF}&: p(x)=\frac{1}{\eta}\exp\left(-\left(\frac{x-\lambda}{\eta}+\exp({-\frac{x-\lambda}{\eta}})\right)\right),\\
    {\rm CDF}&: P(x)=\exp\left(-\exp(-\frac{x-\lambda}{\eta})\right).
\end{align*}
\end{definition}

\begin{definition}[Logistic Distribution]
\label{def:logistic}
If a random variable $x$ follows a $\rm{Logistic}$ distribution , \ie $x\sim \rm Logistic(\lambda,\eta)$ with the location parameter $\lambda$ and positive scale parameter $\eta$, then its expectation is $\lambda$. The corresponding PDF and CDF are:
\begin{align*}
    {\rm PDF}: p(x)=\frac{1}{\eta}\frac{\exp\left({-(x-\lambda)}/{\eta}\right)}{\left(1+\exp\left({-(x-\lambda)}/{\eta}\right)\right)^2},\quad
    {\rm CDF}: P(x)=\frac{1}{1+\exp\left(-(x-\lambda)/{\eta}\right)}.
\end{align*}
\end{definition}

\section{Characterization of Bellman Error with Logistic Distribution}
\label{sec:3}
This section undertakes a profound analysis of the Bellman error. We initiate our exploration in Section~\ref{sec:3.1} by defining the Bellman error with parameterization $\theta$ and analyzing the distribution of the Bellman error under Gumbel initialization. Due to the lack of universality in Gumbel initialization, we present in Section~\ref{sec:3.2} the formulation of the Normal approximation for the Gumbel distribution. This approximation allows for the substitution of Gumbel initialization in Section~\ref{sec:3.1} with Normal initialization.

\subsection{Gumbel Initialization Approximation for Logistic Bellman Error}
\label{sec:3.1}
To establish a foundational basis for subsequent Q-learning algorithms, it is crucial to highlight that (\ref{Eq4}) serves as the keystone equation. Consequently, our investigation is conducted based on it. We do not analyze the soft update (\ref{Eq6}) in this paper.

We now delve into the exact updating process. As per (\ref{Eq4}), providing initial values is imperative for initiating the iteration. Thus, we designate the $t$-th iteration value associated with the pair $(\vs,\va)$ as $\hat{Q}^t(\vs,\va)$ and take $\hat{Q}^0(\vs,\va)$ with $t=0$ as the start. Denote $Q^*(\vs,\va)$ as the optimal solution for (\ref{Eq4}), it should satisfy the optimal Bellman equation:
\begin{align}
    {Q}^*(\vs,\va)&=r(\vs,\va)+\gamma \max_{\va'}({Q}^*(\vs',\va')). \notag
\end{align}
The $t$-th iteration $\hat{Q}^t(\vs,\va)$ naturally exhibits bias to $Q^*(\vs,\va)$. Define the error between them as $\epsilon^t(\vs,\va)$, then
\begin{equation}
\label{Eq12}
    \hat{Q}^{t}(\vs,\va)={Q}^*(\vs,\va)+\epsilon^{t}(\vs,\va).
\end{equation}

Thus, iterative solutions can be obtained by (\ref{Eq4}) upon randomly initialized $\hat{Q}^0(\vs,\va)$. We will show in Lemma~\ref{Lemma3} that the random variable $\epsilon^{t}(\vs,\va)$ follows a biased Gumbel distribution under some mild assumptions.

While (\ref{Eq4}) is capable of updating tabular Q-values, complex environments often employ neural networks to parameterize the Q-function. We thus parameterize $\hat{Q}$ and $\epsilon$  by $\theta$. We refine the Q-function as $\hat{Q}_\theta(\vs,\va)$ for the $(\vs,\va)$ pair and represent the gap as $\epsilon^{\theta}(\vs,\va)$, which revise 
in (\ref{Eq12}) to:

\begin{align}
    \hat{Q}_\theta(\vs,\va)={Q}^*(\vs,\va)+\epsilon^{\theta}(\vs,\va).
\end{align}

We now define the parameterized Bellman error $\varepsilon^{\theta}$, \ie
\begin{equation}
\label{Eq11}
\varepsilon^{\theta}(\vs,\va)=\biggl[r(\vs,\va)+\gamma \max_{\va'} \hat{Q}_{\theta}(\vs',\va')\biggr]-\hat{Q}_{\theta}(\vs,\va).
\end{equation}
Notably, this parameterization focuses solely on the gap generated by (\ref{Eq4}). We omit potential errors from other parameterization aspects, such as optimizer, gradient updating methods, or network architecture, to avoid excessive complexity. As supported by empirical evidence in Section~\ref{sec:LQL}, neglecting these additional errors would not affect the validity of our theory.

Below we present Lemmas~\ref{Lemma1}-\ref{Lemma5} associated with $\epsilon^t(\vs,\va)$ and $\varepsilon^\theta(\vs,\va)$. The complete proofs are in \ref{Appendix_extract}-\ref{Appendix_Lemma6}. In particular, we will show in Lemma~\ref{Lemma3} that the distribution of $\epsilon^t(\vs,\va)$ is indeed biased and time-dependent, which contradicts the assumption in \citep{garg2023extreme} that $\E[\epsilon^t(\vs,\va)]=0$. We commence our analysis under these four assumptions:
\begin{assumption}
\label{assumption1}
The action space $\gA$ contains a finite number of $n$ elements, \ie $|\gA|=n$.
\end{assumption}
We will see the reasons for Assumption~\ref{assumption1} in the upcoming Lemma~\ref{Lemma2}. The reason for this assumption is an infinite action space does not necessarily guarantee the effectiveness of the max operator.
\begin{assumption}
\label{assumption2}
There is an injection mapping $\gT: (\vs,\va)\rightarrow \vs'$, such that the next state $\vs'$ is uniquely determined by the current state $\vs$ and action $\va$ \footnote{This assumption aligns with practical scenarios, especially when disregarding state transition probabilities \citep{littman1993optimization,mendez2019multi}.}
\end{assumption}
Assumption~\ref{assumption2} is for the convenience of our theoretical analysis, as the state transition probability $\sP(\vs'|\vs,\va)$ can affect the analysis of the error distribution that is not conducive to our analysis.

\begin{assumption}
\label{assumption3}
The initial $\hat{Q}^0(\vs,\va)$ follows some $\rm{Gumbel}$ distribution that is independent from $(\vs,\va)$ pairs.
\end{assumption}
We will show in Lemma~\ref{Lemma1} and \ref{Lemma2} that the direct way to obtain a true Gumbel distribution and continue its type during iterations is to assume the initial state is a Gumbel distribution. Lemma~\ref{Lemma1} shows that under finite conditions, it is not possible to obtain a true Gumbel distribution using other distributions.
\begin{assumption}
\label{assumption4}
$1-\gamma=\kappa<\kappa_0\leq\delta_0$, where $\delta_0>0$ is close to $0$.
\end{assumption}
We will see in Theorem~\ref{Theorem1} that Assumption~\ref{assumption4} is necessary, as it ensures the correctness of the Logistic distribution.

Assumption~\ref{assumption1}, \ref{assumption2} and \ref{assumption4}, aside from Assumption~\ref{assumption3}, are standard. Therefore, such assumptions are reasonable. For Assumption~\ref{assumption3}, although the Gumbel initialization does not align with our practical understanding, in Section~\ref{sec:3.2}, we will provide a method to replace the Gumbel initialization with the Normal initialization and give the standard \textbf{$Assumption 3^*$}. Hence, Assumption~\ref{assumption3} can also be considered as a standard assumption.

Based on the assumptions above, we have the associated lemmas.

\begin{lemma} \citep{fisher1928limiting}
\label{Lemma1} 
    For \textit{i.i.d.} random variables $X_1,\dots, X_n  \sim f(X)$, where $f(X)$ has the exponential tails, let $M_n=\max(\{X_1,\dots, X_n\})$. If there exists two constants $a_n$, $b_n$ with respect to size $n$, where $a_n>0$, satisfying:
    $$
    \lim_{n\to\infty} P\left(\frac{M_n-b_n}{a_n} \leq x\right)= G(x),
    $$
    then $ G(x)$ is the CDF of the standard $\rm{Gumbel}$ distribution, \ie $G(x)=e^{-e^{-x}}$.
\end{lemma}

The key idea of Lemma~\ref{Lemma1} is that obtaining the true Gumbel distribution using the maximum operator under a finite sample size $n$ is generally impossible. However, we can approximate the Gumbel distribution under Normal conditions. We will delve into this discussion in Section~\ref{sec:3.2}. The following Lemma~\ref{Lemma_extract}-\ref{Lemma2} describe the basic properties of the Gumbel distribution.
\begin{lemma}
\label{Lemma_extract}
    If a random variable $X\sim{\rm Gumbel}(A, B)$ follows ${\rm Gumbel}$ distribution with location $A$ and scale $B$, then $X+C\sim{\rm Gumbel}(C+A,B)$ and $DX\sim{\rm Gumbel}(DA,DB)$ with arbitrary constants $C\in\R$ and $D>0$.
\end{lemma}

\begin{lemma}
\label{Lemma2}
For a set of mutually independent random variables $X_i \sim {\rm Gumbel}(C_i, \beta)$ ($1\leq i\leq n$), where $C_i$ is a constant related to $X_i$ and $\beta$ is a positive constant, then $\max_i(X_i) \sim {\rm Gumbel}({\beta}\ln\sum_{i=1}^n e^{\frac{1}{\beta} C_i}, \beta)$.
\end{lemma}

The key idea of Lemma~\ref{Lemma_extract} indicates that the Gumbel distribution maintains its distributional type under linear transformations. Lemma~\ref{Lemma2} demonstrates that a sequence of independent Gumbel distributions scaled with the same constant maintains their distributional type with the maximum operation.

It is worth noting that, in conjunction with the Assumption~\ref{assumption1} and \ref{assumption3}, Lemma~\ref{Lemma2} shows us that
$\gamma \max_{\va'}(\hat{Q}^{0}(\vs',\va')) \sim {\rm Gumbel} (C_1,\beta_1)$ with constants $C_1\in\R, \beta_1>0$ that are determined by the initialization and are independent from $(\vs,\va)$ pair. Based on this analysis, we next propose Lemma~\ref{Lemma3} to establish the relationship between ${Q}^*$ and $\hat{Q}^t$ for $\epsilon^t(\vs,\va)$ defined in (\ref{Eq12}).
\begin{lemma}
\label{Lemma3}
For $\epsilon^t(\vs,\va)$ defined in (\ref{Eq12}), under Assumptions \ref{assumption1}-\ref{assumption3}, we show that:
\begin{equation*}
    \epsilon^{t}(\vs,\va) \sim {\rm Gumbel}(C_t(\vs,\va)-\gamma \max_{\va'}({Q}^*(\vs',\va')), \beta_t),
\end{equation*}
where 
\begin{equation*}
\begin{aligned}
    C_1(\vs,\va)&=C_1,\\
    C_2(\vs,\va)&=\gamma (C_{1}(\vs,\va)+ \beta_{1} ln \sum_{i=1}^n e^\frac{r(\vs',\va_i)}{\beta_{1}}),
\end{aligned}
\end{equation*}
and
\begin{equation*}
    C_t(\vs,\va)=\gamma (\beta_{t-1} ln \sum_{i=1}^n e^\frac{r(\vs',\va_i)+C_{t-1}(\vs',\va_i)}{\beta_{t-1}}).
\end{equation*}
for $t\geq3$. For $\beta_t$, it always holds that
\begin{equation*}
    \beta_t=\gamma^{t-1}\beta_1
\end{equation*}
for $t\geq1$. Besides, $\epsilon^t(\vs,\va)$ are independent for arbitrary pairs $(\vs,\va)$.
\end{lemma}

\begin{remark}
\label{remark1}
There is a special case when the $\rm {Gumbel}$ distribution follows a simpler expression. For $\forall \vs_1, \vs_2$, define two sets $S_1=[r(\vs_1,\va_1),r(\vs_1,\va_2),..., r(\vs_1,\va_n)]$ and
$S_2=[r(\vs_2,\va_1),r(\vs_2,\va_2),..., r(\vs_2,\va_n)]$. If $S_1 \triangle S_2 = \emptyset$, then 
$$
\epsilon^{t}(\vs,\va) \sim {\rm Gumbel}(C_t-\gamma \max_{\va'}({Q}^*(\vs',\va')), \beta_t).
$$
with 
\begin{equation*}
\begin{aligned}
    C_t&=\gamma(C_{t-1}+\beta_{t-1} ln \sum_{i=1}^n e^\frac{r(\vs',\va_i)}{\beta_{t-1}}) (t \geq 2),\\
    \beta_t&=\gamma^{t-1}\beta_1 (t \geq 1).
\end{aligned}
\end{equation*}
\end{remark}

The key idea of Lemma~\ref{Lemma3} indicates that under certain assumptions, $\epsilon^t(\vs,\va)$ follows a Gumbel distribution, where the location parameter is associated with the $(\vs,\va)$ pair, and the scale parameter is time-dependent. This suggests that the independent unbiased assumption of $\epsilon^t(\vs,\va)$ in \cite{garg2023extreme} is not adequately considered. Before delving into the new theorem for Bellman error, we need Lemmas~\ref{Lemma4}-\ref{Lemma5}.
\begin{lemma}
\label{Lemma4}
For random variables $X \sim {\rm Gumbel}(C_X,\beta)$ and $Y \sim {\rm Gumbel}(C_Y,\beta)$, if $X$ and $Y$ are independent, then $(X-Y) \sim {\rm Logistic}(C_X-C_Y,\beta)$.
\end{lemma}

The key idea of Lemma~\ref{Lemma4} shows that subtracting two Gumbel distributions with the same scale parameter results in a Logistic distribution. We will see later that it plays a crucial role in the proof of Theorem~\ref{Theorem1}. It is important to note that $X+Y$ will no longer follow the Logistic distribution. However, it can be approximated by Generalized Integer Gamma (GIG) or Generalized Near-Integer Gamma (GNIG) distributions \citep{marques2015distribution}.

\begin{lemma}
\label{Lemma5}
If $X\sim{\rm Gumbel}(A,1)$, then both $\E[e^{-X}]$ and $\E[Xe^{-X}]$ are bounded:
$$
\E[e^{-X}]<(\frac{20}{e^2} +10e^{-e^{\frac{1}{2}}}+ \frac{1}{2}-\frac{1}{2e})e^{-A}.
$$
When $A>0$:
$$
\E[Xe^{-X}]<(\frac{3}{20}+A(\frac{20}{e^2} +10e^{-e^{\frac{1}{2}}}+ \frac{1}{2}-\frac{1}{2e}))e^{-A}.
$$
When $A\leq0$:
$$
\E[Xe^{-X}]<(\frac{3}{20})e^{-A}.
$$
\end{lemma}

Lemma~\ref{Lemma5} provides the bounds for $\E[e^{-X}]$ and $\E[X e^{-X}]$, these bound are also prepared for Theorem~\ref{Theorem1}. Note that the bounds presented in Lemma~\ref{Lemma5} are upper bounds and do not represent the supremum. 

Next, We will present Theorem~\ref{Theorem1} that defines the Logistic distribution of the Bellman error $\varepsilon^{\theta}(\vs,\va)$ (formulated in (\ref{Eq11})). We parameterize $C_t$ and $\beta_t$ in Lemma~\ref{Lemma3} as $C_\theta$ and $\beta_\theta$, respectively. 
Theorem~\ref{Theorem1} is formulated as follows:

\begin{theorem}[Logistic distribution of Bellman error]
\label{Theorem1}
The Bellman error $\varepsilon^{\theta}(\vs,\va)$ approximately follows the $\rm{Logistic}$ distribution under the Assumptions~\ref{assumption1}-\ref{assumption4}. The degree of approximation can be measured by the upper bound of KL divergence between:
$$X \sim {\rm Gumbel}(\beta_{\theta} {\rm ln}\sum_{i=1}^n e^{\frac{r(\vs',\va_i)+C_{\theta}(\vs',\va_i)}{\beta_{\theta}}},\beta_{\theta}).
$$ 
and
$$
Y\sim { \rm Gumbel}(\gamma \beta_{\theta} {\rm ln}\sum_{i=1}^n e^{\frac{r(\vs',\va_i)+C_{\theta}(\vs',\va_i)}{\beta_{\theta}}},\gamma \beta_{\theta}).
$$
Let $A^*={\rm ln}\sum_{i=1}^n e^{\frac{r(\vs',\va_i)+C_{\theta}(\vs',\va_i)}{\beta_{\theta}}}$, we have these conclusions:
\begin{enumerate}[leftmargin=*]
    \item If $A^*>0$, then ${\rm KL}(Y||X)<\log(\frac{1}{\gamma})+(1-\gamma)[A^*(\frac{20}{e^2} +10e^{-e^{\frac{1}{2}}}- \frac{1}{2}-\frac{1}{2e})+\frac{3}{20}-v]$.
    \item If $A^* \leq 0$, then ${\rm KL}(Y||X)<\log(\frac{1}{\gamma})+(1-\gamma)[\frac{3}{20}-A^*-v]$.
    \item The order of the KL divergence error is controlled at $O(\log(\frac{1}{1-\kappa_0})+\kappa_0A^*)$.
\end{enumerate}
If the upper bound of KL divergence is sufficiently small. Then ${\rm \varepsilon}^{\theta}(\vs,\va)$ follows the $\rm{Logistic}$ distribution, \ie
\begin{equation*}
    {\rm \varepsilon}^{\theta}(\vs,\va) \sim {\rm Logistic} (C_{\theta}(\vs,\va)-\beta_{\theta} {\rm ln}\sum_{i=1}^n e^{\frac{r(\vs',\va_i)+C_{\theta}(\vs',\va_i)}{\beta_{\theta}}},\beta_{\theta}).
\end{equation*}
\end{theorem}

\begin{remark}
\label{remark2}
For the special case discussed in Remark~\ref{remark1}, ${\rm \varepsilon}^{\theta}(\vs,\va)$ satisfies:
\begin{equation*}
    {\rm \varepsilon}^{\theta}(\vs,\va) \sim {\rm Logistic} (-\beta_{\theta} ln\sum_{i=1}^n e^{\frac{r(\vs',\va_i)}{\beta_{\theta}}},\beta_{\theta}).
\end{equation*}
\end{remark}

The proof of Theorem~\ref{Theorem1} can be found in \ref{Appendix3}.

\begin{figure*}[t]
    \centering
    \includegraphics[width=\textwidth]{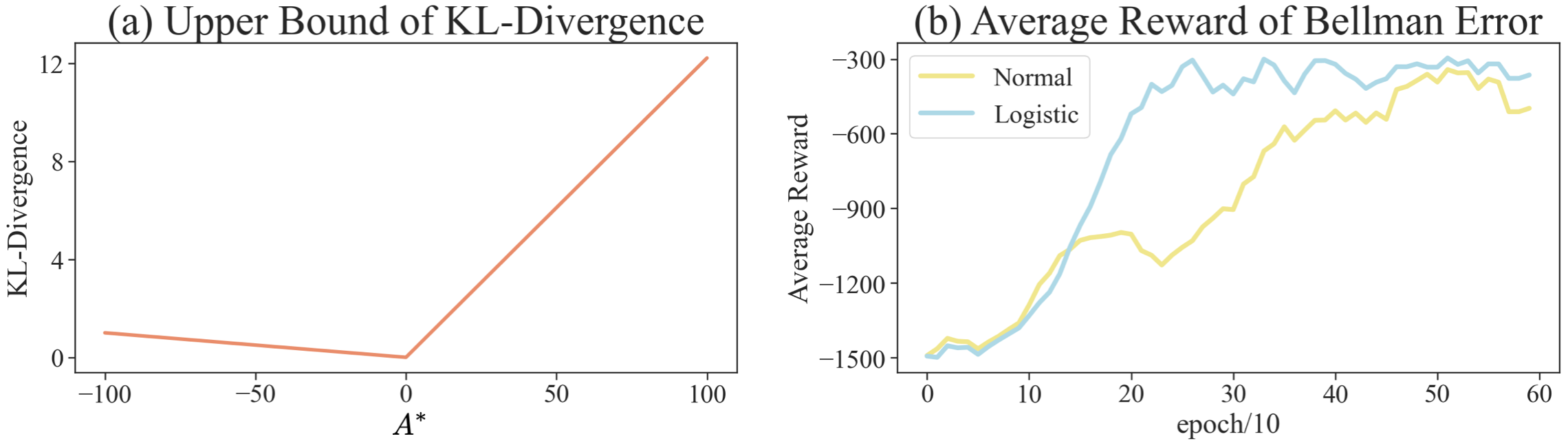}
    \caption{Demonsrtation with \textbf{Pendulum-v1} training. (a) The upper bound of KL divergence with respect to different $A^*$ ratios following Theorem~\ref{Theorem1}. (b) The average reward by Implicit-Q-Learning, with the Bellman error modeled by Normal and Logistic distributions, respectively.}
    \label{fig:evidence}
\end{figure*}

The key idea of Theorem~\ref{Theorem1} indicates that within a certain range of KL divergence, the Bellman error can be effectively modeled by a Logistic distribution. Given that the discount factor $\gamma$ is typically set to a value close to 1 (\eg $0.99$), as long as $A^*$ is not excessively large, the KL divergence range will be sufficiently small. For instance, in Figure~\ref{fig:evidence} (a), the upper bounds of KL divergence for different $A^*$ in \textbf{Pendulum-v1} demonstrates that the small $A^*$ generally correspond to small KL divergence values. It is important to note that this observation holds only for small $A^*$. When $A^*$ becomes considerably large, the upper bound of KL divergence would increase to an inefficient stage (\eg $A^*=100$ results in $\rm{KL}(Y||X)<13$ in Figure~\ref{fig:evidence} (a) ), where the impact of the discount factor becomes significant. Hence, employing the Logistic approximation proves advantageous over a Normal approximation for a more effective training trajectory. This advantage is particularly noticeable during the early stages of training. As depicted in Figure~\ref{fig:evidence} (b),  formulating the Bellman error with a Logistic distribution expedites more effective training for the average reward. This phenomenon correlates with the large $\beta_\theta$ value associated with the substantial gaps in different rewards during the early stages, leading to a remarkably small range of KL divergence. 

We delve further into the discussion in Section~\ref{Section4:Experiment}.

When optimizing the Bellman error, its expectation is anticipated to converge toward zero, \ie
\begin{equation}
\label{Eq13}
  \E\biggl[{\rm \varepsilon}^{\theta}(\vs,\va)\biggr] \to 0.
\end{equation}
However, this expectation contradicts Theorem~\ref{Theorem1}, where the distribution of $\varepsilon^{\theta}(\vs,\va)$ is proven to be biased with an increasing kurtosis over time (For instance, see Figure~\ref{fig:timeline}). While establishing a direct approximation for the Bellman error is challenging due to its expectation being related to the $(\vs,\va)$ pair, we adopt (\ref{Eq13}) as an alternative objective for regression. More details will be provided in Section~\ref{sec:LQL}.

Another notable analysis is related to the Assumption~
\ref{assumption3}. Directly initializing neural networks to the Gumbel distribution is generally impractical, whereas the Normal distribution is more commonly used for initialization. To address this, we next discuss the finite approximation for the Gumbel distribution under Normal distribution initialization.

\subsection{Normal Initialization Approximation for Logistic Bellman Error}
\label{sec:3.2}
The theoretical validation of Theorem~\ref{Theorem1} and Lemma~\ref{Lemma3} in Section~\ref{sec:3.1} were based on Assumption~\ref{assumption3} (see \ref{Appendix2} and \ref{Appendix3} ). While this assumption contradicts the prevalent practice in neural networks, which commonly employ Normal initialization, our objective is to find an approximation that allows us to represent the Gumbel distribution using the Normal distribution. To this end, this section investigates the Normal approximation for Bellman errors following a Gumbel distribution.

To commence, we define the exponential family of distributions and relevant notations. For a random variable $X$ following an exponential family distribution with parameter $\nu$, its PDF is given by
\begin{equation*}
    f(X)=\frac{\nu}{2\Gamma(\frac{1}{\nu})}\sqrt{\left(\frac{\Gamma(\frac{3}{\nu})}{\Gamma(\frac{1}{\nu})}\right)} e^{-(\frac{\Gamma(\frac{3}{\nu})}{\Gamma(\frac{1}{\nu})})^{\frac{2}{v}} |X|^{\nu}}.
\end{equation*}
In particular, define:
\begin{equation*}
\begin{aligned}
    \theta^{\nu}&=\nu-1, \quad C^{\nu}=(\frac{\Gamma(\frac{3}{\nu})}{\Gamma(\frac{1}{\nu})})^{\frac{\nu}{2}}, \quad D_0^{\nu}=\frac{(C^{\nu})^{\frac{1-\nu}{\nu}}}{2\Gamma(\frac{1}{\nu})},\\
    \beta_N^\nu&=\frac{\theta^{\nu}}{\nu (C^{\nu})} W_0[\frac{\nu (C^{\nu})}{\theta^{\nu}}(D_0^{\nu}N)^{\frac{\nu}{\theta^{\nu}}}]^{\frac{1}{\nu}},\\
    D_1^\nu&=-(1-\frac{1}{\nu})\frac{1}{C^\nu}, \quad D_2^\nu=(1-\frac{1}{\nu})(2-\frac{1}{\nu})\frac{1}{(C^\nu)^2}.
\end{aligned}
\end{equation*}
where $W_0[\cdot]$ is the real part of the Lambert W-function, and $\Gamma(\cdot)$ denotes the Gamma function. For $\Gamma(n+\frac{1}{2})$, it is defined:
\begin{equation*}
    \Gamma(n+\frac{1}{2})=\frac{(2n)!\sqrt{\pi}}{n!4^n}.
\end{equation*}
Notably, once $\nu$ is given, these parameters can be computed directly. Since we are interested in the normal distribution, we take $\nu=2$ to reach the case that $f(X)$ follows the standard Normal distribution, actually:
$$
f(X)=\frac{1}{\sqrt{2 \pi}} e^{-\frac{X^2}{2}}.
$$
Based on the definitions above, we now present a numerical method in Lemma~\ref{lemma6} that approximates a Gumbel distribution by the Normal distribution.

\begin{lemma} \citep{zarfaty2021accurately}
\label{lemma6}
Suppose {$X_1$, $X_2$, $X_3$, ..., $X_N$} are \textit{i.i.d} variables from a Normal distribution ${\rm Normal}(0,1)$. Define $f_N(X)$ as the PDF of $X=\max_i(X_i)$ and $g(X)$ as the PDF of a standard ${\rm Gumbel}(0,1)$, then $f_N(X)\simeq \frac{1}{a_N} g(\frac{(X-b_N)}{a_N})$, where
\begin{equation*}
\begin{aligned}
    a_N \simeq &\frac{1}{2C^{\nu}\beta_N^{\nu}}\left[1-\frac{\theta^{\nu}}{2C^{\nu}{(\beta_N^{\nu})^2}}+\frac{(\theta^{\nu})^2-6C^{\nu}D_1^{\nu}}{4(C^{\nu})^2{(\beta_N^{\nu})}^4} \right.\\
    &\qquad \qquad \left.-\frac{2(\theta^{\nu})^3-32\theta^{\nu} C^{\nu}D_1^{\nu}-20(C^{\nu})^2((D_1^{\nu})^2-2D_2^{\nu})}{16(C^{\nu})^3 ({\beta_N}^{\nu})^6}\right],\\
    b_N \simeq &\beta_N^{\nu} \left[1+\frac{D_1^{\nu}}{2C^{\nu}{(\beta_N^{\nu})}^4}-\frac{2\theta^{\nu} D_1^{\nu}+2C^{\nu}(({D_1^{\nu})}^2-2D_2^{\nu})}{16(C^{\nu})^3{(\beta_N^{\nu})}^6}\right].
\end{aligned}
\end{equation*}
\end{lemma}

Lemma~\ref{lemma6} shows us how 
to approximate the Gumbel distribution with a finite set of samples from the Normal distribution. It allows us to relax Assumption~\ref{assumption3} to \textbf{Assumption $3^*$} for Normal initialization. 

\vspace{2mm}
\noindent \textbf{Assumption $3^*$.}
The initial $\hat{Q}^0(\vs,\va)$ follows some \textbf{standard Normal distribution} that is independent from $(\vs,\va)$ pairs.
\vspace{2mm}

In this way, Assumption~\ref{assumption3} becomes standard. Then with the previous Assumptions~\ref{assumption1}, \ref{assumption2}, and \ref{assumption4}, we can extend Lemma~\ref{Lemma3} and Theorem~\ref{Theorem1} to Normal initialization. 
To establish an intuitive understanding of the revised Theorem~\ref{Theorem1} and Lemma~\ref{Lemma3} under Normal initialization, we now present a toy example with a finite state space.

\begin{example}
\label{example1}
    Consider a scenario with five states $\{\vs_i\}_{i=1}^{5}$ and an action space of $5000$ actions $\{\va_j\}_{j=1}^{5000}$. At each state $\vs_i$, taking any action $\va_j$ results in transition from $\vs_i$ to $\vs_{i+1}$ with a constant reward $r(\vs_i,\va_j) \equiv 1$.
\end{example}

Example~\ref{example1} presents a finite state space that can be stored in $\mQ\in\R^{5\times 5000}$. Then the Bellman equation can be uniquely optimized by
\begin{equation*}
    {Q^*}(s_i,:)=\sum_{k=0}^{5-i} 0.99^k,
\end{equation*}
where $\mQ(s_i,:)$ denotes the $i$-th row of $\mQ$ with respect to the $i$-th state.

\begin{figure*}[t]
    \centering
    \includegraphics[width=\textwidth]{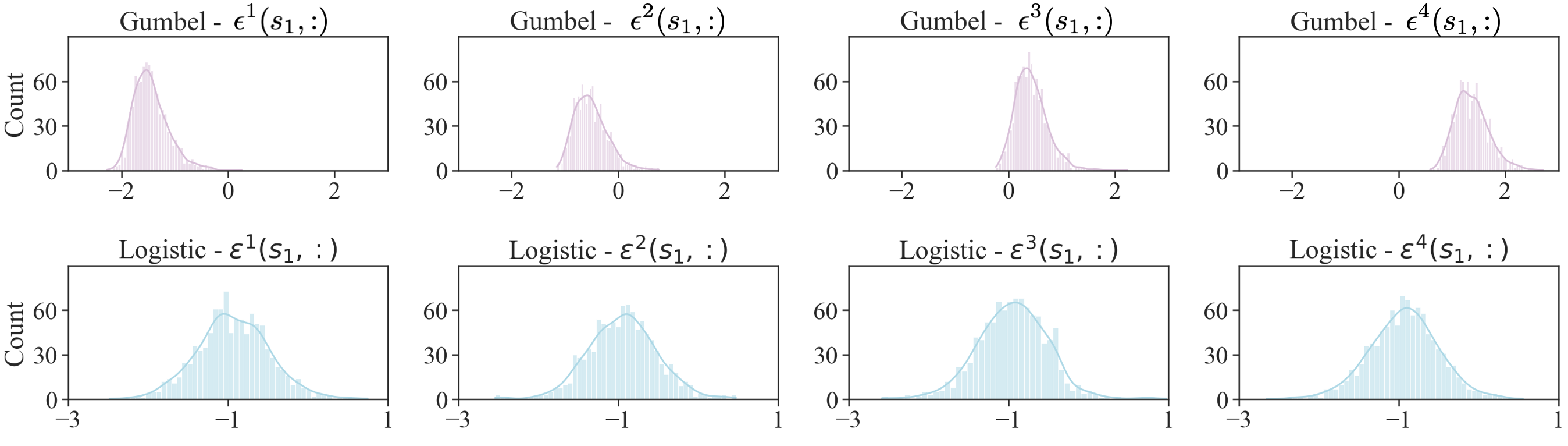}
    \caption{The distribution of $\epsilon(\vs,\va)$ (row 1, in purple) and $\varepsilon(\vs,\va)$ (row 2, in blue) in the first four iterations with a randomly initialized $Q$ table. The former roughly follow Gumbel distributions, and the latter follow Logistic distributions.}
    \label{fig:example}
\end{figure*}

Figure~\ref{fig:example} verifies Lemma~\ref{Lemma3} and Theorem~\ref{Theorem1} with this toy example by visualizing the first $4$ iterations of the Bellman errors. We use ${\rm Normal}(0,1)$ random initialization for each element in the Q table and employ (\ref{Eq4}) for iterating. Define
\begin{equation*}
\begin{aligned}
    &\epsilon^{t}(\vs_1,:)=\hat{Q}^t(\vs_1,:)-{Q}^*(\vs_1,:),\\
    \text{and }\; &\varepsilon^{t}(\vs_1,:)=(r(\vs_1,:)+\max_{\va'}(\hat{Q}^t(\vs_2,:)))-\hat{Q}^t(\vs_1,:),
\end{aligned}
\end{equation*}
where $\epsilon^{t}(\vs_1,:)$ (row 1) follows a Gumbel distribution, and $\varepsilon^{t}(\vs_1,:)$ (row 2) follows a Logistic distribution. While this toy example makes many simplifications for illustrative purposes only, we will provide further validation in Section~\ref{Section4:Experiment} on complex real-world environments, in which cases obtaining the optimal $Q^*$ values is generally impossible. 

\section{The Proportional Reward Scaling Phenomenon and Bellman Error}
\label{sec:RewardScaling}
The \textit{proportional reward scaling} \citep{haarnoja2018soft} problem describes a saturation phenomenon preventing the reward scaling factor from expanding infinitely in RL. Gao \textit{et al.} \citep{gao2023scaling} explained this problem by policy gradient. This section demonstrates a rational explanation for the reward scaling problem from the standpoint of the distribution of Bellman errors. We will establish a natural connection between reward scaling and the expectation of the Logistic distribution.

We start by connecting the proportional reward scaling problem and the Bellman error. Following Remark~\ref{remark2}, we facilitate the analysis with the special case of Theorem~\ref{Theorem1}, where
\begin{equation*}
    {\rm \varepsilon}^{\theta}(\vs,\va) \sim {\rm Logistic} (-\beta_{\theta} {\rm ln}\sum_{i=1}^n e^{\frac{r(\vs',\va_i)}{\beta_{\theta}}},\beta_{\theta}).
\end{equation*}
We have shown in Section~\ref{sec:3} that this distribution is biased, which poses a challenge for approximation with a neural network. Alternatively, we seek to transform it to support a nearly unbiased approximation. We explore whether we can alleviate the bias in the Logistic distribution through reward scaling. Specifically, we present the following theorem to guide scaling.

\begin{theorem}[Positive Scaling upper bounds under Remark~\ref{remark2}]
\label{theorem1-extract}
Denote $r^{+}$ and $r^{-}$ the positive and negative rewards with $r>0$ and $r<0$, respectively. With $i_1+i_2+i_3=n$, assume that:
\begin{equation*}
    \sum_{i=1}^n e^{\frac{r(\vs',\va_i)}{\beta_{\theta}}}=\sum_{i=1}^{i_1} e^{\frac{r^{+}(\vs',\va_i)}{\beta_{\theta}}}+\sum_{i=1}^{i_2} e^{\frac{r^{-}(\vs',\va_i)}{\beta_{\theta}}}+i_3.
\end{equation*}

If it satisfies
\begin{enumerate}[leftmargin=*]
    \item $i_1 \neq 0$,
    \item $\sum_{i=1}^{i_1} e^{\frac{r^{+}(\vs',\va_i)}{\beta_{\theta}}}r^{+}(\vs',\va_i)+\sum_{i=1}^{i_2} e^{\frac{r^{-}(\vs',\va_i)}{\beta_{\theta}}}r^{-}(\vs',\va_i)<0$,
\end{enumerate}
then there exists an optimal scaling ratio $\varphi^*>1$, such that for any scaling ratio $\varphi$ that can effectively reduce the expectation of the Bellman error, it must satisfy
$$
1 \leq \varphi \leq \varphi^*.
$$
\end{theorem}

The proof of Theorem~\ref{theorem1-extract} can be found in \ref{AppendixTheorem2}.

Theorem~\ref{theorem1-extract} offers valuable insights into reward scaling bounds and associated phenomena. Specifically, under the conditions outlined in Theorem~\ref{theorem1-extract}, appropriate scaling can rectify the biased expectation of the Logistic distribution toward zero, thereby enhancing training performance significantly. However, the existence of the upper bounds prohibits excessive scaling to revert the corrected expectation back to its previous state or worse state. We provide an example to further demonstrate this scenario.

\begin{figure*}[t]
    \centering
    \includegraphics[width=\textwidth]{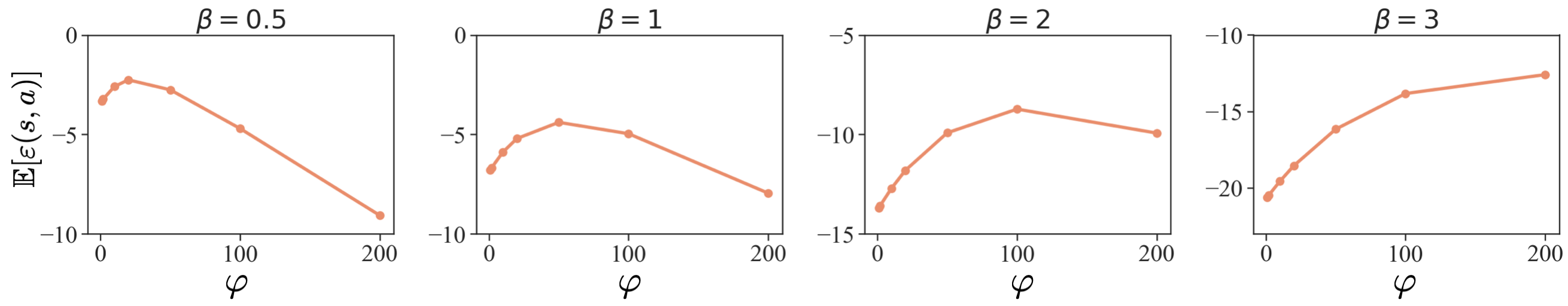}
    \caption{The change of $\E[\varepsilon^{\theta}(\vs,\va)]$ by assigning different $\beta$s. An optimal scaling ratio $\varphi^*$ exists in all the scenarios.}
    \label{fig:example2}
\end{figure*}

\begin{example}
    Consider different scaling factors $\varphi$ in training the \textbf{BipedalWalker-v3} environment.     Figure~\ref{fig:example2} explores the expectation of the sampling error ${\rm \varepsilon}^{\theta}(\vs,\va)$. Let $\{\va_j\}_{j=1}^{5000}$ provide a sufficiently large action space for sampling. We find that there are always positive rewards in these 5000 samples, which satisfies the first condition in Theorem~\ref{theorem1-extract} that $i_1 \neq 0$. Meanwhile, for $\beta_\theta=(0.5,1,2,3)$, $\sum_{i=1}^{i_1} e^{\frac{r^{+}(\vs',\va_i)}{\beta_{\theta}}}r^{+}(\vs',\va_i)+\sum_{i=1}^{i_2} e^{\frac{r^{-}(\vs',\va_i)}{\beta_{\theta}}}r^{-}(\vs',\va_i)\approx(-88,-105,-109,-117)$. In other words, the second condition in Theorem~\ref{theorem1-extract} holds for all the $\beta$s under investigation. Furthermore, the optimal $\varphi^*$ suggested in Theorem~\ref{theorem1-extract} can be observed from the figure. In each of the 4 scenarios in Figure~\ref{fig:example2}, the $\E[\varepsilon^{\theta}(\vs,\va)]$ reaches its upper bound halfway through increasing the scaling factor $\varphi$. Based on the empirical observation, we conclude that when the error variance is considerably small, a scaling ratio of $10 \sim 50$ is recommended. Note that this observation is consistent with the experimental results in \citep{haarnoja2018soft}.
\end{example}

Example 2 explains the existence of the upper bound on the scaling ratio. It comprehends from a distributional perspective for enhancing the model performance during training.

\section{Sampling strategy of Bellman Error with Logistic Distribution}
\label{sec:Sampling}
This section delves into the considerations of batch size in neural networks, building upon the theorem established in Section~\ref{sec:3} that the Bellman error conforms to a Logistic distribution. As the direct application of tabular Q-learning proves inadequate for complex environments, the extension of established theorems, such as training a neural network, becomes crucial for gaining practical significance. In this context, we explore the empirical choice of the batch size $N$ used in sampling Bellman errors for parameter updates. The goal is to regulate the error bound while maintaining computational efficiency. We employ the Bias-Variance decomposition to analyze the sampling distribution and substantiate the identification of a suitable  $N^*$.

Firstly, we outline the problem we are addressing in this section. 

\begin{problem}
    Assume that we have a Logistic distribution denoted as $\rm{Logistic}(A, B)$. We aim to draw points from this distribution to represent it. The more sampling points we have, the more representative they are, and vice versa. The representativeness is measured using the sampling error $S_e$ in (\ref{Eq14}). Our objective is to determine an appropriate batch size $N^*$ that fits the Logistic distribution with $S_e \approx 1\times 10^{-6}$.
\end{problem}
To address this issue, we first define the empirical distribution function for sampling. For $\{x_1,x_2,...,x_N\}$ sampled from $\rm{Logistic}(A, B)$, the associated empirical distribution function for this sequence is 
\begin{equation*}
    \hat{F}^{(x_1,x_2,...,x_N)}_N(t)=\frac{1}{N}\sum_{i=1}^N 1_{x_i\leq t}.
\end{equation*}
Following Definition~\ref{def:logistic}, we denote $F(t), f(t)$ as the CDF and PDF of the $\rm{Logistic}(A, B)$ ($A$ replaces $\lambda$, and $B$ replaces $\eta$). The sampling error $S_e$ reads
\begin{equation}
\label{Eq14}
S_e=\E_t\E_{(x_1,x_2,...,x_N)}\left[(\hat{F}^{(x_1,x_2,...,x_N)}_N(t)-F(t))^2\right].
\end{equation}

We next define the Bias-Variance decomposition for (\ref{Eq14}) in Lemma~\ref{lemma8}.
\begin{lemma}
\label{lemma8}
\label{lemma_SE}
The sampling error $S_e$ in (\ref{Eq14}) can be decomposed into Bias and Variance terms. If we define:
$$
\overline{F}(t)=\E_{(x_1,x_2,...,x_N)}\left[\hat{F}^{(x_1,x_2,...,x_N)}_N(t)\right],
$$
then
\begin{equation*}
S_e=\E_t\left[{\rm Variance}(t)+{\rm Bias}(t)\right]={\rm Variance}+{\rm Bias},
\end{equation*}
where
\begin{equation*}
\begin{aligned}
{\rm Variance}(t)&=\E_{(x_1,x_2,...,x_N)}\left[(\hat{F}^{(x_1,x_2,...,x_N)}_N(t))^2\right]-\E_{(x_1,x_2,...,x_N)}^2[\hat{F}^{(x_1,x_2,...,x_N)}_N(t)],\\
{\rm Bias}(t)&=(\overline{F}(t)-F(t))^2.
\end{aligned}
\end{equation*}
\end{lemma}

\begin{figure*}[t]
    \centering
    \includegraphics[width=\textwidth]{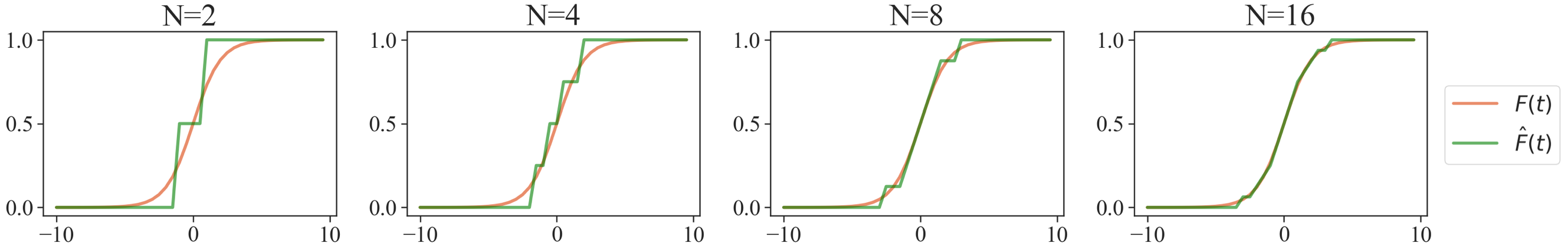}
    \caption{The differences between $F(t)$ (true CDF of $\rm{Logistic}(0,1)$) and $\overline{F}(t)$ (empirical CDF) with varying sample sizes $N$.}
    \label{fig:sampling}
\end{figure*}

We provide a detailed derivation of Lemma~\ref{lemma_SE} in 
\ref{Appendix_Lemma8}.

The key to estimating the sampling error $S_e$ lies in calculating $\overline{F}(t)$. For $N=1$, we show that:
\begin{equation*}
    \overline{F}(t)=\E_{x_1}[\hat{F}^{x_1}_N(t)]=\E_{x_1}[1_{x_1\leq t}]=1_{\E[x_1]\leq t}=1_{A \leq t}.
\end{equation*}
For $N\geq 2$, considering order statistics becomes essential, since
\begin{equation*}
    \overline{F}(t)=\E_{(x_1,x_2,...,x_N)}[\hat{F}^{(x_1,x_2,...,x_N)}_N(t)]=\E_{(x_1,x_2,...,x_N)}[\frac{1}{N}\sum_{i=1}^N 1_{x_i\leq t}]=\frac{1}{N}\sum_{i=1}^N 1_{\E[x_{(i)}]\leq t}.
\end{equation*}
Here $x_{(i)}$ denotes the $i$-th order statistics. To find $\E[x_{(i)}]$ for each $x_{(i)}$, we perform piecewise segmentation \citep{gentle2010computational} on the CDF of each $x_{(i)}$, which reads
\begin{equation*}
    f^{x_{(i)}}(t)=\frac{N!}{(i-1)!(N-i)!}(F(t))^{i-1}(1-F(t))^{n-i}f(t).
\end{equation*}

We then have the following estimation theorem:
\begin{theorem}[The Expectation of order statistics for the Logistic distribution]
\label{theorem_SE}
\begin{equation*}
    \E[x_{(i)}]=\left[B[\sum_{k=1}^{i-1} \frac{1}{k}-\sum_{k=1}^{N-i} \frac{1}{k}]+A\right].
\end{equation*}
\end{theorem} 

The proof of Theorem~\ref{theorem_SE} can be found in \ref{Appendix_Theorem5}.

With Theorem~\ref{theorem_SE}, we can directly calculate each $\E[x_{(i)}]$ analytically. Figure~\ref{fig:sampling} compares $\overline{F}(t)$ with $F(t)$ under $\rm{Logistic}(0,1)$ with sample points $N=(2, 4, 8, 16)$, with $\overline{F}(t)$ calculating by following Theorem~\ref{theorem_SE}. We show that increasing $N$ leads to a more accurate estimation.

Note that the empirical distribution is a type of step function, \ie
\begin{equation*}
\begin{aligned}
    \E_{(x_1,x_2,...,x_N)}[(\hat{F}^{(x_1,x_2,...,x_N)}_N(t))^2]
    &= \E_{(x_1,x_2,...,x_N)}[(\frac{1}{N}\sum_{i=1}^N 1_{x_i\leq t})^2] \\
    &= \sum_{i=1}^{N-1} (\frac{i}{N})^2 1_{(\E[x_{(i)}]\leq t \leq \E[x_{(i+1)}])}+1_{t>\E[x_{(N)}]}, \\
    \E_{(x_1,x_2,...,x_N)}^2[\hat{F}^{(x_1,x_2,...,x_N)}_N(t)]
    &= (\sum_{i=1}^N (\frac{i}{N}) 1_{\E[x_{(i)}]\leq t})^2 \\
    &= \E_{(x_1,x_2,...,x_N)}[(\hat{F}^{(x_1,x_2,...,x_N)}_N(t))^2].
\end{aligned}
\end{equation*}
Consequently, the Variance of this empirical distribution is zero, leading to the estimation of the bias term as the sole concern in approaching $S_e$. Suppose $t$ is uniformly sampled from the range $\left[\E[x_{(1)}], \E[x_{(N)}]\right]$, then
\begin{equation}
\label{Eqse}
    S_e=\E_t[{\rm Bias}(t)]=\frac{1}{\E[x_{(N)}]-\E[x_{(1)}]}\left[\sum_{i=1}^{N-1} \int_{\E[x_{(i)}]}^{\E[x_{(i+1)}]}(F(t)-\frac{i}{N})^2 dt\right],
\end{equation}
where $\E[x_{(i)}]$ follows the definition in Theorem~\ref{theorem_SE} with a fixed size $N$. We can then obtain the upper and lower limits of the integral in (\ref{Eqse}), leading to a direct computation of $\E[x_{(i)}]$ in $S_e$ \footnote{We employ symbolic integration with the built-in `\texttt{int}' function in MATLAB. Instead of using numerical integration techniques, we leverage indefinite integral to achieve a direct numerical result.}. Table~\ref{Table_SE} reports the associated $S_e$ with respect to varying $N$s, where the sampling error keeps reducing with an increased sample size $N$. Notably, the errors are related to $N$ and are independent of both parameters $A$ and $B$. Moreover, a moderate batch size of $256$ is sufficient for achieving a small $S_e \approx 1\times10^{-6}$.

\begin{table}[!t]
\caption{The relationship between the sample size $N$ and sampling error $S_e$ using (\ref{Eqse}).}
\label{Table_SE}
\centering
\resizebox{\linewidth}{!}
{
\begin{tabular}{lrrrrrrrr}
    \toprule
    & \textbf{2} & \textbf{4} &\textbf{8} & \textbf{16} & \textbf{32} & \textbf{64} & \textbf{128} & \textbf{256} \\
    \midrule
    $S_e$ & $2\times 10^{-2}$ & $4\times 10^{-3}$ & $1\times 10^{-3}$ &  $3\times 10^{-4}$ & $8\times 10^{-5}$ & $2\times 10^{-5}$ & $5\times 10^{-6}$ & $1\times 10^{-6}$ \\
    \bottomrule
\end{tabular}
}
\end{table}

The results verify that assuming a Logistic distribution for the Bellman error facilitates the determination of an appropriate batch size for training in RL. Rather than being driven by performance metrics, the batch size can be selected based on the precision requirements.

\section{Logistic Likelihood Q-Learning} 
\label{sec:LQL}
We have justified the rationality of modeling the Bellman error with the Logistic distribution instead of the Normal distribution. Based on this, we propose training a neural network with the Logistic maximum likelihood loss function in place of the conventional MSE-based loss function.

As the typical choice in deep RL networks for Q-updating, $\rm MSELoss$ is based on the assumption that the estimation error follows a Normal distribution $\rm{Normal}(0,\sigma)$. $\rm MSELoss$ is derived from the maximum likelihood estimation function:
\begin{equation}
\label{Eqeq}
    \log[\prod_{i=1}^{n}p({\rm \varepsilon}_i)]=-n\log(\sqrt{2\pi}\sigma)-\sum_{i=1}^n \frac{{\rm \varepsilon}^2_i}{2\sigma^2}\\
    \propto \sum_{i=1}^n -\frac{1}{2} ({\rm \varepsilon}_i)^2.
\end{equation}

In Section~\ref{sec:3.1}, we have deduced that the Bellman error should follow a biased Logistic distribution. While estimating the expectation for this distribution is not straightforward for a neural network, we simplified the objective to (\ref{Eq13}). To this end, we assume the Bellman error follows $\rm{Logistic}(0,\sigma)$ and derive the associated likelihood function as a replacement for $\rm MSELoss$.

We start from the PDF of $\varepsilon_i \sim {\rm Logistic}(0,\sigma)$, which reads
\begin{equation}
p({\rm \varepsilon}_i)=\frac{1}{\sigma}\frac{e^{\frac{-{\rm \varepsilon}_i}{\sigma}}}{(1+e^{\frac{-{\rm \varepsilon}_i}{\sigma}})^2}.
\end{equation}
By employing the log-likelihood function, we have
\begin{equation}
  \log\left[\prod_{i=1}^{n}p({\rm \varepsilon}_i)\right]=-n \log(\sigma)+\sum_{i=1}^n \left[-\frac{{\rm \varepsilon}_i}{\sigma}-2\log\left(1+e^{\frac{-{\rm \varepsilon}_i}{\sigma}}\right)\right].
\end{equation}
This returns the Logistic Loss function ($\rm LLoss$), \ie
\begin{equation}
{\rm LLoss}=\frac{1}{N}\sum_{i=1}^N \left[\frac{{\rm \varepsilon}_i}{\sigma}+2\log\left(1+e^{\frac{-{\rm \varepsilon}_i}{\sigma}}\right)\right].
\end{equation}

\begin{figure*}[t]
    \centering
    \includegraphics[width=\textwidth]{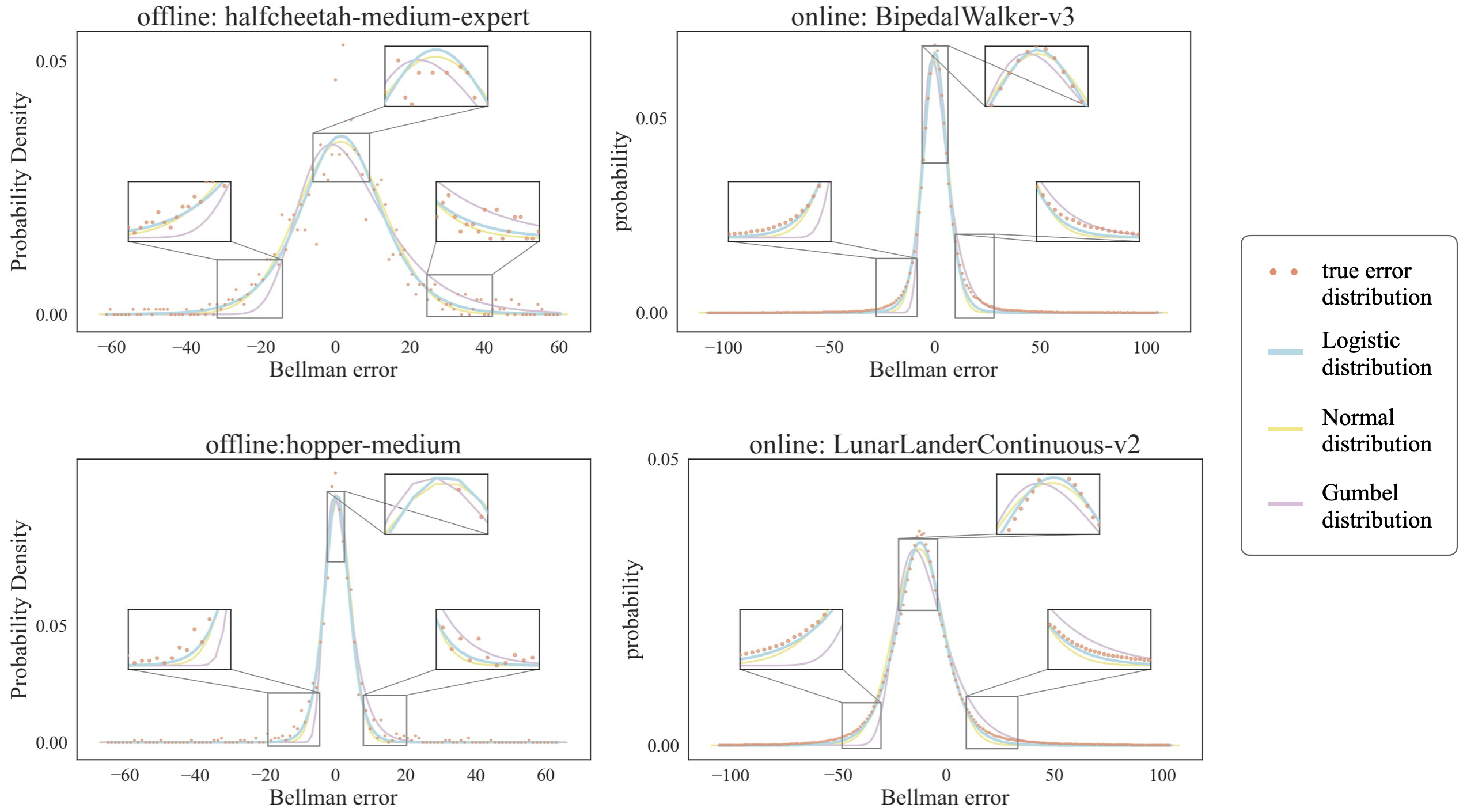}
    \caption{The distribution of Bellman error. For all four online and offline environments, the Bellman errors fit better to the Logistic distribution than the Gumbel and Normal distributions. More details are provided in Table~\ref{Goodness_Online}-\ref{Goodness_Offline}. }
    \label{fig:dist}
\end{figure*}

We have demonstrated in Figure~\ref{fig:timeline} (also see \ref{Appendix11} for additional visualizations) that the distribution of Bellman error evolves along training steps and exhibits a stronger fit to the logistic distribution. Figure~\ref{fig:dist} further compares the closeness of empirical Bellman error to Logistic, Normal, and Gumbel distributions. In all four environments, the Logistic distribution performs better in fitting the empirical Bellman error. In addition to these visualized comparisons, numerical evaluations will be provided later in Table~\ref{Goodness_Online}-\ref{Goodness_Offline} with Kolmogorov–Smirnov (KS) statistic magnitudes \citep{an1933sulla}.

It is noteworthy that $\rm MSELoss$ and $\rm LLoss$ are strongly correlated, with $\rm LLoss$ serving as a corrective function for $\rm MSELoss$. The following Theorem~\ref{theorem5} reveals the relationship between $\rm MSELoss$ and $\rm LLoss$ when $\varepsilon$ is sufficiently small.

\begin{theorem}[Relationship between $\rm LLoss$ and $\rm MSELoss$]
\label{theorem5}
The $\rm MSELoss$ can be used as an approximate estimation of $\rm LLoss$ when $\varepsilon$ is sufficiently small, \ie
\begin{equation*}
    \rm LLoss=ln4+\frac{1}{2} \rm MSELoss+o(\rm \varepsilon^3),
\end{equation*}
where $o(\rm \varepsilon^3)$ third-order infinitesimal of $\varepsilon$ when $\varepsilon$ is sufficiently small.
\end{theorem}

Theorem~\ref{theorem5} demonstrates that the $\rm MSELoss$ can be regarded as an approximate form of $\rm LLoss$ when higher-order terms are neglected. The proof is in \ref{Appendix9}.

\paragraph{\textbf{Summary}}

Through the analysis in Sections~\ref{sec:3}-\ref{sec:LQL}, we concluded that it is more theoretically sound to model the Bellman error with a Logistic distribution in comparison to a Normal or Gumbel distribution. Meanwhile, presenting the Bellman error with a Logistic regression can naturally address the proportional reward scaling problem, which relationship can not be revealed by assuming a Normal distribution. In the next section, we validate the effectiveness of our theory through experiments.

\begin{table}[t]
\caption{Hyperparameters setting for online training. }
\label{Table 5}
\centering
\resizebox{0.8\linewidth}{!}
{
\begin{tabular}{lccccc}
    \toprule
    & \textbf{SAC ($\sigma$)} & \textbf{CQL ($\sigma$)} & \textbf{TAU} 
    & \textbf{Scaling} & \textbf{max. Step} \\
    \midrule
    LunarLanderContinuous-v2  & 10 & 10 & 0.005 & 1 & 200 \\
    HalfCheetah-v2  & 10 & 10 & 0.005 & 1 & 200 \\
    Hopper-v4  & 10 & 10 & 0.005 & 1 & 200  \\
    Walker2d-v2  & 10 & 10 & 0.005 & 1 & 200 \\
    HumanoidStandup-v4  & 90 & 60 & 0.005 & 1 & 100 \\
    InvertedPendulum-v4  & 10 & 10 & 0.005 & 1 & 1000 \\
    InvertedDoublePendulum-v2  & 10 & 10 & 0.005 & 1 & 1000 \\
    BipedalWalker-v3  & 20 & 20 & 0.005 & 50 & 200 \\
    \bottomrule
\end{tabular}
}
\caption{Hyperparameters setting for offline training.}
\label{Table 6}
\centering
\resizebox{0.95\linewidth}{!}
{
\begin{tabular}{lccccc}
    \toprule
    & \textbf{IQL} ($\sigma$) & \textbf{Eval steps} & \textbf{Train steps} &\textbf{Expl. steps} &  \textbf{max. Step} \\
    \midrule
    hopper-medium-v2 & 10 & 100 & 100 & 100 & 100  \\
    walker2d-medium-v2 & 3 & 100 & 100 & 100 & 100  \\
    halfcheetah-medium-v2 & 3 & 100 & 100 & 100 & 100   \\
    hopper-medium-replay-v2 & 3 & 100 & 100 & 100 & 100  \\
    walker2d-medium-replay-v2 & 20 & 100 & 100 & 100 & 100   \\
    halfcheetah-medium-replay-v2 & 20 & 100 & 100 & 100 & 100 \\
    hopper-medium-expert-v2  & 10 & 100 & 100 & 100 & 100  \\
    halfcheetah-medium-expert-v2 & 3 & 100 & 100 & 100 & 100  \\
    walker2d-medium-expert-v2 & 5 & 100 & 100 & 100 & 100  \\
    \bottomrule
\end{tabular}
}
\end{table}

\section{Experiment}
\label{Section4:Experiment}
This section conducts empirical evaluations on wildly assessed online and offline environments for validating the effectiveness of adopting $\rm LLoss$ in practice. Section ~\ref{sec:4.1} elucidates the experimental setups. Section ~\ref{sec:4.2} analysis model performance on $17$ online and offline environments. Section~\ref{sec:4.3} performs additional investigations on our proposed method, including the Kolmogorov-Smirnov (KS) test on the distribution of Bellman error and ablation studies.

\begin{figure*}[t]
    \centering
    \includegraphics[width=\textwidth]{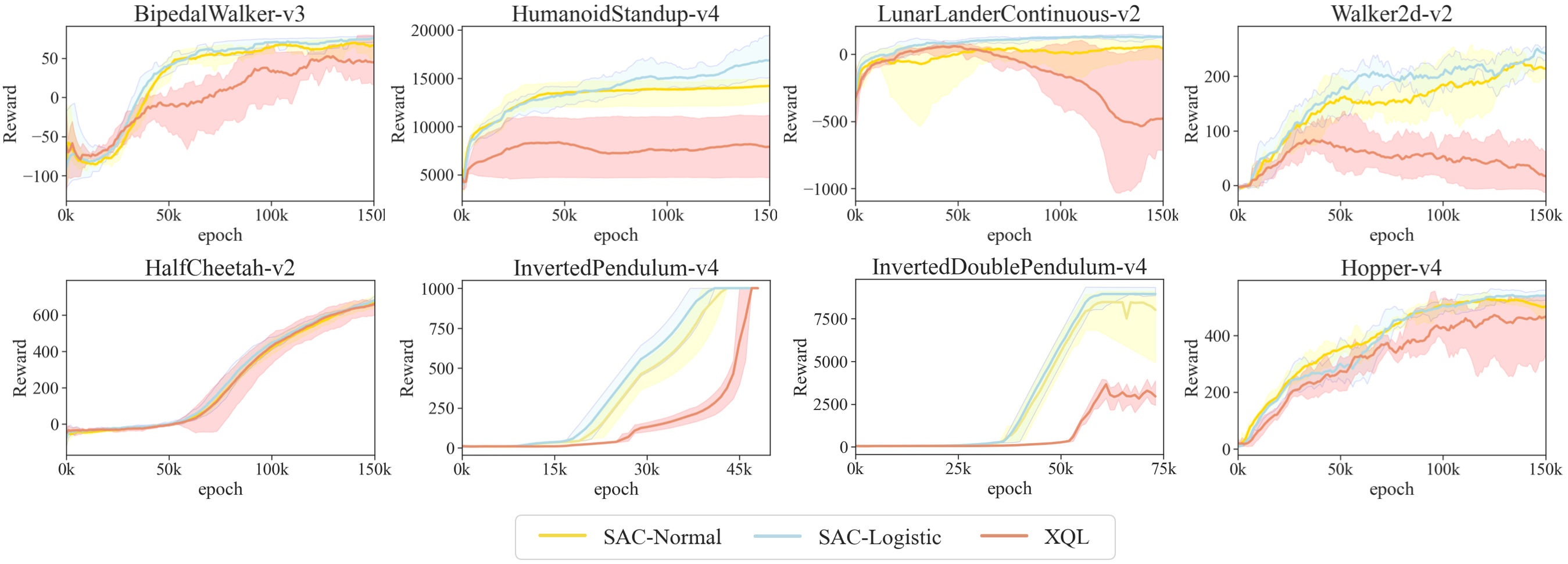}
    \caption{The average reward of SAC, LSAC, and XQL in online training.}
    \label{fig:onlineCurve}
    \centering
    \includegraphics[width=\textwidth]{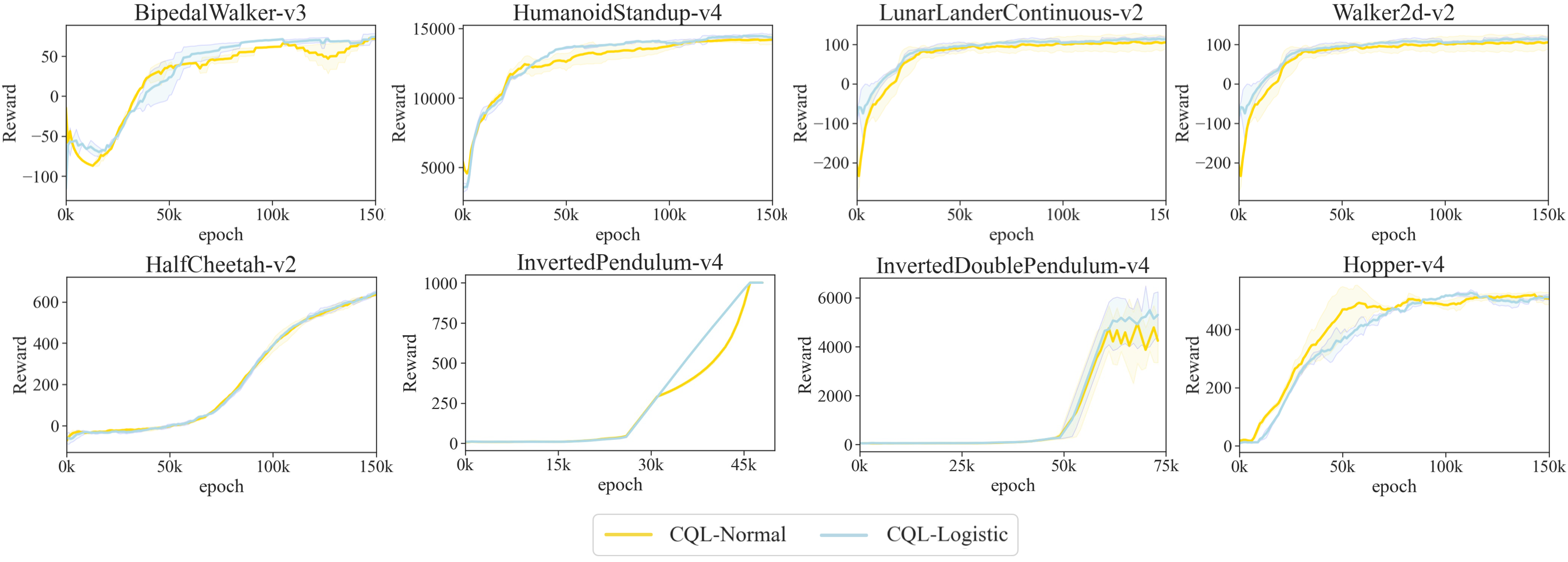}
    \caption{The average reward of CQL, LCQL in online training.}
    \label{fig:onlineCurve2}
\end{figure*}

\subsection{Experiment Protocol}
\label{sec:4.1}
We conducted our experiments using \textbf{gym} (ver.0.23.1), \textbf{mujoco} (ver.2.3.7), and \textbf{D4RL} (ver.1.1). For online RL tasks, training was carried out over $160,000$ iterations across $8$ \textbf{gym} environments. Due to the training simplicity in offline RL, models trained for offline tasks underwent up to $500$ iterations across $9$ \textbf{D4RL} environments \citep{fu2020d4rl}. In both online and offline scenarios, training is stopped after $50$ non-improving steps.

Next, we provide the configurations. Following the analysis in Section~\ref{sec:Sampling}, we set the batch size to $256$ for both online and offline RL. Tables~\ref{Table 5} and \ref{Table 6} report the initializations for online and offline training, respectively. For online RL, we validate the improvement of employing $\rm LLoss$ on SAC \citep{haarnoja2018soft}and CQL \citep{kumar2020conservative}. Consequently, we specify the associated $\sigma$ initialization. For unspecified settings, we adhere to the default setup in \citep{haarnoja2018soft}. Similarly, in Table~\ref{Table 6}, we report the $\sigma$ initialization for IQL \citep{kostrikov2021offline}. The ``Expl. steps`` in Table~\ref{Table 6} represent the agent we need to specify the number of task-agnostic environment steps.
All the programs are sourced from \texttt{rlkit} \footnote{\url{https://github.com/rail-berkeley/rlkit}}.

\subsection{Results Analysis}
\label{sec:4.2}
\paragraph{\textbf{Online RL}} We made improvements based on the official implementation of SAC \footnote{\url{https://github.com/haarnoja/sac}} and CQL \footnote{\url{https://github.com/aviralkumar2907/CQL}}. We replaced $\rm MSELoss$ with $\rm LLoss$ to observe the performance improvement. The enhanced methods with $\rm LLoss$ are referred to as \textbf{LSAC} and \textbf{LCQL}, respectively. During training, we fixed the learning rate to $3\times 10^{-4}$, the discount factor $\gamma$ to $0.99$. 
To guarantee that the performance enhancement is completely attributed to the modification on the loss function, all other initializations are kept identical for $\rm MSELoss$ and $\rm LLoss$ variants. In the comparison with XQL, we fine-tuned XQL based on the $\beta$ range proposed by its authors. The purpose of comparing with XQL is to assess the correctness of Gumbel distribution versus Logistic distribution. The purpose of comparing with SAC is to evaluate the correctness of Normal distribution versus Logistic distribution. In our setting, Most of the environments were run under the condition of 200 max steps. We trained each environment for 160,000 iterations. Based on the rolling epoch timeline, we plotted and stored the Average Reward for each 2000 epoch. It can be observed from Figure~\ref{fig:onlineCurve} that compared to $\rm MSELoss$, $\rm LLoss$ demonstrates more prominent performance in both SAC and CQL, the detailed results are available in Table~\ref{tab:online_main}. The enhancement in Table~\ref{tab:online_main} is calculated as follows:

For the $i$th environment, denote $R^i_{\rm Model}$ and $R^i_{\rm LModel}$ as the average reward obtained by the baseline Model (SAC, CQL) and our variant (LSAC, LCQL), respectively. The online enhancement is defined as follows:
$$
{\rm Enhancement_{online}}(i)=\frac{R^i_{\rm LModel}-R^i_{\rm Model}}{R^i_{\rm Model}}.
$$

\begin{figure*}[t]
    \centering
    \includegraphics[width=\textwidth]{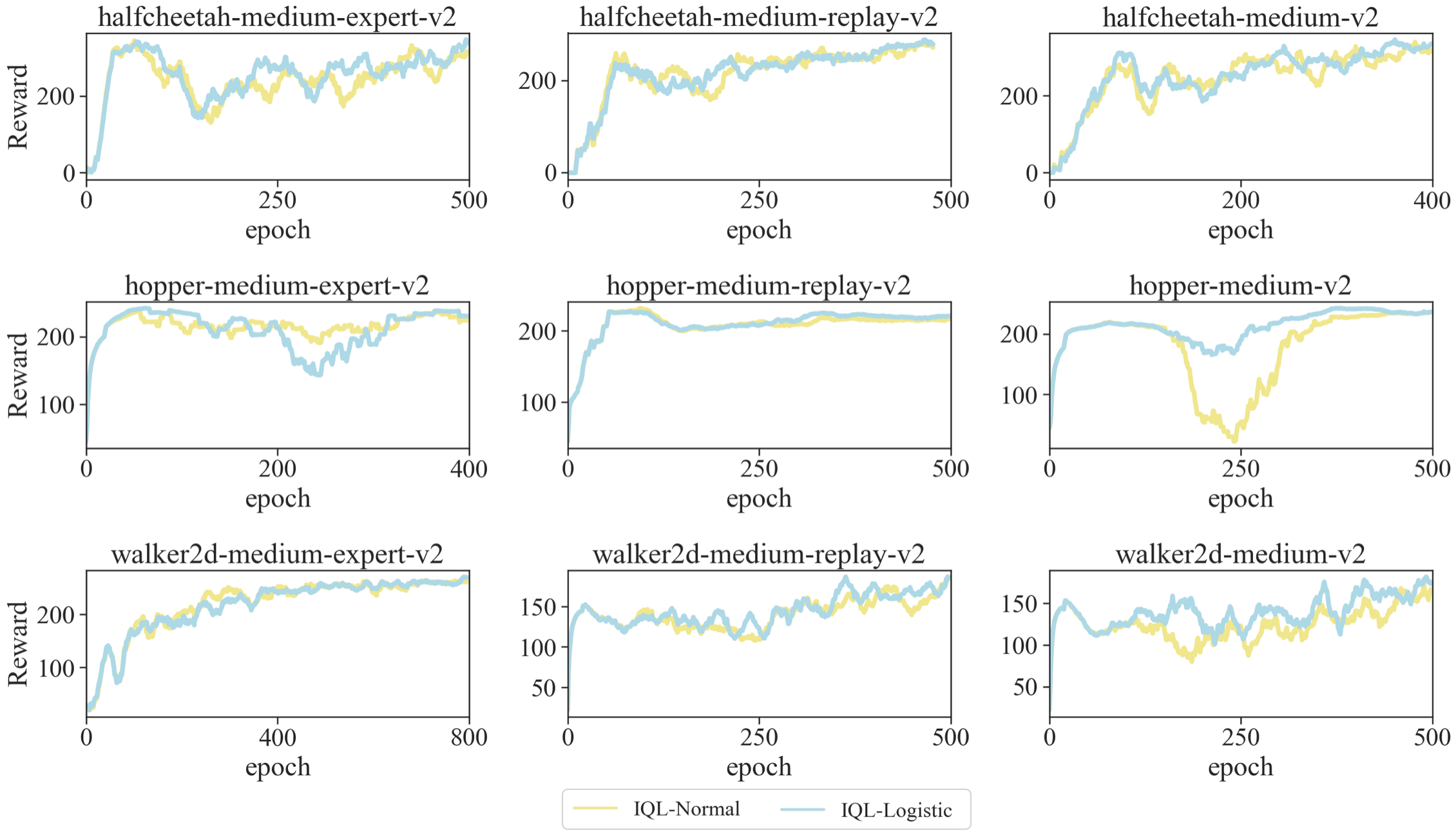}
    \caption{The average reward of IQL and LIQL in offline training.}
    \label{fig:offlineCurve}
\end{figure*}

\paragraph{\textbf{Analysis of Maximum Rewards in Online RL}}
As mentioned in Section~\ref{sec:3.1}, one significant advantage of the Logistic distribution is its ability to expedite training in the early stages, implying a faster training rate compared to the Normal distribution. so we analyze when these environments under online training can reach their maximum reward values and what these maximum reward values are here, as shown in Table~\ref{tab:Analysis_max_Online}. From the results, we can see that LLoss can accelerate the attainment of the maximum reward to some extent and achieve a better maximum reward.

\begin{table}[t]
\caption{The maximum reward of online training over $10$ random repetitions with parentheses reporting the number of epochs (in hundreds) to achieve the results.}

\label{tab:Analysis_max_Online}
\centering
\resizebox{\linewidth}{!}
{
\begin{tabular}{lrrrrr}
    \toprule
    & \textbf{SAC} & \textbf{CQL} & \textbf{XQL} & \textbf{LSAC} (Ours)  & \textbf{LCQL} (Ours) \\
    \midrule
    LunarLander-Continuous-v2 & 194.85 (900) & 154.02 (1350) & 211.90 (340) & \textcolor{red!70!black!}{221.75 (110)} & \textcolor{red!70!black!}{156.72 (220)} \\
    HalfCheetah-v2 & 847.20 (1520) & 739.38 (1500) & 835.24 (1530) & \textcolor{red!70!black!}{856.14 (1300)} & \textcolor{red!70!black!}{761.77 (1420)} \\
    Hopper-v4 & 628.20 (1510) & 616.30 (430) & 618.32 (1200) & \textcolor{red!70!black!}{635.09 (1100)} & 594.98 (\textcolor{red!70!black!}{140}) \\
    Walker2d-v2 & 427.70 (1340) & 360.42 (1570) & 327.14 (340) & \textcolor{red!70!black!}{465.08 (1270)} & \textcolor{red!70!black!}{387.23 (1210)} \\
    HumanoidStandup-v4 & 15142.51 (1590) & 15209.97 (1550) & 13032.94 (1280) & \textcolor{red!70!black!}{23771.01 (760)} & \textcolor{red!70!black!}{15487.68 (1230)} \\
    InvertedPendulum-v4 & 1001.00 (210) & 1001.00 (280) & 1001.00 (270) & 1001.00 (\textcolor{red!70!black!}{190}) & 1001 (280) \\
    InvertedDouble-Pendulum-v2 & 9359.82 (380) & 9361.33 (510) & 9360.56 (540) & \textcolor{red!70!black!}{9362.28} (380) & \textcolor{red!70!black!}{9363.40 (500)} \\
    BipedalWalker-v3 & 79.05 (1490) & 80.11 (1560) & 81.10 (1540) & \textcolor{red!70!black!}{82.53 (1330)} & \textcolor{red!70!black!}{83.77 (1410)}  \\
    \bottomrule
\end{tabular}
}
\end{table}

\begin{table}[th]
\caption{Average reward of online training over $10$ random repetitions. The red values indicate the enhancement of ${\rm LLoss}$ over its $\rm MSELoss$ counterparts.}

\label{tab:online_main}
\centering
\resizebox{\linewidth}{!}
{
\begin{tabular}{lrrrrr}
    \toprule
    & \textbf{SAC} & \textbf{CQL} & \textbf{XQL} & \textbf{LSAC} (Ours) & \textbf{LCQL} (Ours)\\
    \midrule
    LunarLander-Continuous-v2 & 19.99 & 104.15 & -489.19 & 133.95 (\textcolor{red!70!black!}{570.09 \%}) & 112.57 (\textcolor{red!70!black!}{8.08 \%}) \\
    HalfCheetah-v2 & 696.96 & 653.62 & 684.96 & 714.54 (\textcolor{red!70!black!}{2.52 \%}) & 675.33 (\textcolor{red!70!black!}{3.32 \%}) \\
    Hopper-v4 & 509.47 & 495.34 & 487.08 & 544.72 (\textcolor{red!70!black!}{6.92 \%} ) & 515.30 (\textcolor{red!70!black!}{4.03 \%} ) \\
    Walker2d-v2 & 221.46 & 194.59 & 3.42 & 251.63 (\textcolor{red!70!black!}{13.62 \%} ) & 219.27 (\textcolor{red!70!black!}{12.68 \%} ) \\
    HumanoidStandup-v4 & 14,157.95 & 14,166.01 & 8,030.26 & 16,781.59 (\textcolor{red!70!black!}{18.53 \%} ) & 14,258.06 (\textcolor{red!70!black!}{0.65 \%} ) \\
    InvertedPendulum-v4 & 1001 & 1001 & 1001 & 1001 (\textcolor{red!70!black!}{0.00 \%}) & 1001 (\textcolor{red!70!black!}{0.00 \%}) \\
    InvertedDouble-Pendulum-v2 & 8466.48 & 4295.46 & 3290.36 & 8941.93 (\textcolor{red!70!black!}{5.62 \%}) & 4647.64 (\textcolor{red!70!black!}{8.20 \%} ) \\
    BipedalWalker-v3 & 68.69 & 43.91 & 64.56 & 77.59 (\textcolor{red!70!black!}{12.96 \%}) & 71.96 (\textcolor{red!70!black!}{63.88 \%}) \\
    \midrule
    avg. enhancement &  &  &  & \textcolor{red!70!black!}{78.78 \%} & \textcolor{red!70!black!}{12.61 \%} \\
    \bottomrule
\end{tabular}
}
\end{table}

\begin{table}[th]
\caption{The average reward and enhanced ratio after the offline training, all algorithms have calculated the enhancement ratios relative to the IQL. 
}
\label{tab:offline_main}
\centering
\resizebox{\linewidth}{!}
{
\begin{tabular}{llrrrrrrr}
    \toprule
    & & & &  \multicolumn{5}{c}{enhancement over IQL (\%)} \\\cmidrule(lr){5-9} 
    & & \textbf{IQL} & \textbf{LIQL} & \textbf{LIQL} & \textbf{XQL} & \textbf{CQL} & \textbf{TD3+BC} & \textbf{one-step RL} \\ 
    &&&& (ours) \\ \midrule
    \multirow{3}{*}{\rotatebox[origin=c]{90}{medium}} & hopper-v2 & {228.19} & \textbf{240.71} & \textcolor{red!70!black!}{5.49} & \textcolor{red!70!black!}{7.24} & \textcolor{green!70!black!}{-11.77} & \textcolor{green!70!black!}{-10.56} & \textcolor{red!70!black!}{2.11} \\
    & walker2d-v2 & {138.42} & \textbf{161.26} & \textcolor{red!70!black!}{16.50} & \textcolor{red!70!black!}{4.09} & \textcolor{green!70!black!}{-7.41} & \textcolor{red!70!black!}{6.89} & \textcolor{green!70!black!}{-10.11} \\
    & halfcheetah-v2 & {319.41} & \textbf{335.44} & \textcolor{red!70!black!}{5.02} & \textcolor{red!70!black!}{0.63} & \textcolor{green!70!black!}{-7.18} & \textcolor{red!70!black!}{1.89} & \textcolor{red!70!black!}{4.47} \\
    \midrule
    \multirow{3}{*}{\rotatebox[origin=c]{90}{replay}} & hopper-v2 & {243.92} & \textbf{264.30} & \textcolor{red!70!black!}{8.35} & \textcolor{red!70!black!}{1.36} & \textcolor{red!70!black!}{2.94} & \textcolor{red!70!black!}{0.90} & \textcolor{green!70!black!}{-13.80} \\
    & walker2d-v2 & {153.09} & \textbf{176.44} & \textcolor{red!70!black!}{15.25} & \textcolor{red!70!black!}{2.71} & \textcolor{red!70!black!}{4.46} & \textcolor{red!70!black!}{10.69} & \textcolor{red!70!black!}{2.96} \\
    & halfcheetah-v2 & {215.28} & \textbf{221.31} & \textcolor{red!70!black!}{2.80} & \textcolor{red!70!black!}{2.75} & \textcolor{red!70!black!}{0.31} & \textcolor{green!70!black!}{-35.70} & \textcolor{green!70!black!}{-33.02} \\
    \midrule
    \multirow{3}{*}{\rotatebox[origin=c]{90}{expert}} & hopper-v2 & {224.49} & \textbf{237.38} & \textcolor{red!70!black!}{5.74} & \textcolor{red!70!black!}{17.05} & \textcolor{red!70!black!}{15.19} & \textcolor{red!70!black!}{7.10} & \textcolor{red!70!black!}{7.73} \\
    & walker2d-v2 & {261.75} & \textbf{270.15} & \textcolor{red!70!black!}{3.21} & \textcolor{red!70!black!}{0.46} & \textcolor{green!70!black!}{-0.73} & \textcolor{red!70!black!}{0.45} & \textcolor{red!70!black!}{4.61} \\
    & halfcheetah-v2 & {305.14} & \textbf{347.30} & \textcolor{red!70!black!}{13.82} & \textcolor{red!70!black!}{3.58} & \textcolor{red!70!black!}{5.65} & \textcolor{red!70!black!}{4.61} & \textcolor{red!70!black!}{12.90} \\ \midrule
    \multicolumn{2}{l}{avg. enhancement} &  & & \textcolor{red!70!black!}{8.46} & \textcolor{red!70!black!}{4.43} & \textcolor{red!70!black!}{0.16} & \textcolor{green!70!black!}{-1.53} & \textcolor{red!70!black!}{3.10} \\
    \bottomrule
\end{tabular}
}
\end{table}

\paragraph{\textbf{Offline RL}}
We conducted improved experiments based on the IQL components. We set the maximum iteration count as 500 and incorporated a variance threshold of 5 to determine convergence for 50 epochs. The method of controlling variables is the same as in the online setting. Our algorithm is referred to as LIQL. Due to some dimensional discrepancies between the IQL algorithm provided by rlkit and the IQL algorithm, we use the improvement ratio relative to the IQL baseline as the measure of algorithm performance. The change in the average reward during training is depicted in Figure~\ref{fig:offlineCurve}, and relevant details are presented in Table~\ref{tab:offline_main}. The results also indicate that our model exhibits the highest enhancement ratio. The enhancement in Table~\ref{tab:offline_main} is calculated as follows:

If the average reward in environment $i$ obtained by the Model (IQL) is denoted as $\rm{R^i_{IQL}}$, and correspondingly, the average reward obtained by other models (LIQL, XQL, CQL, TD3+BC, one-step RL) are denoted as $\rm{R^i_{Model}}$
, then the offline enhancement in environment $i$ is defined as follows:
$$
\rm{Enhancement_{offline}(i)}=\frac{R^i_{Model}-R^i_{IQL}}{R^i_{IQL}}.
$$

\subsection{KS tests and Ablation Study}
\label{sec:4.3}
\paragraph{\textbf{KS Tests}}
The Kolmogorov-Smirnov (KS) tests, introduced by \citep{an1933sulla}, are employed to examine whether data conforms to a particular distribution. We conducted KS tests on the Bellman error for each environment. It includes statistical $R^2$, and other statistical test parameters. The test results are presented in Table~\ref{Goodness_Online} and Table~\ref{Goodness_Offline}. The KS tests results indicate that our assumption of the Logistic distribution is more accurate than the other two distributions.

\begin{table}[t]
\caption{The fitness and KS tests of Bellman-error for online RL.}
\label{Goodness_Online}
\centering
\resizebox{\linewidth}{!}
{
\begin{tabular}{lcccccccccccc}
    \toprule
    & \multicolumn{3}{c}{R$^2$ $\uparrow$} & \multicolumn{3}{c}{SSE ($\times 10^{-4}$) $\downarrow$} & \multicolumn{3}{c}{RMSE ($\times 10^{-4}$) $\downarrow$} & \multicolumn{3}{c}{KS statistic $\downarrow$} \\\cmidrule(lr){2-4}\cmidrule(lr){5-7}\cmidrule(lr){8-10}\cmidrule(lr){11-13}
    & Logistic & Gumbel & Normal & Logistic & Gumbel & Normal & Logistic & Gumbel & Normal & Logistic & Gumbel & Normal \\
    \midrule
    
    \textbf{LunarLanderContinuous-v2} & \bm{$0.985$} & $0.971$ & $0.975$ & \bm{$1.119$} & $2.224$ & $1.902$ & \bm{$7.555$} & $10.653$ & $9.850$ & \bm{$0.052$} & $0.070$ & $0.071$ \\

    \textbf{HalfCheetah-v2} & \bm{$0.991$} & $0.990$ & $0.989$ &  \bm{$1.344$} & $1.425$ & $1.549$ & \bm{$8.282$} & $8.405$ & $8.888$ & \bm{$0.026$} & $0.047$ & $0.033$ \\
    
    \textbf{Hopper-v4} & \bm{$0.989$} & $0.985$ & $ 0.981$ & \bm{$2.697$} & $3.793$ & $4.807$ & \bm{$11.734$} & $13.912$ & $15.661$ & \bm{$0.067$} & $0.073$ & $0.085$ \\
    
    \textbf{Walker2d-v2} & \bm{$0.988$} & $0.967$ & $0.975$ & \bm{$0.900$} & $2.549$ & $1.903$ & \bm{$6.778$} & $11.404$ & $9.854$ & \bm{$0.054$} & $0.084$ & $0.072$\\

    \textbf{HumanoidStandup-v4} & \bm{$0.667$} & $0.641$ & $0.628$ & \bm{$27.164$} & $29.279$ & $30.318$ & \bm{$37.228$} & $38.652$ & $39.331$ &\bm{$0.269$} &$0.322$ & $0.291$\\
    
    \textbf{InvertedPendulum-v4} & \bm{$0.983$} & $0.963$ & $0.971$ & \bm{$20.961$} & $46.307$ & $35.772$ & \bm{$32.702$} & $48.606$ & $42.721$ &\bm{$0.115$} &$0.175$ &$0.117$ \\

    \textbf{InvertedDoublePendulum-v4} & \bm{$0.999$} & $0.981$ & $0.998$ & \bm{$0.249$} & $5.623$ & $0.324$ & \bm{$3.959$} & $16.938$ & $4.063$ &\bm{$0.021$} & $0.079$ & $0.023$\\
    
    \textbf{BipedalWalker-v3} & \bm{$0.997$} & $0.979$ & $0.990$ & \bm{$1.206$} & $7.888$ & $3.563$ & \bm{$7.843$} & $20.061$ & $13.482$ &\bm{$0.039$} & $0.101$ &$0.057$\\
    \bottomrule
\end{tabular}
}
\end{table}

\begin{table}[t]
\caption{The fitness and KS tests of Bellman-error for offline RL.}
\label{Goodness_Offline}
\centering
\resizebox{\linewidth}{!}
{
\begin{tabular}{lcccccccccccc}
    \toprule
    & \multicolumn{3}{c}{R$^2$ $\uparrow$} & \multicolumn{3}{c}{SSE ($\times 10^{-4}$) $\downarrow$} & \multicolumn{3}{c}{RMSE ($\times 10^{-4}$) $\downarrow$} & \multicolumn{3}{c}{KS statistic $\downarrow$} \\\cmidrule(lr){2-4}\cmidrule(lr){5-7}\cmidrule(lr){8-10}\cmidrule(lr){11-13}
    & Logistic & Gumbel & Normal & Logistic & Gumbel & Normal & Logistic & Gumbel & Normal & Logistic & Gumbel & Normal \\
    \midrule
    
    \textbf{hopper-medium-v2} & \bm{$0.981$} & $0.975$ & $0.976$ & \bm{$10.175$} & $13.975$ & $13.191$ & \bm{$29.617$} & $34.709$ & $33.722$ & \bm{$0.040$} & $0.094$ & $0.053$ \\
    
    \textbf{walker2d-medium-v2} & \bm{$0.927$} & $0.923$ & $0.913$ &  \bm{$15.714$} & $15.915$ & $18.563$ & \bm{$36.801$} & $37.129$ & $40.003$ & \bm{$0.062$} & $0.072$ &$0.076$ \\
    
    \textbf{halfcheetah-medium-v2} & \bm{$0.836$} & $0.831$ & $ 0.833$ & \bm{$9.087$} & $9.394$ & $9.223$ & \bm{$12.314$} & $12.941$ & $12.444$ &\bm{$0.050$} &$0.052$ &$0.069$ \\
    
    \textbf{halfcheetah-medium-replay-v2} & \bm{$0.852$} & $0.813$ & $0.836$ & \bm{$12.864$} & $16.242$ & $14.221$ & \bm{$33.301$} & $37.416$ & $35.012$ & \bm{$0.031$} & $0.105$ & $0.048$\\
    
    \textbf{walker2d-medium-replay-v2} & \bm{$0.950$} & $0.908$ & $0.937$ & \bm{$10.514$} & $19.256$ & $13.262$ & \bm{$30.101$} & $40.743$ & $33.812$ &\bm{$0.075$} &$0.149$ & $0.093$\\
    
    \textbf{hopper-medium-replay-v2} & \bm{$0.954$} & $0.927$ & $0.948$ & \bm{$10.773$} & $17.278$ & $12.265$ & \bm{$30.475$} & $38.594$ & $32.516$ &\bm{$0.038$} &$0.104$ &$0.049$ \\

    \textbf{hopper-medium-expert-v2} & \bm{$0.985$} & $0.970$ & $0.982$ & \bm{$8.054$} & $17.097$ & $10.162$ & \bm{$26.351$} & $38.391$ & $29.598$ &\bm{$0.056$}& $0.105$ & $0.059$\\

    \textbf{walker2d-medium-expert-v2} & \bm{$0.981$} & $0.959$ & $0.973$ & \bm{$11.186$} & $23.807$ & $15.584$ & \bm{$31.053$} & $45.302$ & $36.653$ &\bm{$0.067$}& $0.138$ & $0.075$\\

    \textbf{halfcheetah-medium-expert-v2} & \bm{$0.919$} & $0.869$ & $0.913$ & \bm{$12.722$} & $20.568$ & $13.719$ & \bm{$33.117$} & $42.108$ & $34.391$ &\bm{$0.036$}& $0.098$ &$0.045$\\

    \bottomrule
\end{tabular}
}
\end{table}

\paragraph{\textbf{Sensitivity Analysis}}
We conducted some sensitivity analysis on the variation of $\sigma$ across different environments in both online and offline settings.
We conducted proportional $\sigma$ variations for each environment to observe the changes in the final average reward and maximum average reward.
We put these details in \ref{Appendix15}. The results indicate that the performance of our approach within a certain range of $\sigma$ variations outperforms the $\rm MSELoss$ and exhibits a certain level of robustness.

\section{Conclusion and Future Direction}
\label{sec:conclusion}
In this research, we discussed different formulations of Bellman error from a distributional perspective. By assuming a Logistic distribution for the Bellman error and integrating the Logistic maximum likelihood function into the associated loss function, we observed enhanced training efficacy in both online and offline RL, marking a departure from the typical use of normal or Gumbel distributions. Our theory's validity is substantiated by rigorous analysis and proofs, as well as empirical evaluations. Moreover, we naturally integrate the Bellman error distribution with the reward scaling problem and propose a sampling scheme based on this distribution for error limit control.

While we have introduced a novel avenue for improving RL optimization focusing on the Bellman error, there remain compelling future directions for exploration. For example, extending our analysis beyond the Bellman iterative equation to include soft Bellman iterations could offer further insights. The formulation of the state transition function might also benefit from a linear combination of Gumbel distributions. Moreover, exploring innovative methods for learning from an unknown biased distribution could be another promising direction, aligning with the inherently biased nature of the distribution of Bellman error.

\bibliography{main.bbl}

\begin{thebibliography}{44}
\providecommand{\natexlab}[1]{#1}
\providecommand{\url}[1]{\texttt{#1}}
\expandafter\ifx\csname urlstyle\endcsname\relax
  \providecommand{\doi}[1]{doi: #1}\else
  \providecommand{\doi}{doi: \begingroup \urlstyle{rm}\Url}\fi

\bibitem[An(1933)]{an1933sulla}
Kolmogorov An.
\newblock Sulla determinazione empirica di una legge didistribuzione.
\newblock \emph{Giorn Dell'inst Ital Degli Att}, 4:\penalty0 89--91, 1933.

\bibitem[Baird(1995)]{baird1995residual}
Leemon Baird.
\newblock Residual algorithms: Reinforcement learning with function approximation.
\newblock In \emph{Machine Learning Proceedings 1995}, pp.\  30--37. Elsevier, 1995.

\bibitem[Bas-Serrano et~al.(2021)Bas-Serrano, Curi, Krause, and Neu]{bas2021logistic}
Joan Bas-Serrano, Sebastian Curi, Andreas Krause, and Gergely Neu.
\newblock Logistic q-learning.
\newblock In \emph{International Conference on Artificial Intelligence and Statistics}, pp.\  3610--3618. PMLR, 2021.

\bibitem[Bayramo{\u{g}}lu et~al.(2021)Bayramo{\u{g}}lu, Erzin, Sezgin, and Yemez]{bayramouglu2021engagement}
{\"O}yk{\"u}~Zeynep Bayramo{\u{g}}lu, Engin Erzin, Tevfik~Metin Sezgin, and Y{\"u}cel Yemez.
\newblock Engagement rewarded actor-critic with conservative q-learning for speech-driven laughter backchannel generation.
\newblock In \emph{Proceedings of the 2021 International Conference on Multimodal Interaction}, pp.\  613--618, 2021.

\bibitem[Bellman(1954)]{bellman1954theory}
Richard Bellman.
\newblock The theory of dynamic programming.
\newblock \emph{Bulletin of the American Mathematical Society}, 60\penalty0 (6):\penalty0 503--515, 1954.

\bibitem[Bertsekas(2014)]{bertsekas2014constrained}
Dimitri~P Bertsekas.
\newblock \emph{Constrained optimization and Lagrange multiplier methods}.
\newblock Academic press, 2014.

\bibitem[Bi et~al.(2022)Bi, Ma, Wang, Cao, Chen, Sun, and Chee]{bi2022learning}
Jieyi Bi, Yining Ma, Jiahai Wang, Zhiguang Cao, Jinbiao Chen, Yuan Sun, and Yeow~Meng Chee.
\newblock Learning generalizable models for vehicle routing problems via knowledge distillation.
\newblock \emph{arXiv:2210.07686}, 2022.

\bibitem[Cabi et~al.(2019)Cabi, Colmenarejo, Novikov, Konyushkova, Reed, Jeong, Zolna, Aytar, Budden, Vecerik, et~al.]{cabi2019scaling}
Serkan Cabi, Sergio~G{\'o}mez Colmenarejo, Alexander Novikov, Ksenia Konyushkova, Scott Reed, Rae Jeong, Konrad Zolna, Yusuf Aytar, David Budden, Mel Vecerik, et~al.
\newblock Scaling data-driven robotics with reward sketching and batch reinforcement learning.
\newblock \emph{arXiv preprint arXiv:1909.12200}, 2019.

\bibitem[Christodoulou(2019)]{christodoulou2019soft}
Petros Christodoulou.
\newblock Soft actor-critic for discrete action settings.
\newblock \emph{arXiv:1910.07207}, 2019.

\bibitem[Dai et~al.(2018)Dai, Shaw, Li, Xiao, He, Liu, Chen, and Song]{dai2018sbeed}
Bo~Dai, Albert Shaw, Lihong Li, Lin Xiao, Niao He, Zhen Liu, Jianshu Chen, and Le~Song.
\newblock Sbeed: Convergent reinforcement learning with nonlinear function approximation.
\newblock In \emph{International Conference on Machine Learning}, pp.\  1125--1134. PMLR, 2018.

\bibitem[Feng et~al.(2019)Feng, Li, and Liu]{feng2019kernel}
Yihao Feng, Lihong Li, and Qiang Liu.
\newblock A kernel loss for solving the bellman equation.
\newblock \emph{Advances in Neural Information Processing Systems}, 32, 2019.

\bibitem[Fisher \& Tippett(1928)Fisher and Tippett]{fisher1928limiting}
Ronald~Aylmer Fisher and Leonard Henry~Caleb Tippett.
\newblock Limiting forms of the frequency distribution of the largest or smallest member of a sample.
\newblock In \emph{Mathematical proceedings of the Cambridge philosophical society}, volume~24, pp.\  180--190. Cambridge University Press, 1928.

\bibitem[Fu et~al.(2020)Fu, Kumar, Nachum, Tucker, and Levine]{fu2020d4rl}
Justin Fu, Aviral Kumar, Ofir Nachum, George Tucker, and Sergey Levine.
\newblock D4rl: Datasets for deep data-driven reinforcement learning.
\newblock \emph{arXiv preprint arXiv:2004.07219}, 2020.

\bibitem[Fujimoto et~al.(2022)Fujimoto, Meger, Precup, Nachum, and Gu]{fujimoto2022should}
Scott Fujimoto, David Meger, Doina Precup, Ofir Nachum, and Shixiang~Shane Gu.
\newblock Why should i trust you, bellman? the bellman error is a poor replacement for value error.
\newblock \emph{arXiv:2201.12417}, 2022.

\bibitem[Gao et~al.(2023)Gao, Schulman, and Hilton]{gao2023scaling}
Leo Gao, John Schulman, and Jacob Hilton.
\newblock Scaling laws for reward model overoptimization.
\newblock In \emph{International Conference on Machine Learning}, pp.\  10835--10866. PMLR, 2023.

\bibitem[Garg et~al.(2023)Garg, Hejna, Geist, and Ermon]{garg2023extreme}
Divyansh Garg, Joey Hejna, Matthieu Geist, and Stefano Ermon.
\newblock Extreme q-learning: Maxent rl without entropy.
\newblock \emph{arXiv:2301.02328}, 2023.

\bibitem[Geist et~al.(2017)Geist, Piot, and Pietquin]{geist2017bellman}
Matthieu Geist, Bilal Piot, and Olivier Pietquin.
\newblock Is the bellman residual a bad proxy?
\newblock \emph{Advances in Neural Information Processing Systems}, 30, 2017.

\bibitem[Gentle(2010)]{gentle2010computational}
James~E Gentle.
\newblock \emph{Computational statistics}.
\newblock Springer, 2010.

\bibitem[Gong et~al.(2020)Gong, Bai, Hou, and Ji]{gong2020stable}
Chen Gong, Yunpeng Bai, Xinwen Hou, and Xiaohui Ji.
\newblock Stable training of bellman error in reinforcement learning.
\newblock In \emph{Neural Information Processing: 27th International Conference, ICONIP 2020, Bangkok, Thailand, November 18--22, 2020, Proceedings, Part V 27}, pp.\  439--448. Springer, 2020.

\bibitem[Haarnoja et~al.(2018{\natexlab{a}})Haarnoja, Zhou, Abbeel, and Levine]{haarnoja2018soft2}
Tuomas Haarnoja, Aurick Zhou, Pieter Abbeel, and Sergey Levine.
\newblock Soft actor-critic: Off-policy maximum entropy deep reinforcement learning with a stochastic actor.
\newblock In \emph{International conference on machine learning}, pp.\  1861--1870. PMLR, 2018{\natexlab{a}}.

\bibitem[Haarnoja et~al.(2018{\natexlab{b}})Haarnoja, Zhou, Hartikainen, Tucker, Ha, Tan, Kumar, Zhu, Gupta, Abbeel, et~al.]{haarnoja2018soft}
Tuomas Haarnoja, Aurick Zhou, Kristian Hartikainen, George Tucker, Sehoon Ha, Jie Tan, Vikash Kumar, Henry Zhu, Abhishek Gupta, Pieter Abbeel, et~al.
\newblock Soft actor-critic algorithms and applications.
\newblock \emph{arXiv:1812.05905}, 2018{\natexlab{b}}.

\bibitem[Hansen-Estruch et~al.(2023)Hansen-Estruch, Kostrikov, Janner, Kuba, and Levine]{hansen2023idql}
Philippe Hansen-Estruch, Ilya Kostrikov, Michael Janner, Jakub~Grudzien Kuba, and Sergey Levine.
\newblock Idql: Implicit q-learning as an actor-critic method with diffusion policies.
\newblock \emph{arXiv:2304.10573}, 2023.

\bibitem[Hejna \& Sadigh(2023)Hejna and Sadigh]{hejna2023inverse}
Joey Hejna and Dorsa Sadigh.
\newblock Inverse preference learning: Preference-based rl without a reward function.
\newblock \emph{arXiv:2305.15363}, 2023.

\bibitem[Hottung et~al.(2021)Hottung, Kwon, and Tierney]{hottung2021efficient}
Andr{\'e} Hottung, Yeong-Dae Kwon, and Kevin Tierney.
\newblock Efficient active search for combinatorial optimization problems.
\newblock \emph{arXiv:2106.05126}, 2021.

\bibitem[Kaiser et~al.(2019)Kaiser, Babaeizadeh, Milos, Osinski, Campbell, Czechowski, Erhan, Finn, Kozakowski, Levine, et~al.]{kaiser2019model}
Lukasz Kaiser, Mohammad Babaeizadeh, Piotr Milos, Blazej Osinski, Roy~H Campbell, Konrad Czechowski, Dumitru Erhan, Chelsea Finn, Piotr Kozakowski, Sergey Levine, et~al.
\newblock Model-based reinforcement learning for atari.
\newblock \emph{arXiv:1903.00374}, 2019.

\bibitem[Konda \& Tsitsiklis(1999)Konda and Tsitsiklis]{konda1999actor}
Vijay Konda and John Tsitsiklis.
\newblock Actor-critic algorithms.
\newblock \emph{Advances in neural information processing systems}, 12, 1999.

\bibitem[Kostrikov et~al.(2021)Kostrikov, Nair, and Levine]{kostrikov2021offline}
Ilya Kostrikov, Ashvin Nair, and Sergey Levine.
\newblock Offline reinforcement learning with implicit q-learning.
\newblock \emph{arXiv:2110.06169}, 2021.

\bibitem[Kumar et~al.(2020)Kumar, Zhou, Tucker, and Levine]{kumar2020conservative}
Aviral Kumar, Aurick Zhou, George Tucker, and Sergey Levine.
\newblock Conservative q-learning for offline reinforcement learning.
\newblock \emph{Advances in Neural Information Processing Systems}, 33:\penalty0 1179--1191, 2020.

\bibitem[Kwon et~al.(2020)Kwon, Choo, Kim, Yoon, Gwon, and Min]{kwon2020pomo}
Yeong-Dae Kwon, Jinho Choo, Byoungjip Kim, Iljoo Yoon, Youngjune Gwon, and Seungjai Min.
\newblock Pomo: Policy optimization with multiple optima for reinforcement learning.
\newblock \emph{Advances in Neural Information Processing Systems}, 33:\penalty0 21188--21198, 2020.

\bibitem[Littman(1993)]{littman1993optimization}
Michael~L Littman.
\newblock An optimization-based categorization of reinforcement learning environments.
\newblock \emph{From animals to animats}, 2:\penalty0 262--270, 1993.

\bibitem[Lyu et~al.(2022)Lyu, Ma, Li, and Lu]{lyu2022mildly}
Jiafei Lyu, Xiaoteng Ma, Xiu Li, and Zongqing Lu.
\newblock Mildly conservative q-learning for offline reinforcement learning.
\newblock \emph{arXiv:2206.04745}, 2022.

\bibitem[Marques et~al.(2015)Marques, Coelho, and De~Carvalho]{marques2015distribution}
Filipe~J Marques, Carlos~A Coelho, and Miguel De~Carvalho.
\newblock On the distribution of linear combinations of independent gumbel random variables.
\newblock \emph{Statistics and Computing}, 25:\penalty0 683--701, 2015.

\bibitem[M{\'e}ndez-Hern{\'a}ndez et~al.(2019)M{\'e}ndez-Hern{\'a}ndez, Rodr{\'\i}guez-Bazan, Martinez-Jimenez, Libin, and Now{\'e}]{mendez2019multi}
Beatriz~M M{\'e}ndez-Hern{\'a}ndez, Erick~D Rodr{\'\i}guez-Bazan, Yailen Martinez-Jimenez, Pieter Libin, and Ann Now{\'e}.
\newblock A multi-objective reinforcement learning algorithm for jssp.
\newblock In \emph{Artificial Neural Networks and Machine Learning--ICANN 2019: Theoretical Neural Computation: 28th International Conference on Artificial Neural Networks, Munich, Germany, September 17--19, 2019, Proceedings, Part I 28}, pp.\  567--584. Springer, 2019.

\bibitem[Mnih et~al.(2013)Mnih, Kavukcuoglu, Silver, Graves, Antonoglou, Wierstra, and Riedmiller]{mnih2013playing}
Volodymyr Mnih, Koray Kavukcuoglu, David Silver, Alex Graves, Ioannis Antonoglou, Daan Wierstra, and Martin Riedmiller.
\newblock Playing atari with deep reinforcement learning.
\newblock \emph{arXiv:1312.5602}, 2013.

\bibitem[Neu et~al.(2017)Neu, Jonsson, and G{\'o}mez]{neu2017unified}
Gergely Neu, Anders Jonsson, and Vicen{\c{c}} G{\'o}mez.
\newblock A unified view of entropy-regularized markov decision processes.
\newblock \emph{arXiv:1705.07798}, 2017.

\bibitem[Osband et~al.(2019)Osband, Doron, Hessel, Aslanides, Sezener, Saraiva, McKinney, Lattimore, Szepesvari, Singh, et~al.]{osband2019behaviour}
Ian Osband, Yotam Doron, Matteo Hessel, John Aslanides, Eren Sezener, Andre Saraiva, Katrina McKinney, Tor Lattimore, Csaba Szepesvari, Satinder Singh, et~al.
\newblock Behaviour suite for reinforcement learning.
\newblock \emph{arXiv preprint arXiv:1908.03568}, 2019.

\bibitem[Pan et~al.(2023)Pan, Huang, Cheng, and Zeng]{pan2023reinforcement}
Jie Pan, Jingwei Huang, Gengdong Cheng, and Yong Zeng.
\newblock Reinforcement learning for automatic quadrilateral mesh generation: A soft actor--critic approach.
\newblock \emph{Neural Networks}, 157:\penalty0 288--304, 2023.

\bibitem[Patterson et~al.(2022)Patterson, White, and White]{patterson2022generalized}
Andrew Patterson, Adam White, and Martha White.
\newblock A generalized projected bellman error for off-policy value estimation in reinforcement learning.
\newblock \emph{The Journal of Machine Learning Research}, 23\penalty0 (1):\penalty0 6463--6523, 2022.

\bibitem[Qi et~al.(2023)Qi, Fan, Karimi, and Su]{qi2023adaptive}
Wen Qi, Haoyu Fan, Hamid~Reza Karimi, and Hang Su.
\newblock An adaptive reinforcement learning-based multimodal data fusion framework for human--robot confrontation gaming.
\newblock \emph{Neural Networks}, 164:\penalty0 489--496, 2023.

\bibitem[Ward et~al.(2019)Ward, Smofsky, and Bose]{ward2019improving}
Patrick~Nadeem Ward, Ariella Smofsky, and Avishek~Joey Bose.
\newblock Improving exploration in soft-actor-critic with normalizing flows policies.
\newblock \emph{arXiv:1906.02771}, 2019.

\bibitem[Watkins \& Dayan(1992)Watkins and Dayan]{watkins1992q}
Christopher~JCH Watkins and Peter Dayan.
\newblock Q-learning.
\newblock \emph{Machine learning}, 8:\penalty0 279--292, 1992.

\bibitem[Zarfaty et~al.(2021)Zarfaty, Barkai, and Kessler]{zarfaty2021accurately}
Lior Zarfaty, Eli Barkai, and David~A Kessler.
\newblock Accurately approximating extreme value statistics.
\newblock \emph{Journal of Physics A: Mathematical and Theoretical}, 54\penalty0 (31):\penalty0 315205, 2021.

\bibitem[Zhang et~al.(2020)Zhang, Zohren, and Stephen]{zhang2020deep}
Zihao Zhang, Stefan Zohren, and Roberts Stephen.
\newblock Deep reinforcement learning for trading.
\newblock \emph{The Journal of Financial Data Science}, 2020.

\bibitem[Ziebart(2010)]{ziebart2010modeling}
Brian~D Ziebart.
\newblock \emph{Modeling purposeful adaptive behavior with the principle of maximum causal entropy}.
\newblock Carnegie Mellon University, 2010.

\end{thebibliography}
\bibliographystyle{iclr2023_conference}
\newpage
\appendix
\onecolumn
\section{Proof for Lemmas and Theorems.}
\subsection{Proof for Lemma~\ref{Lemma_extract}:}

\label{Appendix_extract}
\paragraph{Lemma 2}
If a random variable $X\sim{\rm Gumbel}(A, B)$ follows ${\rm Gumbel}$ distribution with location $A$ and scale $B$, then $X+C\sim{\rm Gumbel}(C+A,B)$ and $DX\sim{\rm Gumbel}(DA,DB)$ with arbitrary constants $C\in\R$ and $D>0$.
\begin{proof}
The Cumulative Probability Density Function (CDF) $P$ for $\rm{Gumbel(A,B)}$ has been given in Section~\ref{sec:2.4} that:
$$
P(X)=e^{-e^{-\frac{(X-A)}{B}}}.
$$
So we have:
$$
P(Y<\alpha)=P(X<\alpha-C)=e^{-e^{-\frac{(\alpha-(C+A))}{B}}},
$$
$$
P(Z<\alpha)=P(X<\frac{\alpha}{D})=e^{-e^{-\frac{(\frac{\alpha}{D}-A)}{B}}}=e^{-e^{-\frac{({\alpha}-DA)}{DB}}},
$$
Which means:
$$
Y \sim {\rm Gumbel}(C+A,B).
$$
$$
Z \sim {\rm Gumbel}(DA,DB).
$$
\end{proof}

\subsection{Proof for Lemma~\ref{Lemma2}}
\label{Appendix1}
\paragraph{Lemma 3}
For a set of mutually independent random variables $X_i \sim {\rm Gumbel}(C_i, \beta)$ ($1\leq i\leq n$), where $C_i$ is a constant related to $X_i$ and $\beta$ is a positive constant, then $\max_i(X_i) \sim {\rm Gumbel}({\beta}\ln\sum_{i=1}^n e^{\frac{1}{\beta} C_i}, \beta)$.
\begin{proof}

As mentioned in Section~\ref{sec:2.4}, the Cumulative Probability Density Function (CDF) $P$ for $\rm{Gumbel}( C_i, \beta)$ is : $e^{-e^{-\frac{(x-C_i)}{\beta}}}$. Based on the independence, we have:
$$
P(\max_i(X_i)<A)=P(X_1<A,X_2<A,X_3<A,..., X_n<A).
$$
where
$$
P(X_i<A)=e^{-e^{-\frac{(A-C_i)}{\beta}}}.
$$
Then
$$
P(X_1<A,X_2<A,X_3<A,..., X_n<A)=e^{-e^{-\frac{(A-C_1)}{\beta}}} \cdot e^{-e^{-\frac{(A-C_2)}{\beta}}} ... \cdot e^{-e^{-\frac{(A-C_n)}{\beta}}} = e^{-\sum_{i=1}^n e^{-\frac{(A-C_i)}{\beta}}},
$$
$$
P(X_1<A,X_2<A,X_3<A,..., X_n<A)=e^{-e^{-\frac{A}{\beta}}\sum_{i=1}^n e^{\frac{C_i}{\beta}}}=e^{-e^{-\frac{A}{\beta}}e^{\ln(\sum_{i=1}^n e^{\frac{C_i}{\beta}}})}=e^{-e^{-\frac{A}{\beta}+\ln(\sum_{i=1}^n e^{\frac{C_i}{\beta}}})},
$$
$$
P(\max_i(X_i)<A)=e^{-e^{-\frac{1}{\beta}[A-{\beta}\ln(\sum_{i=1}^n e^{\frac{C_i}{\beta}})]}},
$$
So this means
$$
\max_i(X_i) \sim {\rm Gumbel}({\beta}\ln\sum_{i=1}^n e^{\frac{1}{\beta} C_i}, \beta).
$$
\end{proof}

\subsection{Proof for Lemma~\ref{Lemma3}}
\label{Appendix2}
\paragraph{Lemma 4}
For $\epsilon^t(\vs,\va)$ defined in (\ref{Eq12}), under Assumptions \ref{assumption1}-\ref{assumption3}, we show that:
\begin{equation*}
    \epsilon^{t}(\vs,\va) \sim {\rm Gumbel}(C_t(\vs,\va)-\gamma \max_{\va'}({Q}^*(\vs',\va')), \beta_t),
\end{equation*}
where 
\begin{equation*}
\begin{aligned}
    C_1(\vs,\va)&=C_1,\\
    C_2(\vs,\va)&=\gamma (C_{1}(\vs,\va)+ \beta_{1} ln \sum_{i=1}^n e^\frac{r(\vs',\va_i)}{\beta_{1}}),
\end{aligned}
\end{equation*}
and
\begin{equation*}
    C_t(\vs,\va)=\gamma (\beta_{t-1} ln \sum_{i=1}^n e^\frac{r(\vs',\va_i)+C_{t-1}(\vs',\va_i)}{\beta_{t-1}}).
\end{equation*}
for $t\geq3$. For $\beta_t$, it always holds that
\begin{equation*}
    \beta_t=\gamma^{t-1}\beta_1
\end{equation*}
for $t\geq1$. Besides, $\epsilon^t(\vs,\va)$ are independent for arbitrary pairs $(\vs,\va)$.

\begin{proof}
If we use the Bellman operator during updating at the $t$-th iteration for estimating from (\ref{Eq4}), we will have:
$$
\hat{Q}^{t}(\vs,\va)=r(\vs,\va)+\gamma \max_{\va'}(\hat{Q}^{t-1}(\vs',\va')).
$$
In Section~\ref{sec:3.1}, we have shown that $Q^{*}$ for (\ref{Eq4}) should satisfy:
$$
Q^{*}(\vs,\va)=r(\vs,\va)+\gamma \max_{\va'}(Q^{*}(\vs',\va')).
$$
By subtracting these two equations, it can be deduced that the error $\epsilon^t(\vs,\va)$ at the $t$-th step is:
$$
\epsilon^{t}(\vs,\va)=\gamma [\max_{\va'}(\hat{Q}^{t-1}(\vs',\va'))- \max_{\va'}(Q^{*}(\vs',\va'))].
$$
Let's see what's going on :

When $t=1$, apparently:
$$
\epsilon^{1}(\vs,\va)=\gamma \max_{\va'}(\hat{Q}^{0}(\vs',\va'))- \gamma \max_{\va'}(Q^{*}(\vs',\va')).
$$
We have shown that:
$$
\gamma \max_{\va'}(\hat{Q}^{0}(\vs',\va')) \sim {\rm Gumbel} (C_1,\beta_1).
$$
While $\gamma \max_{\va'}(Q^{*}(\vs',\va'))$ is a constant, not a random variable, so it does not affect the Gumbel distribution type, but it affects the location of this $\rm Gumbel$ distribution. According to Lemma~\ref{Lemma_extract}, we will have:
$$
\epsilon^{1}(\vs,\va) \sim  {\rm Gumbel}(C_1-\gamma \max_{\va'}(Q^{*}(\vs',\va')),\beta_1).
$$
Let us see what happens if we replace $\vs$ with $\vs'$. By assumption, the action space $\gA$ has finite elements, which means $\gA=[\va_1,\va_2,...,\va_n]$, so we can enumerate all actions to a list with the state-action pair: $[(\vs',\va_1,r_1,\vs^{''}_1), (\vs',\va_2,r_2,\vs^{''}_2),...,(\vs',\va_n,r_n,\vs^{''}_n)]$, where $\vs^{''}_i$ is gotten from $\gT(\vs',\va_i)$. According to the above discussion, we have:
$$
\epsilon^{1}(\vs',\va_i) \sim  {\rm Gumbel}(C_1-\gamma \max_{\va'}(Q^{*}(\vs^{''}_i,\va')),\beta_1).
$$
Noticed that $\epsilon^{1}(\vs',\va_i)$ and $\epsilon^{1}(\vs',\va_j)$ are independent when $i\neq j$, this is because there is obviously no relationship between the two different actions $\va_i $ and $\va_j$. This is not a difficult fact to understand. In fact, we will show in \textbf{Fact 1} that for any different $(\vs, \va)$ pair, $\epsilon^1(\vs,\va)$ will be independent.

\textbf{Fact 1:}
For any different $(\vs, \va)$ pair, $\epsilon^1(\vs,\va)$ will be independent.

This may be surprising, because according to the mapping $\gT$, this has established a relationship between $\vs'_k$ and $(\vs,\va_k)$ with $\vs'_k=\gT(\vs,\va_k)$.

\textbf{Proof for the Fact 1}:

For any two different pairs $(\vs_1,\va_k)$ and $(\vs_2,\va_j)$, define  $\gT(\vs_1,\va_k)=\vs_{1k}'$ and $\gT(\vs_2,\va_j)=\vs_{2j}'$, noticed that:
$$
\epsilon^{1}(\vs_1,\va_k)=\gamma \max_{\va'}(\hat{Q}^{0}(\vs_{1k}',\va'))- \gamma \max_{\va'}(Q^{*}(\vs_{1k}',\va')) \sim {\rm Gumbel}(C_1-\gamma \max_{\va'}(Q^{*}(\vs_{1k}',\va')),\beta_1).
$$
$$
\epsilon^{1}(\vs_2,\va_j)=\gamma \max_{\va'}(\hat{Q}^{0}(\vs_{2j}',\va'))- \gamma \max_{\va'}(Q^{*}(\vs_{2j}',\va')) \sim {\rm Gumbel}(C_1-\gamma \max_{\va'}(Q^{*}(\vs_{2j}',\va')),\beta_1).
$$

According to Assumption~\ref{assumption3}, we show that $\gamma \max_{\va'}(\hat{Q}^{0}(\vs_{1k}',\va'))$ and $\gamma \max_{\va'}(\hat{Q}^{0}(\vs_{2j}',\va'))$ are independent. This is due to the randomness of the initialization. On the other hand, $\gamma \max_{\va'}(Q^{*}(\vs_{1k}',\va'))$ and $\gamma \max_{\va'}(Q^{*}(\vs_{2j}',\va'))$ are two fixed number. Although they are constrained by the Bellman equation, they are not variables. Therefore, $\epsilon^{1}(\vs_1,\va_k)$ is independent with $\epsilon^{1}(\vs_2,\va_j)$, in this way we have proved \textbf{Fact 1}. We will see later that only at the same time $t$ can keep this property.

Let us continue our discussion. When $t=2$, we will have:
$$
\epsilon^{2}(\vs,\va)=\gamma \max_{\va'}(Q^{*}(\vs',\va')+ \epsilon^{1}(\vs',\va'))- \gamma \max_{\va'}(Q^{*}(\vs',\va')).
$$
Let $Q^{*}(\vs',\va_i)+ \epsilon^{1}(\vs',\va_i)$ be $L_i$, according to \textbf{Fact 1} we have discussed above, $L_i$ is a sequence of mutually independent countable random variables, using Lemma~\ref{Lemma2}, we can have:
$$
L_i \sim  {\rm Gumbel}(Q^{*}(\vs',\va_i)+C_1-\gamma \max_{\va'}(Q^{*}(\vs^{''}_i,\va')), \beta_1).
$$
Noticed that:
$$
Q^{*}(\vs',\va_i)=r(\vs',\va_i)+\gamma \max_{\va'}(Q^{*}(\vs^{''}_i,\va')).
$$
So:
$$
L_i \sim  {\rm Gumbel}(r(\vs',\va_i)+C_1, \beta_1).
$$
According to Lemma~\ref{Lemma2}, then 
$$
max_{\va_i}(Q^{*}(\vs',\va_i)+ \epsilon^{1}(\vs',\va_i))=max_i(L_i) \sim {\rm Gumbel}(C_1+\beta_1 ln \sum_{i=1}^n e^\frac{r(\vs',\va_i)}{\beta_1}, \beta_1). 
$$
Let $\gamma max_i(L_i) \sim {\rm Gumbel}(C_2 (\vs,\va),\beta_2)$. Because the discounted factor $\gamma$ is positive number, according to Lemma~\ref{Lemma2}:
$$
C_2 (\vs,\va) = \gamma (C_1+\beta_1 ln \sum_{i=1}^n e^\frac{r(\vs',\va_i)}{\beta_1}).
$$
$$
\beta_2 = \gamma (\beta_1).
$$
So:
$$
\epsilon^{2}(\vs,\va) \sim {\rm Gumbel}(C_2(\vs,\va)-\gamma \max_{\va'}(Q^{*}(\vs',\va')), \beta_2).
$$
We also have a \textbf{Fact 2} similar to \textbf{Fact 1}.

\textbf{Fact 2:} For any different $(\vs, \va)$ pair, $\epsilon^2(\vs,\va)$ will be independent.

\textbf{Proof for the Fact 2}:
For any two different pairs $(\vs_1,\va_k)$ and $(\vs_2,\va_j)$, define  $\gT(\vs_1,\va_k)=\vs_{1k}'$ and $\gT(\vs_2,\va_j)=\vs_{2j}'$, noticed that:
$$
\epsilon^{2}(\vs_1,\va_k)=\gamma \max_{\va'}(Q^{*}(\vs_{1k}',\va')+ \epsilon^{1}(\vs_{1k}',\va'))- \gamma \max_{\va'}(Q^{*}(\vs_{1k}',\va')).
$$
From here we can see that, \textbf{$\epsilon^{2}(\vs_1,\va_k)$ and any $\epsilon^{1}(\vs_{1k}',\va_j)$ are not independent}, on the other hand:
$$
\epsilon^{2}(\vs_2,\va_j)=\gamma \max_{\va'}(Q^{*}(\vs_{2j}',\va')+ \epsilon^{1}(\vs_{2j}',\va'))- \gamma \max_{\va'}(Q^{*}(\vs_{2j}',\va')).
$$

According to \textbf{Fact 1}, for any action $a_m$, $a_n$, $\epsilon^{1}(\vs_{1k}',\va_n)$ is independent with any $\epsilon^{1}(\vs_{2j}',\va_m)$. $Q^*(\vs_{2j}',\va_m)$ is a fixed number. It does not introduce any randomness, so it does not affect independence and randomness. So $\epsilon^{1}(\vs_{1k}',\va_n)$ is independent with $\gamma \max_{\va'}(Q^{*}(\vs_{2j}',\va')+ \epsilon^{1}(\vs_{2j}',\va'))$ for any $n$. Similarly, $\gamma \max_{\va'}(Q^{*}(\vs_{2j}',\va')+ \epsilon^{1}(\vs_{2j}',\va'))$ is independent with $\gamma \max_{\va'}(Q^{*}(\vs_{1k}',\va')+ \epsilon^{1}(\vs_{1k}',\va'))$. For the rest of the part $\gamma \max_{\va'}(Q^{*}(\vs_{1k}',\va'))$ and $\gamma \max_{\va'}(Q^{*}(\vs_{2j}',\va'))$. they can all be treated as the fix constants, so they don't affect the independence, so we have proved the \textbf{Fact 2}.

At this point, we can already discern some patterns. However, to ensure thoroughness, we will conduct one more iteration here for $t=3$:
$$
\epsilon^{3}(\vs,\va)=\gamma \max_{\va'}(Q^{*}(\vs',\va')+ \epsilon^{2}(\vs',\va'))- \gamma \max_{\va'}(Q^{*}(\vs',\va')).
$$
In this case, Let $Q^{*}(\vs',\va_i)+ \epsilon^{2}(\vs',\va_i)$ be $M_i$, obviously there is no connection between $a_i$ and $a_j$ when $i  \neq j$. So $M_i$ is a sequence of mutually independent countable random variables, using Lemma~\ref{Lemma2} again, we can have:
$$
M_i \sim  {\rm Gumbel}(Q^{*}(\vs',\va_i)+C_2(\vs',\va_i)-\gamma \max_{\va'}(Q^{*}(\vs^{''}_i,\va')), \beta_2).
$$
$$
M_i \sim  {\rm Gumbel}(r(\vs',\va_i)+C_2(\vs',\va_i), \beta_2).
$$

According to Lemma~\ref{Lemma2}, then 
$$
max_{\va_i}(Q^{*}(\vs',\va_i)+ \epsilon^{2}(\vs',\va_i))=max_i(M_i) \sim {\rm Gumbel}(\beta_2 ln \sum_{i=1}^n e^\frac{r(\vs',\va_i)+C_2(\vs',\va_i)}{\beta_2}, \beta_2). 
$$
Let:
$$
C_3(s,a)=\gamma (\beta_2 ln \sum_{i=1}^n e^\frac{r(\vs',\va_i)+C_2(\vs',\va_i)}{\beta_2}).
$$
$$
\beta_3=\gamma \beta_2.
$$
We will have:
$$
\epsilon^{3}(\vs,\va) \sim {\rm Gumbel}(C_3(\vs,\va)-\gamma \max_{\va'}(Q^{*}(\vs',\va')), \beta_3).
$$
Similar to \textbf{Fact 1} and \textbf{Fact 2}, of course, there is \textbf{Fact 3} to hold.

\textbf{Fact 3:} For any different $(\vs, \va)$ pair, $\epsilon^3(\vs,\va)$ will be independent.

The proof for \textbf{Fact 3} is the same as \textbf{Fact 2}.

Continuing in this manner, we will find that when $(t \geq 3)$, the approach becomes identical. We will have a general iteration format:
$$
\epsilon^{t}(\vs,\va) \sim {\rm Gumbel}(C_t(s,a)-\gamma \max_{\va'}(Q^{*}(\vs',\va')), \beta_t). 
$$
Where 
$$
C_2(\vs,\va)=\gamma (C_{1}+ \beta_{1} ln \sum_{i=1}^n e^\frac{r(\vs',\va_i)}{\beta_{1}}).
$$
$$
C_t(\vs,\va)=\gamma (\beta_{t-1} ln \sum_{i=1}^n e^\frac{r(\vs',\va_i)+C_{t-1}(\vs',\va_i)}{\beta_{t-1}}) (t \geq 3).
$$
We also have a summary fact here.

\textbf{Summary Fact:}
For any different $(\vs, \va)$ pair, $\epsilon^{t}(\vs,\va)$ will be independent for the same $t$.

The reason we do not merge $C_2$ and $C_t$ is to emphasize that $C_1$ is a constant. If the following special cases in Remark~\ref{remark1} can be satisfied, it will be found that all $C_i$ for any $i$ are constants without distinction.

\textbf{Proof for Remark~\ref{remark1}:}

If for $\forall s_1, s_2$, let us define $S_1$ and $S_2$ sets as follows:
$$S_1=[r(\vs_1,\va_1),r(\vs_1,\va_2),..., r(\vs_1,\va_n)]. $$
$$S_2=[r(\vs_2,\va_1),r(\vs_2,\va_2),..., r(\vs_2,\va_n)]. $$
Obviously neither $S_1$ and $S_2$ are empty set, if $S_1$ and $S_2$ satisfy:
$$
S_1 \triangle S_2 = \emptyset.
$$

Then when $t=2$:
$$
C_2=\gamma (C_1+\beta_1 ln \sum_{i=1}^n e^\frac{r(\vs',\va_i)}{\beta_1}).
$$
$$
\beta_2=\gamma (\beta_1).
$$
So:
$$
\epsilon^{2}(\vs,\va) \sim {\rm Gumbel}(C_2-\gamma \max_{\va'}(Q^{*}(\vs',\va')), \beta_2).
$$
This means this condition removes the correlation between $C_i$ and $(\vs,\va)$ under our assumption.
So:
$$
\epsilon^{t}(\vs,\va) \sim {\rm Gumbel}(C_t-\gamma \max_{\va'}(Q^{*}(\vs',\va')), \beta_t).
$$
Where
$$
C_t=\gamma(C_{t-1}+\beta_{t-1} ln \sum_{i=1}^n e^\frac{r(\vs',\va_i)}{\beta_{t-1}}) (t \geq 2).
$$
\end{proof}

\subsection{Proof for Lemma~\ref{Lemma4}:}
\paragraph{Lemma 5}
For random variables $X \sim {\rm Gumbel}(C_X,\beta)$ and $Y \sim {\rm Gumbel}(C_Y,\beta)$, if $X$ and $Y$ are independent, then $(X-Y) \sim {\rm Logistic}(C_X-C_Y,\beta)$.

\begin{proof}
Let $p_1(X),p_2(Y)$ as the PDF for $X, Y$. $P_1(X),P_2(Y)$ as the CDF for $X, Y$.
$$
P(X-Y<z)=P(X<Y+z)=\int_{-\infty}^{+\infty} \int_{-\infty}^{Y+z} p_1(X) p_2(Y) dXdY=\int_{-\infty}^{+\infty} P_1(Y+z)p_2(Y) dY.
$$
So:
$$
\int_{-\infty}^{+\infty} P_1(Y+z)p_2(Y) dY=\int_{-\infty}^{+\infty} e^{-e^{-\frac{Y+z-C_X}{\beta}}} \frac{1}{\beta}e^{-(\frac{Y-C_Y}{\beta}+e^{-\frac{Y-C_Y}{\beta}})} dY,
$$
$$
\frac{1}{\beta} \int_{-\infty}^{+\infty} e^{-e^{-\frac{Y+z-C_X}{\beta}}} e^{-(\frac{Y-C_Y}{\beta}+e^{-\frac{Y-C_Y}{\beta}})} dY= \frac{1}{\beta} \int_{-\infty}^{+\infty} e^{-\frac{Y-C_Y}{\beta}} e^{-e^{-\frac{Y-C_Y}{\beta}}(1+e^{\frac{C_X-C_Y-z}{\beta}})} dY,
$$
Take $U=e^{-\frac{Y-C_Y}{\beta}}$, then $dU=-\frac{1}{\beta} U dY$, then:
$$
\frac{1}{\beta} \int_{-\infty}^{+\infty} e^{-\frac{Y-C_Y}{\beta}} e^{-e^{-\frac{Y-C_Y}{\beta}}(1+e^{\frac{C_X-C_Y-z}{\beta}})} dY=\int_0^{+\infty}e^{-U(1+e^{\frac{C_X-C_Y-z}{\beta}})}dU=\frac{1}{(1+e^{\frac{C_X-C_Y-z}{\beta}})},
$$
So, we show that:
$$
P(X-Y<z)=\frac{1}{(1+e^{\frac{C_X-C_Y-z}{\beta}})}=\frac{1}{(1+e^{-\frac{z-(C_X-C_Y)}{\beta}})}.
$$
According to Section~\ref{sec:2.4}, we know that: 
$$
X-Y \sim \rm{Logistic}(C_X-C_Y,\beta).
$$
\end{proof}
\subsection{Proof for Lemma~\ref{Lemma5}:}
\label{Appendix_Lemma6}
\paragraph{Lemma 6}
If the random variable $X \sim \rm{Gumbel}(A,1)$, then $\E[e^{-X}]$, $\E[Xe^{-X}]$ can all be bounded:

(1)
$$
\E[e^{-X}]<(\frac{20}{e^2} +10e^{-e^{\frac{1}{2}}}+ \frac{1}{2}-\frac{1}{2e})e^{-A}.
$$
(2)

When $A>0$:
$$
\E[Xe^{-X}]<(\frac{3}{20}+A(\frac{20}{e^2} +10e^{-e^{\frac{1}{2}}}+ \frac{1}{2}-\frac{1}{2e}))e^{-A}.
$$
When $A \leq 0$:
$$
\E[Xe^{-X}]<(\frac{3}{20})e^{-A}.
$$

\begin{proof}
Let us assume that $X \sim \rm{Gumbel}(0,1)$ first.
$$
X \sim \rm{Gumbel}(0,1), p(X)=e^{-(X+e^{-X})}.
$$
For (1):
$$
\E[e^{-X}]=\int_{-\infty}^{+\infty} e^{-X}p(X) dX=\int_{-\infty}^{+\infty} e^{-X} \cdot e^{-(X+e^{-X})} dX=\int_{-\infty}^{+\infty} e^{-(2X+e^{-X})} dX.
$$
For (2):
$$
\E[Xe^{-X}]=\int_{-\infty}^{+\infty} Xe^{-(2X+e^{-X})} dX.
$$
We split this integral into the parts for $X>0$ and $X<0$ for separate discussions now. When $X>0$, it is easy to see that:
$$
e^{2X}>e^{X} \rightarrow e^{-2X}<e^{-X} \rightarrow e^{-2X}-e^{-X}<0.
$$
So we will have:
$$
\frac{e^{(-2X-e^{-X}})}{e^{(-2X-e^{-2X}})}=e^{(e^{-2X}-e^{-X})}<1 \rightarrow \int_0^{+\infty} e^{-(2X+e^{-X})} dX< \int_0^{+\infty} e^{-(2X+e^{-2X})} dX< \int_{-\infty}^{+\infty} e^{-(2X+e^{-2X})} dX.
$$
In fact:
$$
\int_0^{+\infty} e^{-(2X+e^{-2X})} dX=\frac{1}{2}-\frac{1}{2e}.
$$
On the other hand, obviously:
$$
\int_0^{+\infty} Xe^{-(2X+e^{-X})} dX< \int_0^{+\infty} Xe^{-(2X+e^{-2X})} dX.
$$
When we take $X<-5$, for (1), we will have:
$$
\frac{e^{(-2X-e^{-X}})}{e^{(-0.1X-e^{-0.1X}})}=e^{(-1.9X+e^{-0.1X}-e^{-X})}<1 \rightarrow \int_{-\infty}^{-5} e^{-(2X+e^{-X})} dX< \int_{-\infty}^{-5} e^{-(0.1X+e^{-0.1X})} dX.
$$
In fact:
$$
\int_{-\infty}^{-5} e^{-(0.1X+e^{-0.1X})} dX=10e^{-e^{\frac{1}{2}}}.
$$
For (2), if we take $X<0$:
$$
\frac{e^{(-2X-e^{-X}})}{e^{(-2X-e^{-2X}})}=e^{(e^{-2X}-e^{-X})}>1 \rightarrow Xe^{(-2X-e^{-X}})<Xe^{(-2X-e^{-2X}}).
$$
$$
\int_{-\infty}^{0} Xe^{-(2X+e^{-X})} dX< \int_{-\infty}^{0} Xe^{-(2X+e^{-2X})} dX.
$$
When $-5\leq X \leq0$, let $U(X)=2X+e^{-X}$:
$$
\frac{dU}{dX}=0 \rightarrow x=-ln2 \rightarrow min(U(x))=2-2ln2 \rightarrow max(-U(x))=2ln2-2.
$$
$$
\int_{-5}^0 e^{-(2X+e^{-X})} dX \leq 5(e^{ln4-2})=\frac{20}{e^2}.
$$
So, we can easily observe that:
$$
\int e^{-(2X+e^{-X})} dX=\int_{-\infty}^{-5} e^{-(2X+e^{-X})} dX+\int_{-5}^0 e^{-(2X+e^{-X})} dX+\int_0^{+\infty} e^{-(2X+e^{-X})} dX.
$$
So:
$$
\int e^{-(2X+e^{-X})} dX < \frac{20}{e^2} +10e^{-e^{\frac{1}{2}}}+ \frac{1}{2}-\frac{1}{2e}.
$$
$$
\int Xe^{-(2X+e^{-X})} dX < \int_{-\infty}^{0} Xe^{-(2X+e^{-2X})} dX+\int_0^{+\infty} Xe^{-(2X+e^{-2X})} dX.
$$
The expectation of the Gumbel distribution is known. In fact, if $X\sim \rm{Gumbel}(A,B)$, then $\E[X]=A+vB$ where $v\approx 0.5772 <0.6$ represent the Euler–Mascheroni constant. This has already been discussed in Section~\ref{sec:2.4}.
In summary:
$$
\int e^{-(2X+e^{-X})} dX<\frac{20}{e^2} +10e^{-e^{\frac{1}{2}}}+ \frac{1}{2}-\frac{1}{2e}.
$$
$$
\int Xe^{-(2X+e^{-X})} dX<(\frac{1}{4})v<(\frac{1}{4})\frac{3}{5}=\frac{3}{20}.
$$
These are the boundaries when $X$ follows a $Gumbel(0,1)$ distribution. Now, let's consider the case when $X$ follows a $Gumbel(A,1)$ distribution. If $X\sim Gumbel(A,1)$, according to Lemma~\ref{Lemma_extract},$X-A\sim Gumbel(0,1)$, then $\E(e^{-(X-A)})$ can be bounded:
$$
\E(e^{A-X})<\frac{20}{e^2} +10e^{-e^{\frac{1}{2}}}+ \frac{1}{2}-\frac{1}{2e}  \rightarrow \E[e^{-X}]<(\frac{20}{e^2} +10e^{-e^{\frac{1}{2}}}+ \frac{1}{2}-\frac{1}{2e})e^{-A}.
$$
$$
\E((X-A)e^{A-X})<\frac{3}{20}.\rightarrow e^A\E[Xe^{-X}]-Ae^A\E[e^{-X}]<\frac{3}{20}.
$$
$$
e^A\E[Xe^{-X}]<\frac{3}{20}+Ae^A\E[e^{-X}].
$$
So when $A>0$:
$$
e^A\E[Xe^{-X}]<\frac{3}{20}+A(\frac{20}{e^2} +10e^{-e^{\frac{1}{2}}}+ \frac{1}{2}-\frac{1}{2e}) \rightarrow \E[Xe^{-X}]<(\frac{3}{20}+A(\frac{20}{e^2} +10e^{-e^{\frac{1}{2}}}+ \frac{1}{2}-\frac{1}{2e}))e^{-A}.
$$
But when $A<0$, noticed that:$\E[e^{-X}]>0$, so:
$$
e^A\E[Xe^{-X}]<\frac{3}{20} \rightarrow \E[Xe^{-X}]<(\frac{3}{20})e^{-A}.
$$
\end{proof}

\subsection{Proof for Theorem~\ref{Theorem1}:}
\label{Appendix3}
\paragraph{Theorem 1}
\textbf{(Logistic distribution for Bellman error)}:
The Bellman error $\varepsilon^{\theta}(\vs,\va)$ approximately follows the $\rm{Logistic}$ distribution under the Assumptions~\ref{assumption1}-\ref{assumption4}. The degree of approximation can be measured by the upper bound of KL divergence between:
$$X \sim {\rm Gumbel}(\beta_{\theta} {\rm ln}\sum_{i=1}^n e^{\frac{r(\vs',\va_i)+C_{\theta}(\vs',\va_i)}{\beta_{\theta}}},\beta_{\theta}).
$$ 
and
$$
Y\sim { \rm Gumbel}(\gamma \beta_{\theta} {\rm ln}\sum_{i=1}^n e^{\frac{r(\vs',\va_i)+C_{\theta}(\vs',\va_i)}{\beta_{\theta}}},\gamma \beta_{\theta}).
$$
Let $A^*={\rm ln}\sum_{i=1}^n e^{\frac{r(\vs',\va_i)+C_{\theta}(\vs',\va_i)}{\beta_{\theta}}}$, we have these conclusions:
\begin{enumerate}[leftmargin=*]
    \item If $A^*>0$, then ${\rm KL}(Y||X)<\log(\frac{1}{\gamma})+(1-\gamma)[A^*(\frac{20}{e^2} +10e^{-e^{\frac{1}{2}}}- \frac{1}{2}-\frac{1}{2e})+\frac{3}{20}-v]$.
    \item If $A^* \leq 0$, then ${\rm KL}(Y||X)<\log(\frac{1}{\gamma})+(1-\gamma)[\frac{3}{20}-A^*-v]$.
    \item The order of the KL divergence error is controlled at $O(\log(\frac{1}{1-\kappa_0})+\kappa_0A^*)$.
\end{enumerate}
If the upper bound of KL divergence is sufficiently small. Then ${\rm \varepsilon}^{\theta}(\vs,\va)$ follows the $\rm{Logistic}$ distribution, \ie
\begin{equation*}
    {\rm \varepsilon}^{\theta}(\vs,\va) \sim {\rm Logistic} (C_{\theta}(\vs,\va)-\beta_{\theta} {\rm ln}\sum_{i=1}^n e^{\frac{r(\vs',\va_i)+C_{\theta}(\vs',\va_i)}{\beta_{\theta}}},\beta_{\theta}).
\end{equation*}

\begin{proof}
    According to Equation~\ref{Eq11}, we will have the definition for the Bellman error under the setting of parameter $\theta$:
    $$
    {\varepsilon}^{\theta} (\vs,\va)=\hat{Q}_{\theta}(\vs,\va)-r(\vs,\va)-\gamma \max_{\va'}(\hat{Q}_{\theta}(\vs',\va')).
    $$
    Where:
    $$
    \hat{Q}_{\theta}(\vs,\va)=Q^{*}(\vs,\va)+\epsilon^{\theta}{(\vs,\va)}.
    $$
    From the proof of Lemma~\ref{Lemma3}, we have already known that:
    $$
    \epsilon^{\theta}{(\vs,\va)} \sim {\rm Gumbel} (C_{\theta}(\vs,\va)-\gamma \max_{\va'}(Q^{*}(\vs',\va')), \beta_{\theta}).
    $$
    where:
    $$
    C_{\theta}(\vs,\va)=\gamma (\beta_{(\theta-1)} ln \sum_{i=1}^n e^\frac{r(\vs',\va_i)+C_{(\theta-1)}(\vs',\va_i)}{\beta_{(\theta-1)}}) (\theta \geq 2).
    $$
    $$
    \beta_{\theta}=\gamma^{(\theta-1)}\beta_1.
    $$
    So:
    $$
    {\rm \varepsilon}^{\theta}(\vs,\va) = Q^{*}(\vs,\va)+\epsilon^{\theta}(\vs,\va)-r(\vs,\va)-\gamma \max_{\va'}(\hat{Q}_{\theta}(\vs',\va')).
    $$
    Because of:
    $$
    Q^{*}(\vs,\va)=r(\vs,\va)+\gamma max_{\va'}Q^{*}(\vs',\va')
    $$
    So:
    $$
    {\rm \varepsilon}^{\theta}(\vs,\va) = \gamma \max_{\va'}[Q^{*}(\vs',\va')]+\epsilon^{\theta}(\vs,\va)-\gamma \max_{\va'}[Q^{*}(\vs',\va')+\epsilon^{\theta}(\vs',\va')].
    $$
    Notice that this equation has two parts: $(1) \gamma \max_{\va'}[Q^{*}(\vs',\va')]+\epsilon^{\theta}(\vs,\va)$ and $(2) \gamma \max_{\va'}[Q^{*}(\vs',\va')+\epsilon^{\theta}(\vs',\va')]$. Let us discuss them separately.

    Let us first analyze part $(1)$, according to Lemma~\ref{Lemma_extract}, it is easy to have:
    $$
    \gamma \max_{\va'}[Q^{*}(\vs',\va')]+\epsilon^{\theta}(\vs,\va) \sim  {\rm Gumbel}(C_{\theta}(\vs,\va),\beta_{\theta}).
    $$
    For another part $(2)$:
    $$
    \epsilon^{\theta}{(\vs',\va_i)} \sim {\rm Gumbel}(C_{\theta}(\vs',\va_i) -\gamma \max_{\va'}(Q^{*}(\vs_i{''},\va')), \beta_{\theta}).
    $$
    Thus using Lemma~\ref{Lemma_extract}, we have:
    $$
    [Q^{*}(\vs',\va_i)+\epsilon^{\theta}(\vs',\va_i)] \sim {\rm Gumbel}(C_{\theta}(\vs',\va_i) -\gamma \max_{\va'}(Q^{*}(\vs_i{''},\va'))+Q^{*}(\vs',\va_i), \beta_{\theta}).
    $$
    Because of:
    $$
    -\gamma \max_{\va'}(Q^{*}(\vs_i{''},\va'))+Q^{*}(\vs',\va_i)=r(\vs',\va_i).
    $$
    So:
    $$
    L_i=[Q^{*}(\vs',\va_i)+\epsilon^{\theta}(\vs',\va_i)] \sim {\rm Gumbel}(C_{\theta}(\vs',\va_i)+r(\vs',\va_i),\beta_{\theta}).
    $$
    In the proof of Lemma~\ref{Lemma3}, the independence of $L_i$ has already been taken into account, therefore, using Lemma~\ref{Lemma2}, we can know that:
    $$
    max_{a_i}[Q^{*}(\vs',\va_i)+\epsilon^{\theta}(\vs',\va_i)] \sim {\rm Gumbel}(\beta_{\theta} ln\sum_{i=1}^n e^{\frac{r(\vs',\va_i)+C_{\theta}(\vs',\va_i)}{\beta_{\theta}}},\beta_{\theta}).
    $$
    According to the proof of Lemma~\ref{Lemma3}, $max_{a_i}[Q^{*}(\vs',\va_i)+\epsilon^{\theta}(\vs',\va_i)]$ and $\gamma \max_{\va'}[Q^{*}(\vs',\va')]+\epsilon^{\theta}(\vs,\va)$ are independent under the same time $\theta$. Now we want to use the Lemma~\ref{Lemma4}, according to Lemma~\ref{Lemma_extract}, we noticed that:
    $$
    \gamma max_{a_i}[Q^{*}(\vs',\va_i)+\epsilon^{\theta}(\vs',\va_i)]\sim {\rm Gumbel}(\gamma\beta_{\theta} ln\sum_{i=1}^n e^{\frac{r(\vs',\va_i)+C_{\theta}(\vs',\va_i)}{\beta_{\theta}}},\gamma\beta_{\theta}).
    $$
    $$
    \gamma \max_{\va'}[Q^{*}(\vs',\va')]+\epsilon^{\theta}(\vs,\va) \sim {\rm Gumbel}(C_{\theta}(\vs,\va),\beta_{\theta}).
    $$
    Thus we cannot use Lemma~\ref{Lemma4} directly because the scale parameters are not the same even though they are independent, so we need to give an approximation with certain error conditions now.

    Assume that:
    $$
    X=max_{a_i}[Q^{*}(\vs',\va_i)+\epsilon^{\theta}(\vs',\va_i)]\sim {\rm Gumbel}(A,B).
    $$
    $$
    Y=\gamma max_{a_i}[Q^{*}(\vs',\va_i)+\epsilon^{\theta}(\vs',\va_i)]\sim {\rm Gumbel}(\gamma A,\gamma B).
    $$
    where:
    $$
    A=\beta_{\theta} ln\sum_{i=1}^n e^{\frac{r(\vs',\va_i)+C_{\theta}(\vs',\va_i)}{\beta_{\theta}}}.
    $$
    $$
    B=\beta_{\theta}.
    $$
    $$
    A^*=\frac{A}{B}=ln\sum_{i=1}^n e^{\frac{r(\vs',\va_i)+C_{\theta}(\vs',\va_i)}{\beta_{\theta}}}.
    $$
    Let us see the KL divergence between these two distributions, we treat the PDF for $Gumbel(A, B)$ and $Gumbel(\gamma A,\gamma B)$ as $p(x)$ and $q(x)$, according to Section~\ref{sec:2.4}, we have shown that:
    $$
    p(x)=\frac{1}{B}e^{-(\frac{x-A}{B}+e^{-\frac{x-A}{B}})}.
    $$
    $$
    q(x)=\frac{1}{\gamma B}e^{-(\frac{x-\gamma A}{\gamma B}+e^{-\frac{x-\gamma A}{\gamma B}})}.
    $$    
    $$
    {\rm KL}(q(x)||p(x))=\E_{x~\sim q(x) }[log(\frac{q(x)}{p(x)})]=\E_{x~\sim q(x)}[log(\frac{\frac{1}{\gamma B}e^{-(\frac{x-\gamma A}{\gamma B}+e^{-\frac{x-\gamma A}{\gamma B}})}}{\frac{1}{B}e^{-(\frac{x-A}{B}+e^{-\frac{x-A}{B}})}})],
    $$
    $$
    {\rm KL}(q(x)||p(x))=\E_{x~\sim q(x)}[log(\frac{1}{\gamma})-(\frac{x-\gamma A}{\gamma B}+e^{-\frac{x-\gamma A}{\gamma B}})+(\frac{x-A}{B}+e^{-\frac{x-A}{B}})],
    $$
    $$
    {\rm KL}(q(x)||p(x))=log(\frac{1}{\gamma})+(\frac{1}{B}-\frac{1}{\gamma B})\E_{x\sim q(x)}[x]+e^{\frac{A}{B}} \E_{x\sim q(x)}[e^{-\frac{x}{B}}-e^{-\frac{x}{\gamma B}}].
    $$
    Where $0<\gamma<1$ is the discounted factor, $1-\gamma=\kappa<\kappa_0\leq \delta_0$ with sufficiently small $\delta_0$.

    Using Lemma ~\ref{Lemma_extract}, we have shown that if $x\sim {\rm Gumbel}(\gamma A,\gamma B)$, then $x'=\frac{x}{\gamma B} \sim {\rm Gumbel}(\frac{A}{B},1)$. $dx'=\frac{1}{\gamma B} dx$. 
    So:
    $$
    \E_{x\sim q(x)}[e^{-\frac{x}{\gamma B}}]=\int_{-\infty}^{+\infty} \frac{1}{\gamma B}e^{-(\frac{x-\gamma A}{\gamma B}+e^{-\frac{x-\gamma A}{\gamma B}})} e^{-\frac{x}{\gamma B}} dx= \int_{-\infty}^{+\infty} e^{-(x'-\frac{A}{B}+e^{-(x'-\frac{A}{B})})}e^{-x'}dx'=\E_{x'}[e^{-x'}].
    $$
    According to Lemma~\ref{Lemma5}:
    $$
    \E_{x\sim q(x)}[e^{-\frac{x}{\gamma B}}]=\E_{x'}[e^{-x'}<(\frac{20}{e^2} +10e^{-e^{\frac{1}{2}}}+ \frac{1}{2}-\frac{1}{2e})e^{-\frac{A}{B}}.
    $$
    $$
    \E_{x\sim q(x)}[xe^{-\frac{x}{\gamma B}}]= \E_{x'}[\gamma B x' e^{-x'}]=\gamma B\E_{x'}[x' e^{-x'}]
    $$
    If $A>0$, according to Lemma~\ref{Lemma5}:
    $$
    \E_{x\sim q(x)}[xe^{-\frac{x}{\gamma B}}] <(\frac{3}{20}+\frac{A}{B}(\frac{20}{e^2} +10e^{-e^{\frac{1}{2}}}+ \frac{1}{2}-\frac{1}{2e}))e^{-\frac{A}{B}}\gamma B=(\frac{3}{20}\gamma B+\gamma A(\frac{20}{e^2} +10e^{-e^{\frac{1}{2}}}+ \frac{1}{2}-\frac{1}{2e}))e^{-\frac{A}{B}}.
    $$
    If $A \leq 0$, then:
    $$
    \E_{x\sim q(x)}[xe^{-\frac{x}{\gamma B}}]<(\frac{3\gamma B}{20})e^{-\frac{A}{B}}.
    $$
    According to our assumption, this bound can be kept under a sufficiently small $\delta_0$. Let $H(\frac{1}{t})=\E_{x\sim q(x)}[e^{\frac{-x}{t}}]$. Using Lagrange's mean value theorem, there can be a $l \in [\gamma,1]$, satisfy:
    $$
    \frac{H(\frac{1}{B})-H(\frac{1}{\gamma B})}{(\frac{1}{B}-\frac{1}{\gamma B})}=H'(\frac{1}{lB}).
    $$
    Noticed that $(\frac{1}{B}-\frac{1}{\gamma B})<0$. Under our assumption, we have known that:
    
    (1) $A>0$:
    $$
    H'(\frac{1}{lB})=\E_{x\sim q(x)}[-x e^{-\frac{x}{lB}}]>-(\frac{3}{20}\gamma B+\gamma A(\frac{20}{e^2} +10e^{-e^{\frac{1}{2}}}+ \frac{1}{2}-\frac{1}{2e}))e^{-\frac{A}{B}},
    $$
    So:
    $$
    (\frac{1}{B}-\frac{1}{\gamma B})H'(\frac{1}{lB})<\frac{1-\gamma}{\gamma}(\frac{3}{20}\gamma +\frac{\gamma A}{B}(\frac{20}{e^2} +10e^{-e^{\frac{1}{2}}}+ \frac{1}{2}-\frac{1}{2e}))e^{-\frac{A}{B}}.
    $$
    (2) $A\leq 0$:
    $$
    H'(\frac{1}{lB})=\E_{x\sim q(x)}[-x e^{-\frac{x}{lB}}]>-(\frac{3\gamma B}{20})e^{-\frac{A}{B}},
    $$
    $$
    (\frac{1}{B}-\frac{1}{\gamma B})H'(\frac{1}{lB})<\frac{1-\gamma}{\gamma}(\frac{3}{20}\gamma)e^{-\frac{A}{B}}.
    $$
    Thus, we can rearrange the above equation to obtain:
    
    (1) $A>0$:
    $$
    {\rm KL}(q(x)||p(x))<log(\frac{1}{\gamma})+\frac{1}{B}(\frac{\gamma-1}{\gamma})\E_{x\sim q(x)}[x]+\frac{1-\gamma}{\gamma}(\frac{3}{20}\gamma +\frac{\gamma A}{B}(\frac{20}{e^2} +10e^{-e^{\frac{1}{2}}}+ \frac{1}{2}-\frac{1}{2e})).
    $$
    (2) $A \leq 0$:
    $$
    {\rm KL}(q(x)||p(x))<log(\frac{1}{\gamma})+\frac{1}{B}(\frac{\gamma-1}{\gamma})\E_{x\sim q(x)}[x]+\frac{1-\gamma}{\gamma}(\frac{3}{20}\gamma).
    $$
    We have shown that $\E_{x\sim q(x)}[x]=\gamma A+\gamma B v$, where $v$ is the Euler–Mascheroni constant. 

    (1) $A>0$:
    $$
    {\rm KL}(q(x)||p(x))<log(\frac{1}{\gamma})+(\gamma-1)(\frac{A}{B}+v)+\frac{1-\gamma}{\gamma}(\frac{3}{20}\gamma +\frac{\gamma A}{B}(\frac{20}{e^2} +10e^{-e^{\frac{1}{2}}}+ \frac{1}{2}-\frac{1}{2e})).
    $$
    (2) $A \leq 0$:
    $$
    {\rm KL}(q(x)||p(x))<log(\frac{1}{\gamma})+(\gamma-1)(\frac{A}{B}+v)+\frac{1-\gamma}{\gamma}(\frac{3}{20}\gamma).
    $$
    Finally, summarizing the above equation yields the KL bound:

    (1) $A>0$:
    $$
    {\rm KL}(q(x)||p(x))<log(\frac{1}{\gamma})+(1-\gamma)[\frac{A}{B}(\frac{20}{e^2} +10e^{-e^{\frac{1}{2}}}- \frac{1}{2}-\frac{1}{2e})+\frac{3}{20}-v].
    $$
    (2) $A \leq 0$:
    $$
    {\rm KL}(q(x)||p(x))<log(\frac{1}{\gamma})+(1-\gamma)[\frac{3}{20}-\frac{A}{B}-v].
    $$
    Let us prove that these two upper bounds are well-defined. 

    (1)$A>0$:
    
    Let $f(A)=log(\frac{1}{\gamma})+(1-\gamma)[\frac{A}{B}(\frac{20}{e^2} +10e^{-e^{\frac{1}{2}}}- \frac{1}{2}-\frac{1}{2e})+\frac{3}{20}-v]$. Obviously $f(A)>f(0)$, where:
    $$
    f(0)=log(\frac{1}{\gamma})+(1-\gamma)[\frac{3}{20}-v]>log(\frac{1}{\gamma})+(\gamma-1)[\frac{9}{20}]=g(\gamma),
    $$
    $$
    \frac{\partial g}{\partial \gamma}=-\frac{1}{\gamma}+\frac{9}{20}<0.
    $$
    So:
    $$
    f(A)>f(0)=g(\gamma)>g(1)=0.
    $$

    (2)$A \leq 0$:
     
     Let $f(A)=log(\frac{1}{\gamma})+(1-\gamma)[\frac{3}{20}-\frac{A}{B}-v]$. Obviously, it still holds that  $f(A) \geq f(0)$, it is consistent with the above discussion. So:
     $$
     f(A)\geq f(0) = g(\gamma) > g(1)=0.
     $$
     Therefore, these two bounds are well-defined and meaningful, they indicate that two distributions can be considered approximately identical within the KL divergence error limit. It is obvious that when $\gamma=1$, ${\rm KL}(q(x)||p(x))=0$.

     Next, let's discuss the order of this error, as defined, $\kappa$ is in a small neighborhood near the zero with the radius $\kappa_0$, then the growth order for KL divergence ${\rm KL}(q(x)||p(x))$ is:
     $$
     O(log(\frac{1}{1-\kappa_0})+\kappa_0\frac{A}{B}).
     $$
    Within this error control range, we consider that $\gamma$ does not affect the distribution type and coefficient magnitude, allowing us to apply Lemma~\ref{Lemma4} now.
    
    According to Lemma~\ref{Lemma4}:
    $$
    {\rm \varepsilon}^{\theta}(\vs,\va) \sim {\rm Gumbel}(C_{\theta}(\vs,\va),\beta_{\theta})-{\rm Gumbel}(\beta_{\theta} ln\sum_{i=1}^n e^{\frac{r(\vs',\va_i)+C_{\theta}(\vs',\va_i)}{\beta_{\theta}}},\beta_{\theta}),
    $$
    Which means:
    $$
    {\rm \varepsilon}^\theta(\vs,\va) \sim {\rm Logistic} (C_{\theta}(\vs,\va)-\beta_{\theta} ln\sum_{i=1}^n e^{\frac{r(\vs',\va_i)+C_{\theta}(\vs',\va_i)}{\beta_{\theta}}},\beta_{\theta}).
    $$
    The discussion of Remark~\ref{remark2} is consistent with Lemma~\ref{Lemma3}. The Remark~\ref{remark2} removed the connection between $C_{\theta}$ and pair $(\vs,\va)$, so in this case:
    $$
    {\rm \varepsilon}^{\theta}(\vs,\va) \sim {\rm Logistic} (C_{\theta}-\beta_{\theta} ln\sum_{i=1}^n e^{\frac{r(\vs',\va_i)+C_{\theta}}{\beta_{\theta}}},\beta_{\theta}).
    $$
    So, we have shown that:
    $$
    {\rm \varepsilon}^{\theta}(\vs,\va) \sim {\rm Logistic} (-\beta_{\theta} ln\sum_{i=1}^n e^{\frac{r(\vs',\va_i)}{\beta_{\theta}}},\beta_{\theta}).
    $$
\end{proof}

\subsection{Proof for Theorem~\ref{theorem1-extract}}
\label{AppendixTheorem2}
\paragraph{Theorem 2}
\textbf{(Positive Scaling upper bounds under Remark~\ref{remark2}):} 
Denote $r^{+}$ and $r^{-}$ the positive and negative rewards with $r>0$ and $r<0$, respectively. With $i_1+i_2+i_3=n$, assume that:
\begin{equation*}
    \sum_{i=1}^n e^{\frac{r(\vs',\va_i)}{\beta_{\theta}}}=\sum_{i=1}^{i_1} e^{\frac{r^{+}(\vs',\va_i)}{\beta_{\theta}}}+\sum_{i=1}^{i_2} e^{\frac{r^{-}(\vs',\va_i)}{\beta_{\theta}}}+i_3.
\end{equation*}
If it satisfies
\begin{enumerate}[leftmargin=*]
    \item $i_1 \neq 0$,
    \item $\sum_{i=1}^{i_1} e^{\frac{r^{+}(\vs',\va_i)}{\beta_{\theta}}}r^{+}(\vs',\va_i)+\sum_{i=1}^{i_2} e^{\frac{r^{-}(\vs',\va_i)}{\beta_{\theta}}}r^{-}(\vs',\va_i)<0$,
\end{enumerate}
then there exists an optimal scaling ratio $\varphi^*>1$, such that for any scaling ratio $\varphi$ that can effectively reduce the expectation of the Bellman error, it must satisfy
$$
1 \leq \varphi \leq \varphi^*.
$$

\begin{proof}

According to our discussion, $\varepsilon^{\theta}(\vs,\va)$ should satisfy:
$$
    {\rm \varepsilon}^{\theta}(\vs,\va) \sim {\rm Logistic} (-\beta_{\theta} ln\sum_{i=1}^n e^{\frac{r(\vs',\va_i)}{\beta_{\theta}}},\beta_{\theta}).
$$
According to condition (1), which means $-\beta_{\theta} ln\sum_{i=1}^n e^{\frac{r(\vs',\va_i)}{\beta_{\theta}}}<0$, so if the scaling factor $\varphi$ is effective, it should satisfy:
$$
-\beta_{\theta} ln\sum_{i=1}^n e^{\frac{r(\vs',\va_i)}{\beta_{\theta}}} \leq -\beta_{\theta} ln\sum_{i=1}^n e^{\frac{\varphi r(\vs',\va_i)}{\beta_{\theta}}}.
$$

this means:
$$
ln\sum_{i=1}^n e^{\frac{r(\vs',\va_i)}{\beta_{\theta}}} \geq ln\sum_{i=1}^n e^{\frac{\varphi r(\vs',\va_i)}{\beta_{\theta}}},
$$
$$
ln [ \sum_{i=1}^{i_1} e^{\frac{r^{+}(\vs',\va_i)}{\beta_\theta}}+ \sum_{i=1}^{i_2} e^{\frac{r^{-}(\vs',\va_i)}{\beta_\theta}}+i_3] \geq ln [ \sum_{i=1}^{i_1} e^{\frac{\varphi r^{+}(\vs',\va_i)}{\beta_\theta}}+ \sum_{i=1}^{i_2} e^{\frac{\varphi r^{-}(\vs',\va_i)}{\beta_\theta}}+i_3],
$$
which means:
$$
\sum_{i=1}^{i_1} e^{\frac{r^{+}(\vs',\va_i)}{\beta_\theta}}+ \sum_{i=1}^{i_2} e^{\frac{r^{-}(\vs',\va_i)}{\beta_\theta}} \geq \sum_{i=1}^{i_1} e^{\frac{\varphi r^{+}(\vs',\va_i)}{\beta_\theta}}+ \sum_{i=1}^{i_2} e^{\frac{\varphi r^{-}(\vs',\va_i)}{\beta_\theta}}.
$$
Let $G(\varphi)=\sum_{i=1}^{i_1} e^{\frac{\varphi r^{+}(\vs',\va_i)}{\beta_\theta}}+ \sum_{i=1}^{i_2} e^{\frac{\varphi r^{-}(\vs',\va_i)}{\beta_\theta}}$.
$$
\frac{\partial G}{\partial \varphi}=\sum_{i=1}^{i_1} e^{\frac{\varphi r^{+}(\vs',\va_i)}{\beta_\theta}} r^{+}(\vs',\va_i)+ \sum_{i=1}^{i_2} e^{\frac{\varphi r^{-}(\vs',\va_i)}{\beta_\theta}}r^{-}(\vs',\va_i).
$$
Obviously $\frac{\partial G}{\partial \varphi}$ is monotonically increasing w.r.t. $\varphi$. According to our assumption, we have $\frac{\partial G}{\partial \varphi}(1)=\sum_{i=1}^{i_1} e^{\frac{r^{+}(\vs',\va_i)}{\beta_{\theta}}}r^{+}(\vs',\va_i)+\sum_{i=1}^{i_2} e^{\frac{r^{-}(\vs',\va_i)}{\beta_{\theta}}}r^{-}(\vs',\va_i)<0$, $\lim_{x\rightarrow \infty}\frac{\partial G}{\partial \varphi}(x)>0$. According to zero theorem, there exist a $\varphi^* \in (1,+\infty)$, satisfy $\frac{\partial G}{\partial \varphi}(\varphi^*)=0$. when $1\leq \varphi \leq \varphi^*$, $G(1)\geq G(\varphi)$. Which means:
$$
G(1)=\sum_{i=1}^{i_1} e^{\frac{r^{+}(\vs',\va_i)}{\beta_\theta}}+ \sum_{i=1}^{i_2} e^{\frac{r^{-}(\vs',\va_i)}{\beta_\theta}} \geq \sum_{i=1}^{i_1} e^{\frac{\varphi r^{+}(\vs',\va_i)}{\beta_\theta}}+ \sum_{i=1}^{i_2} e^{\frac{\varphi r^{-}(\vs',\va_i)}{\beta_\theta}}=G(\varphi).
$$
That is what we want to prove.
\end{proof}

\subsection{Sovling for Equation (\ref{Eq7})}
\label{Appendix4}
Because:
$$
\pi^*=arg\max_{\pi}[\sum \sP(\vs'|\vs,\va)\pi(\va'|\vs')\left[Q(\vs',\va')-\zeta \log\frac{\pi(\va'|\vs')}{\mu(\va'|\vs')}\right]].
$$
Subject to the following equality constraints:
$$
\sum_{\va'} \pi^*(\va'|\vs')=1.
$$
Then we construct the Lagrange function as:
$$
f(\pi,L)=\sum_{\vs',\va'} \sP(\vs'|\vs,\va)\pi(\va'|\vs')\left[Q(\vs',\va')-\zeta \log\frac{\pi(\va'|\vs')}{\mu(\va'|\vs')}\right]+L\left[\sum_{\va'} \pi(\va'|\vs')-1\right].
$$
Take
$$
\frac{\partial f}{\partial L}=0, \frac{\partial f}{\partial \pi}=0.
$$
We have:
$$
\mu(\va|\vs)e^{\frac{Q(\vs,\va)+L-1}{\zeta}}=\pi(\va|\vs).
$$
$$
log(e^{\frac{L-1}{\zeta}}\sum \mu(\va|\vs)e^{\frac{Q(\vs,\va)}{\zeta}})=log(1)=0 \to L=1-log\sum \mu(\va|\vs)e^{\frac{Q(\vs,\va)}{\zeta}}.
$$
Then take $L$ back, we can have:
$$
\pi^{*}(\va|\vs)=\frac{\mu(\va|\vs)e^{Q(\vs,\va)/\zeta}}{\sum_\va \mu(\va|\vs)e^{Q(\vs,\va)/\zeta}}.
$$

\subsection{Proof for Theorem~\ref{theorem5}}
\label{Appendix9}
\paragraph{Theorem 4}
\textbf{(Relationship between $\rm LLoss$ and $\rm MSELoss)$}
The $\rm MSELoss$ can be used as an approximate estimation of $\rm LLoss$ when $\varepsilon$ is sufficiently small, \ie
\begin{equation*}
    \rm LLoss=ln4+\frac{1}{2} \rm MSELoss+o(\rm \varepsilon^3),
\end{equation*}
where $o(\rm \varepsilon^3)$ third-order infinitesimal of $\varepsilon$ when $\varepsilon$ is sufficiently small.

\begin{proof}
Let $\frac{\varepsilon}{\sigma}=t$, then:
$$
\rm MSELoss=\frac{1}{2} t^2.
$$
$$
\rm LLoss=t+2log(1+e^{-t}).
$$
If we performed a Taylor expansion on $\rm LLoss$, we can have:
$$
\rm LLoss=t+2[ln2-\frac{1}{2}t +\frac{1}{8} t^2+o(t^3)]=ln4+\frac{1}{4}t^2+o(t^3)=ln4+\frac{1}{2} \rm MSELoss+ o(t^3).
$$
\end{proof}
\subsection{Proof for Lemma~\ref{lemma_SE}}
\label{Appendix_Lemma8}
\paragraph{Lemma 8}
The sampling error $S_e$ in (\ref{Eq14}) can be decomposed into Bias and Variance terms. If we define:
$$
\overline{F}(t)=\E_{(x_1,x_2,...,x_N)}\left[\hat{F}^{(x_1,x_2,...,x_N)}_N(t)\right],
$$
then
\begin{equation*}
S_e=\E_t\left[{\rm Variance}(t)+{\rm Bias}(t)\right]={\rm Variance}+{\rm Bias},
\end{equation*}
where
\begin{equation*}
\begin{aligned}
{\rm Variance}(t)&=\E_{(x_1,x_2,...,x_N)}\left[(\hat{F}^{(x_1,x_2,...,x_N)}_N(t))^2\right]-\E_{(x_1,x_2,...,x_N)}^2[\hat{F}^{(x_1,x_2,...,x_N)}_N(t)],\\
{\rm Bias}(t)&=(\overline{F}(t)-F(t))^2.
\end{aligned}
\end{equation*}

\begin{proof}
According to (\ref{Eq14}):
$$
S_e=\E_t\E_{(x_1,x_2,...,x_N)}[(\hat{F}^{(x_1,x_2,...,x_N)}_N(t))^2-2\hat{F}^{(x_1,x_2,...,x_N)}_N(t)F(t)+F^2(t)],
$$
Take $\overline{F}(t)$ in, we have:
$$
S_e=\E_t\E_{(x_1,x_2,...,x_N)}[(\hat{F}^{(x_1,x_2,...,x_N)}_N(t))^2-2\overline{F}(t)F(t)+F^2(t)+\overline{F}^2(t)-\overline{F}^2(t)],
$$
which means:
$$
S_e=\E_t\E_{(x_1,x_2,...,x_N)}[(\hat{F}^{(x_1,x_2,...,x_N)}_N(t))^2-\overline{F}^2(t)+(\overline{F}(t)-F(t))^2].
$$
So $S_e$ can be rewritten with Bias-Variance decomposition, \ie
\begin{equation*}
    S_e=\E_t\left[{\rm Variance}(t)+{\rm Bias}(t)\right]={\rm Variance}+{\rm Bias}.
\end{equation*}
where
\begin{equation*}
\begin{aligned}
{\rm Variance}(t)&=\E_{(x_1,x_2,...,x_N)}\left[(\hat{F}^{(x_1,x_2,...,x_N)}_N(t))^2\right]-\left(\E_{(x_1,x_2,...,x_N)}[\hat{F}^{(x_1,x_2,...,x_N)}_N(t)]\right)^2,\\
{\rm Bias}(t)&=(\overline{F}(t)-F(t))^2.
\end{aligned}
\end{equation*}
\end{proof}
\subsection{Proof for Theorem~\ref{theorem_SE}}
\label{Appendix_Theorem5}
\paragraph{Theorem 3}
\textbf{(The Expectation of order statistics for the Logistic distribution)}
\begin{equation*}
    \E[x_{(i)}]=\left[B[\sum_{k=1}^{i-1} \frac{1}{k}-\sum_{k=1}^{N-i} \frac{1}{k}]+A\right].
\end{equation*}

\begin{proof}
As we have known that:
\begin{equation*}
\begin{aligned}
    \E[x_{(i)}]&=\frac{N!}{B(i-1)!(N-i)!}\int_{-\infty}^{+\infty} \frac{(e^{-\frac{(t-A)}{B}})^{N+1-i}}{(1+e^{-\frac{(t-A)}{B}})^{N+1}}t dt\\
    &=\frac{N!}{(i-1)!(N-i)!}\int_{-\infty}^{+\infty} \frac{(e^{-g})^{N+1-i}}{(1+e^{-g})^{N+1}}(gB+A)dg.
\end{aligned}
\end{equation*}
where we define $\frac{t-A}{B}=g$ and $Bdg=dt$. Divide the integral term to $L_1(N,i)$ and $L_2(N,i)$ simplifies. So:
\begin{equation*}
    \E[x_{(i)}]=\frac{N!}{(i-1)!(N-i)!} \left[BL_1(N,i)+AL_2(N,i)\right].
\end{equation*}
where:
\begin{equation*}
    \int_{-\infty}^{+\infty} \frac{(e^{-g})^{N+1-i}}{(1+e^{-g})^{N+1}}(gB+A)dg = B \underbrace{\int_{-\infty}^{+\infty} \frac{(e^{-g})^{N+1-i}}{(1+e^{-g})^{N+1}}gdg}_{L_1(N,i)}
    +A \underbrace{\int_{-\infty}^{+\infty} \frac{(e^{-g})^{N+1-i}}{(1+e^{-g})^{N+1}}dg}_{L_2(N,i)}.
\end{equation*}
Let us calculate these two parts:
$$
L_2(N,i)=\lim_{W\to +\infty} \int_{-W}^{+W} \frac{(e^{-g})^{N+1-i}}{(1+e^{-g})^{N+1}} dg=\int_{-\infty}^{+\infty} \frac{(e^{-g})^{N+1-i}}{(1+e^{-g})^{N+1}}dg.
$$
Because:
$$
\int_{-W}^{+W} \frac{(e^{-g})^{N+1-i}}{(1+e^{-g})^{N+1}} dg= \frac{1}{N} \int_{-W}^{+W} (e^{-g})^{N-i} d(\frac{1}{(1+e^{-g})^{N}})=\frac{(e^{-g})^{N-i}}{(1+e^{-g})^{N}}|_{-W}^{+W}+\frac{(N-i)}{N} \int_{-W}^{+W}\frac{(e^{-g})^{N-i}}{(1+e^{-g})^{N}} dg.
$$
Noticed that when ($i<N$):
$$
\lim_{W\to +\infty} \frac{(e^{-g})^{N-i}}{(1+e^{-g})^{N}}|_{-W}^{+W}=\lim_{W\to +\infty} \frac{(e^W)^i}{(e^W+1)^N}-\frac{(e^W)^{(N-i)}}{(e^W+1)^N}=0.
$$
Noticed that:
$$
L_2(i,i)=\int_{-\infty}^{+\infty} \frac{(e^{-g})}{(1+e^{-g})^{i+1}}dg=\lim_{W\to +\infty} \frac{1}{i}\int_{-W}^{+W} d(\frac{1}{(1+e^{-g})^{i}})=\frac{1}{i} \lim_{W\to +\infty}[(\frac{e^W}{e^W+1})^i-(\frac{1}{e^W+1})^i]=\frac{1}{i}.
$$
So when $i<N$ we can use this formula:
$$
L_2(N,i)=\int_{-\infty}^{+\infty} \frac{(e^{-g})^{N+1-i}}{(1+e^{-g})^{N+1}}dg=\frac{(N-i)}{N} \int_{-\infty}^{+\infty}\frac{(e^{-g})^{N-i}}{(1+e^{-g})^{N}} dg=\frac{(N-i)}{N} L_2(N-1,i)
$$
When $i=N$ we can have:
$$
L_2(N,i)=\frac{1}{N}
$$
So, in summary:
$$
L_2(N,i)=\frac{(N-i)}{N}\frac{(N-i-1)}{N-1}\frac{(N-i-2)}{N-2}....\frac{1}{i}=\frac{(N-i)!}{N!}(i-1)!.
$$

Let us consider another part $L_1(N,i)$:
$$
L_1(N,i)=\lim_{W\to +\infty} \int_{-W}^{+W} \frac{(e^{-g})^{N+1-i}}{(1+e^{-g})^{N+1}}gdg=\int_{-\infty}^{+\infty} \frac{(e^{-g})^{N+1-i}}{(1+e^{-g})^{N+1}}gdg.
$$
Because:
$$
\int_{-W}^{+W} \frac{(e^{-g})^{N+1-i}}{(1+e^{-g})^{N+1}}gdg= \frac{1}{N} \int_{-W}^{+W} (e^{-g})^{N-i} g d(\frac{1}{(1+e^{-g})^{N}}),
$$
Which means:
$$
\int_{-W}^{+W} \frac{(e^{-g})^{N+1-i}}{(1+e^{-g})^{N+1}}gdg=\frac{1}{N}[\frac{(e^{-g})^{N-i} g}{(1+e^{-g})^{N}}|_{-W}^{+W}+ \int_{-W}^{+W} (N-i) \frac{(e^{-g})^{N-i} g}{(1+e^{-g})^{N}} dg -\int_{-W}^{+W} \frac{(e^{-g})^{N-i}}{(1+e^{-g})^{N}} dg].
$$
Take $\lim_{W\to +\infty}$, notice that when $(i < N)$:
$$
\lim_{W\to +\infty} \frac{(e^{-g})^{N-i} g}{(1+e^{-g})^{N}}|_{-W}^{+W}=\lim_{W\to +\infty} \frac{W(e^W)^i}{(e^W+1)^N}+\frac{W(e^W)^{(N-i)}}{(e^W+1)^N}=0
$$
Noticed that $L_1(N,N)+L_1(N,1)=0$ and $L_1(1,1)=0$, this is because $L_1(1,1)$ is the expectation of $\rm{Logistic}(0,1)$. So:
$$
L_1(N,1)=\frac{N-1}{N}L_1(N-1,1)-\frac{1}{N(N-1)}=-\frac{1}{N}[\frac{1}{N-1}+\frac{1}{N-2}+\frac{1}{N-3}+...\frac{1}{1}].
$$
So:
$$
L_1(N,N)=\frac{1}{N}[\sum_{i=1}^{N-1} \frac{1}{i}].
$$
In addition, we have the following general iterative expression:
$$
L_1(N,i)=\frac{N-i}{N} L_1(N-1,i)-\frac{1}{N}L_2(N-1,i)=\frac{(N-i)!}{N!}(i-1)![\sum_{k=1}^{i-1} \frac{1}{k}-\sum_{k=1}^{N-i} \frac{1}{k}].
$$
In conclusion, we have:
$$
\E[x_{(i)}]=\frac{N!}{(i-1)!(N-i)!} [B\frac{(N-i)!}{N!}(i-1)![\sum_{k=1}^{i-1} \frac{1}{k}-\sum_{k=1}^{N-i} \frac{1}{k}]+A\frac{(N-i)!}{N!}(i-1)!].
$$
Which means:
$$
\E[x_{(k)}]=[B[\sum_{k=1}^{i-1} \frac{1}{k}-\sum_{k=1}^{N-i} \frac{1}{k}]+A].
$$

\end{proof}

\section{Other experiment result}
\subsection{The evolving Bellman error distribution over training time }
\label{Appendix11}

\begin{figure}[H]
    \centering
    \includegraphics[width=\textwidth]{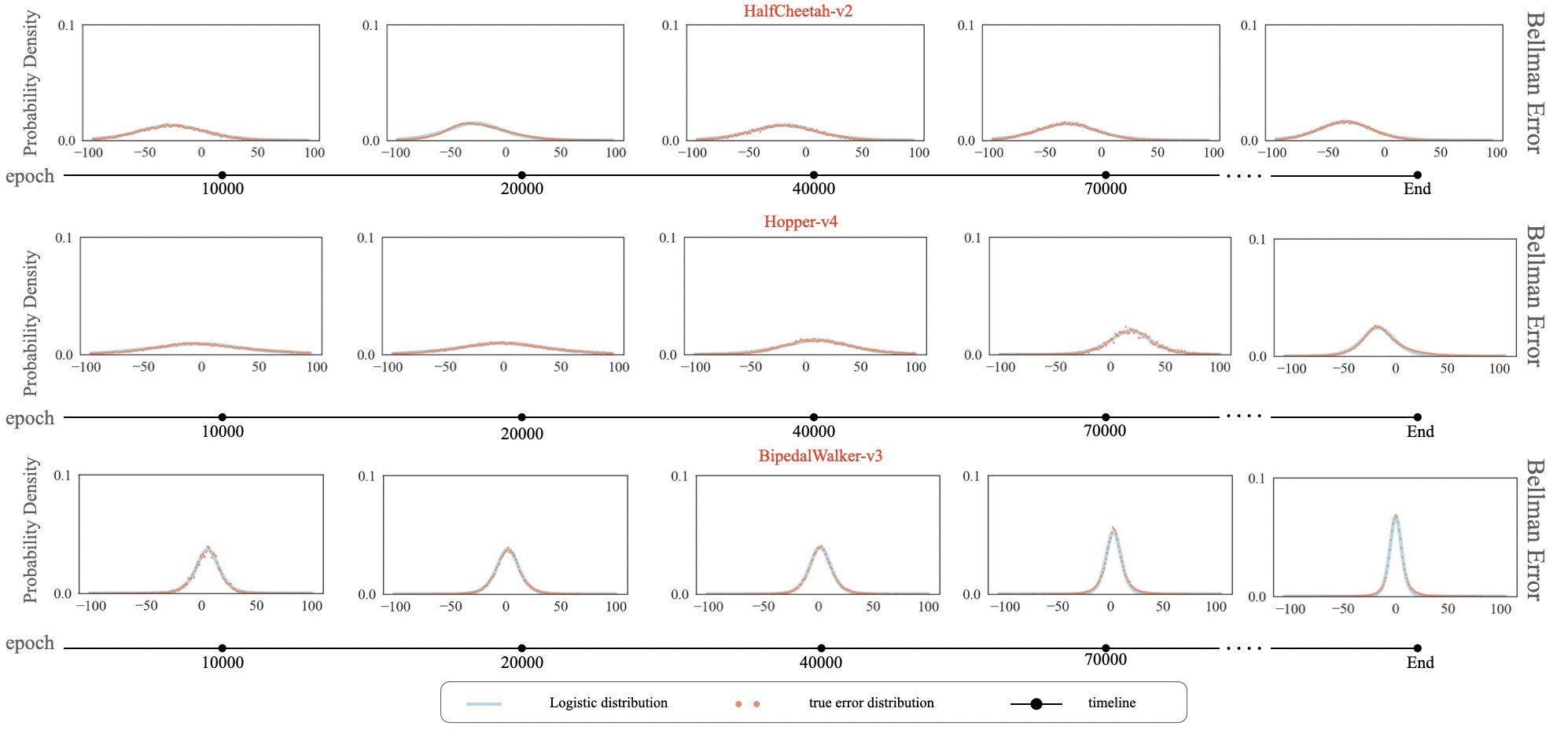}
    \caption{ The evolving distributions of Bellman error computed by (\ref{Eq11}) at different epochs (10000, 20000, 40000, 70000 and final) of online RL training on three environments.}
    \label{figure 20}
\end{figure}

In this section, we present another three distributional images of Bellman errors during training in different environments under the same settings as Figure~\ref{fig:timeline}, revealing a strong alignment with the Logistic distribution. These results are presented in Figure~\ref{figure 20}.

\subsection{The variations of Bellman error distribution in online and offline environments.}
\label{Appendix12}
\begin{figure}[H]
    \centering
    \includegraphics[width=\textwidth]{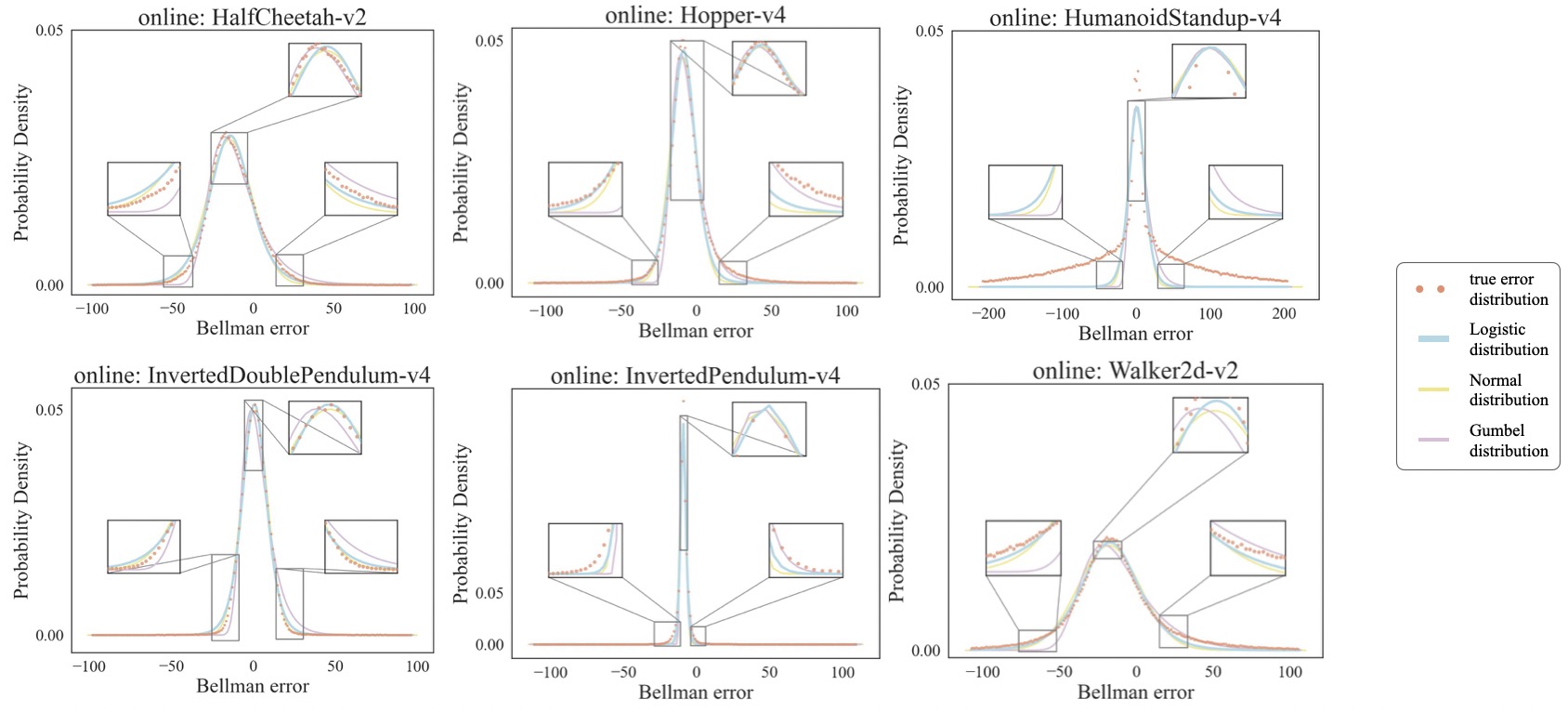}
    \caption{The distribution for Bellman error for the other online environment during half of the training epochs.}
    \label{figure 22}
\end{figure}

\begin{figure}[H]
    \centering
    \includegraphics[width=\textwidth]{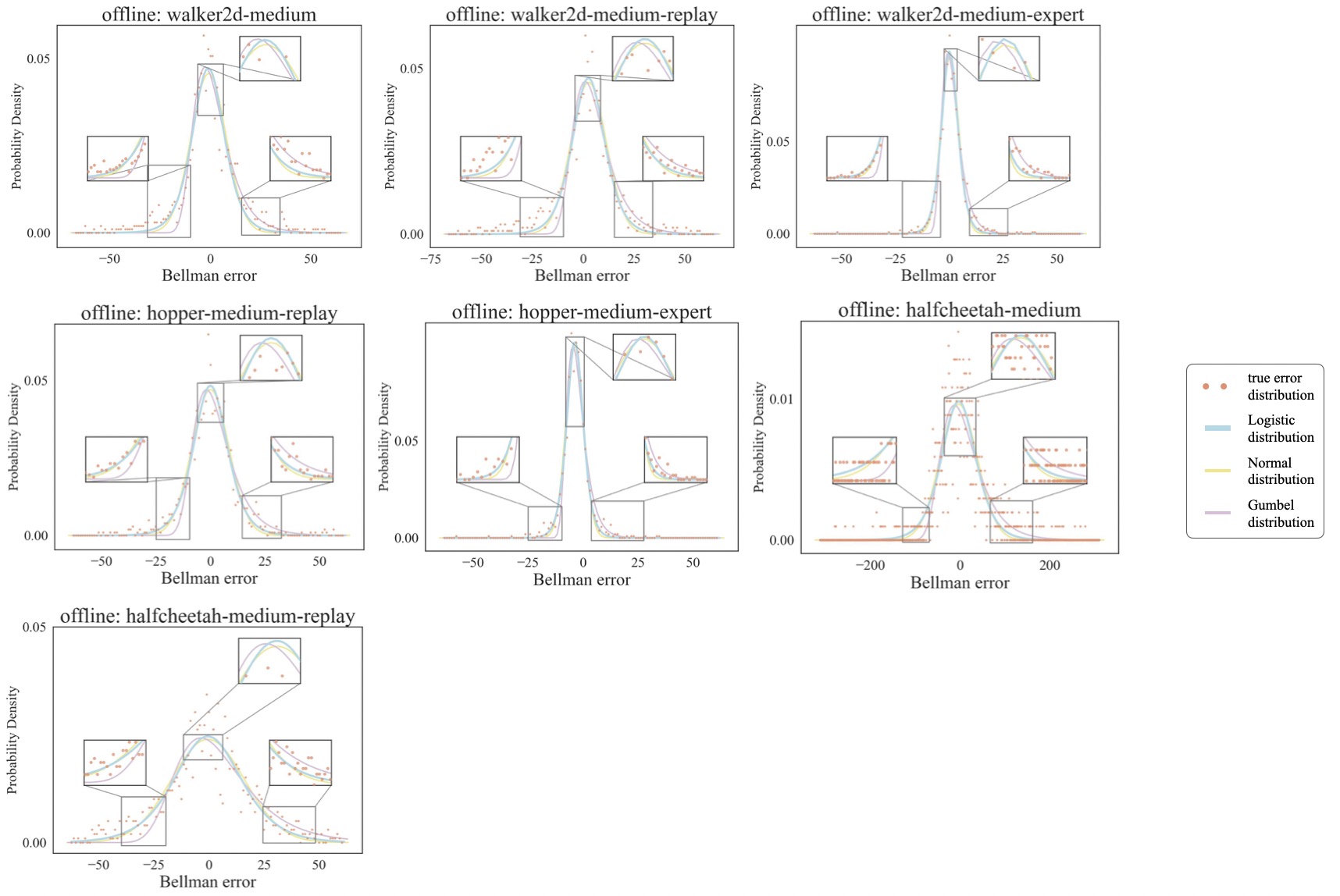}
    \caption{The distribution for Bellman error for the other offline environments during
    the half of the training epochs.}
    \label{figure 21}
\end{figure}

In this section, we will present all distribution details to confirm the characteristics of the Logistic distribution, which is slightly shorter in the tail and slightly longer in the head compared to the Gumbel distribution. In the head region of the distribution, the Logistic distribution fits much better than the Normal and Gumbel distributions, while in the tail region, the Logistic distribution is superior to the Normal distribution and generally outperforms the Gumbel distribution in most environments. These phenomena can be observed from Figures~\ref{figure 21} and \ref{figure 22}. Figure~\ref{figure 21} provides a detailed view of the Bellman error distribution in the offline environment, while Figure~\ref{figure 22} displays the detailed Bellman error distributions in the online environment.

\subsection{The complete version of the containing goodness-of-fit tests and KS tests for Bellman error}
\label{Appendix14}

The specific procedure involves the following three steps:
\begin{enumerate}
    \item[\textbf{step 1}] Collect the data as $x\in [x_1,x_2,...x_n]$ and compute its cumulative distribution function as $F^*(x)$. Fix the distribution to be tested as $F(x)$.
    \item[\textbf{step 2}] Plot the cumulative distribution function $F(x)$ of a specific distribution (Gumbel/Logistic/Normal) under their optimal parameters. For example, after importing the dataset, we first obtain the optimal fitting parameters for the three distributions through fitting. We then use the distributions corresponding to their respective optimal fitting parameters as $F(x)$.
    \item[\textbf{step 3}] Calculate the size of the KS statistic by 
    \begin{equation*}
    KS=\max_i |F^*(x_i)-F(x_i)|.
    \end{equation*}
    A smaller KS statistic indicates a closer similarity between the cumulative distribution function of the data and the specified distribution function. 
\end{enumerate}

\subsection{Sensitivity analysis.}
\label{Appendix15}
\begin{figure}[H]
    \centering
    \includegraphics[width=\textwidth]{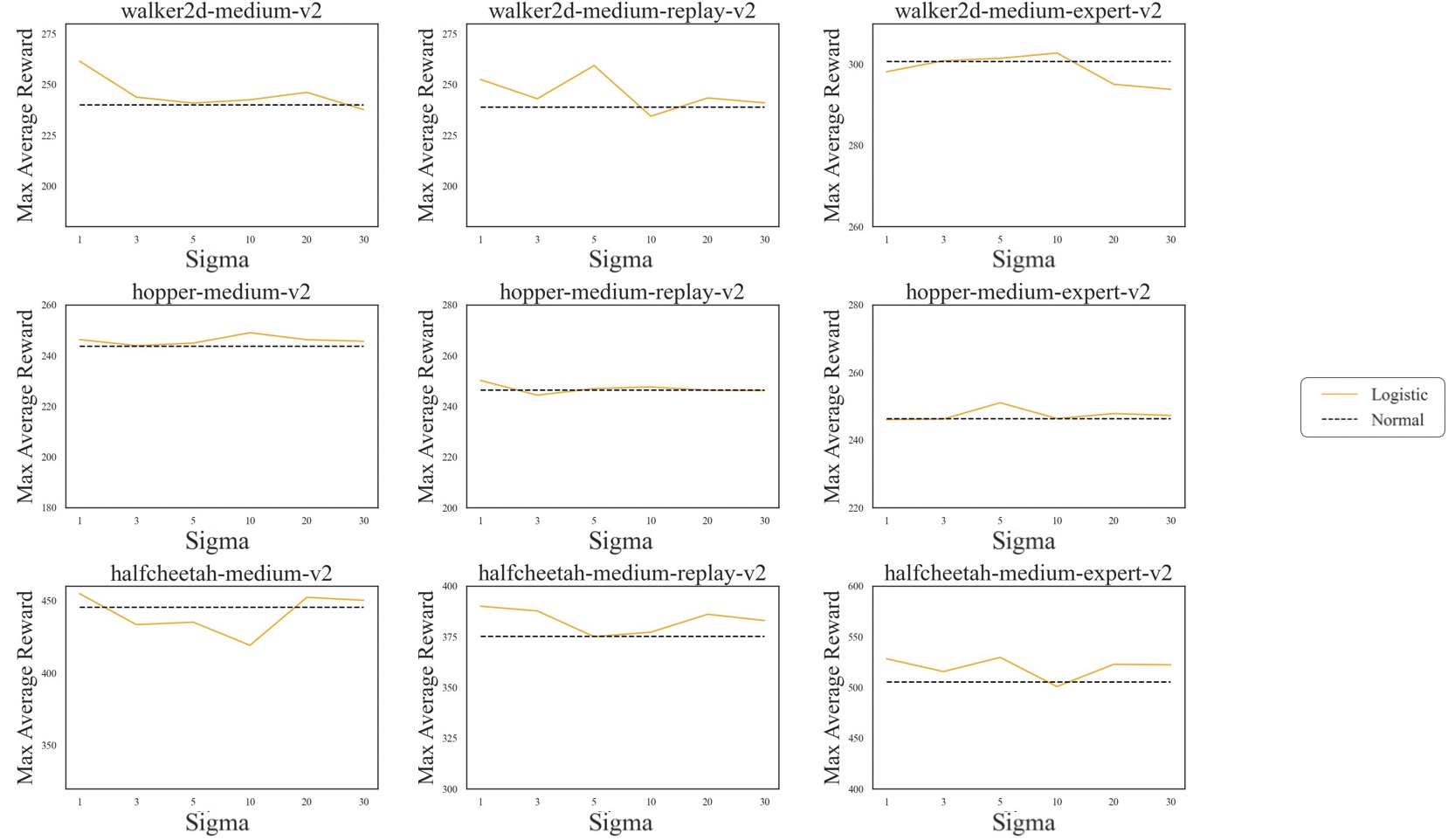}
    \caption{The relationship between the variation of $\sigma$ and the \textbf{maximum average reward} in offline training.}
    \label{figure 26}
\end{figure}

\begin{figure}[H]
    \centering
    \includegraphics[width=\textwidth]{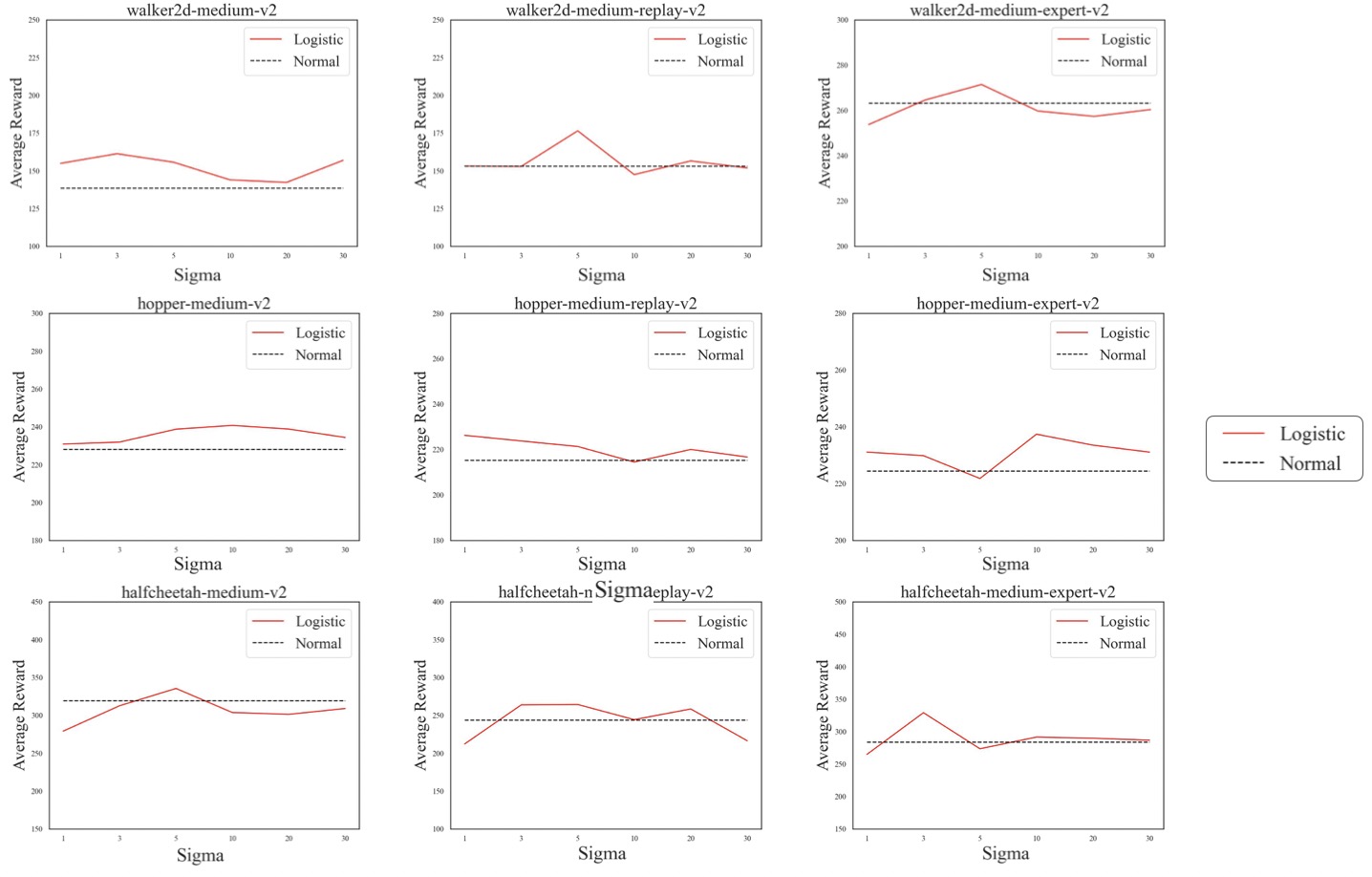}
    \caption{The relationship between the variation of $\sigma$ and the \textbf{average reward} in offline training.}
    \label{figure 27}
\end{figure}

\begin{figure}[H]
    \centering
    \includegraphics[width=\textwidth]{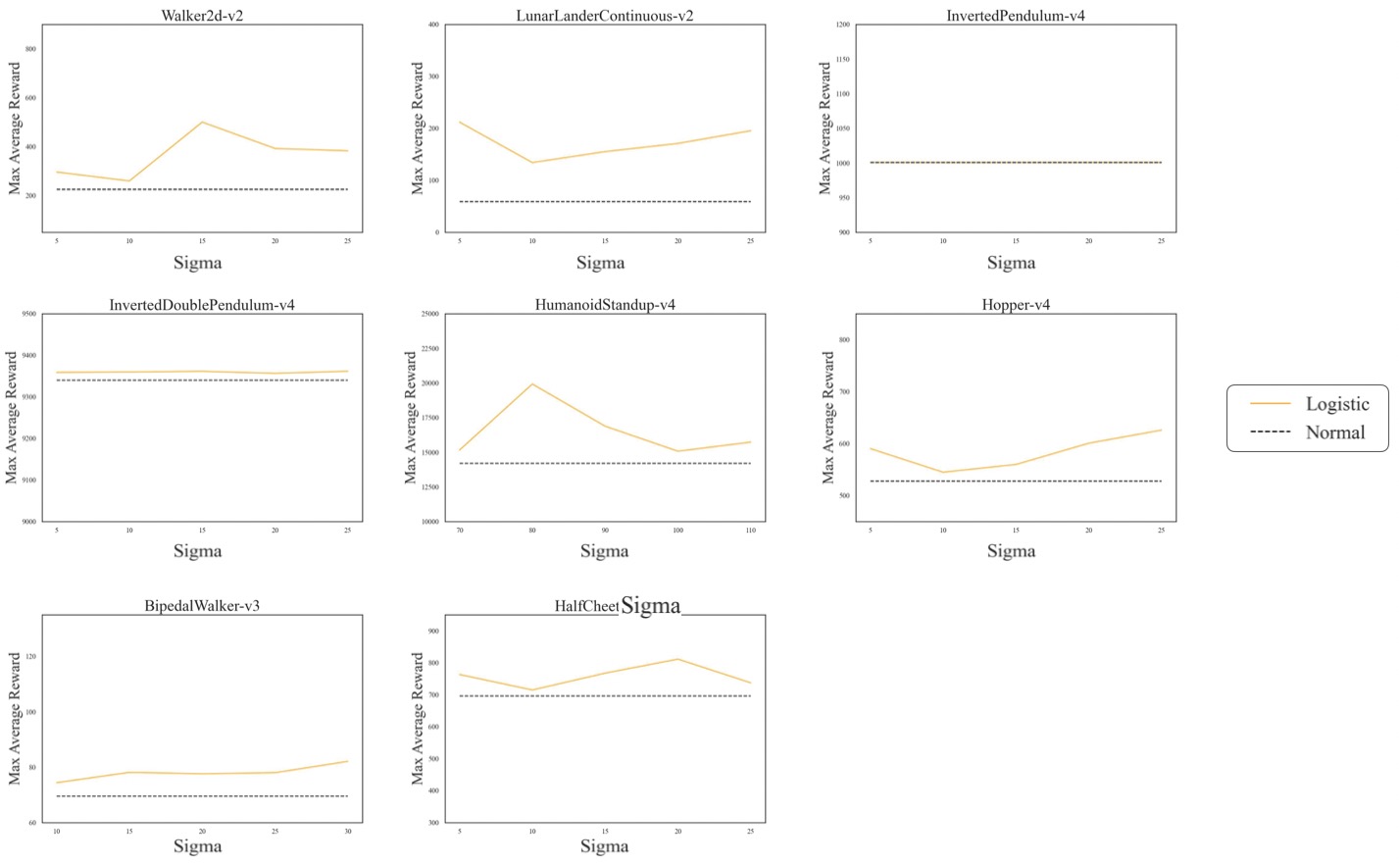}
    \caption{The relationship between the variation of $\sigma$ and the \textbf{maximum average reward} in online training.}
    \label{figure 28}
\end{figure}

\begin{figure}[H]
    \centering
    \includegraphics[width=\textwidth]{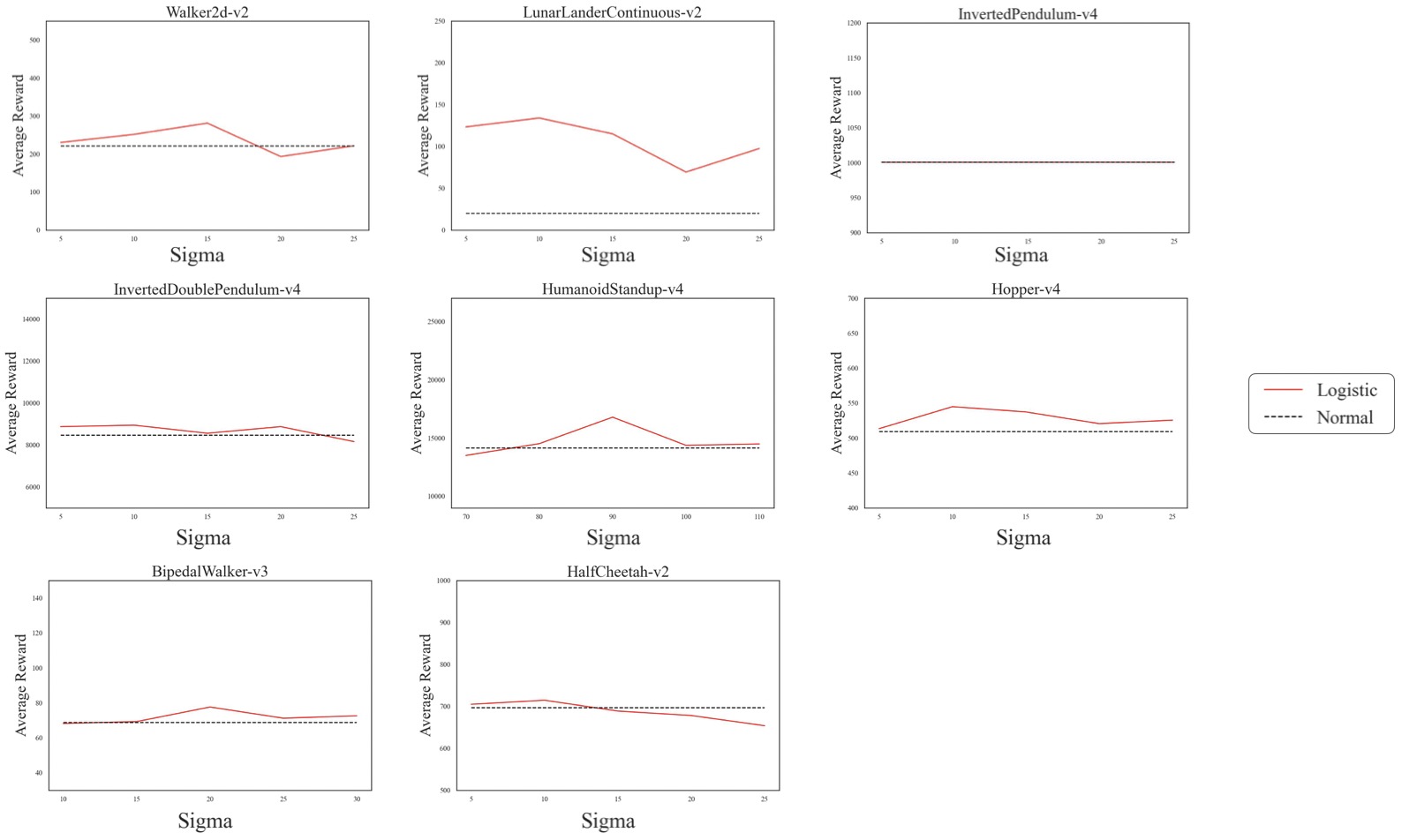}
    \caption{The relationship between the variation of $\sigma$ and the \textbf{average reward} in online training.}
    \label{figure 29}
\end{figure}

In this section, based on the empirical results obtained by adjusting $\sigma$, we provide a range of $\sigma$ variations and their relationship with maximum average reward and final training average reward. From the graphs, it is evident that within a certain range of $\sigma$ variations, $\rm LLoss$ outperforms $\rm MSELoss$ in both the maximum average reward and the final training average reward. This indicates that even within a small range of $\sigma$ adjustments, $\rm LLoss$ consistently yields superior results compared to $\rm MSELoss$ and exhibits a degree of robustness. Figures~\ref{figure 26} and \ref{figure 27} depicts the variations in maximum average reward and final average reward for different $\sigma$ values in the offline setting, with the dashed line representing the $\rm MSELoss$ standard for IQL. Figures~\ref{figure 28} and \ref{figure 29} similarly show the variations in maximum average reward and final average reward for different $\sigma$ values in the online setting, with the dashed line representing the $\rm MSELoss$ standard for SAC.


\end{document}